\pdfoutput=1

\documentclass[11pt]{article}

\usepackage[preprint]{acl}

\usepackage{times}
\usepackage{latexsym}

\usepackage[T1]{fontenc}

\usepackage[utf8]{inputenc}

\usepackage{microtype}

\usepackage{inconsolata}

\usepackage{graphicx}
\usepackage{subcaption}
\usepackage{caption}

\usepackage{balance}

\usepackage{amsmath}
\usepackage{amssymb}
\usepackage{mathtools}
\usepackage{amsthm}
\usepackage{algorithm}
\usepackage{algorithmic}
\usepackage{xcolor} 
\usepackage{colortbl} 

\usepackage{enumitem}
\usepackage{svg}
\usepackage{booktabs}
\usepackage{multirow}

\usepackage[most]{tcolorbox}
\usepackage{dashrule}
\usepackage{wrapfig}

\usepackage[capitalize,noabbrev]{cleveref}
\crefname{section}{Sec.}{Secs.}
\Crefname{section}{Section}{Sections}
\Crefname{table}{Table}{Tables}
\crefname{table}{Tab.}{Tabs.}

\newcommand{\ours}[0]{VersaTune\ }

\title{VersaTune: An Efficient Data Composition Framework for Training Multi-Capability LLMs}

\author{Keer Lu$^{\dagger}$, Keshi Zhao$^{\dagger}$, Zhuoran Zhang$^{\ddagger}$, Zheng Liang$^{\mathsection}$, Da Pan$^{\mathsection}$, Shusen Zhang$^{\mathsection}$, \\
\textbf{Xin Wu$^{\mathsection}$, Guosheng Dong$^{\mathsection}$, Bin Cui$^{\ddagger}$, Tengjiao Wang$^{\ddagger}$, Wentao Zhang$^{\dagger}$} \\
$^\dagger$Center for Data Science, Academy for Advanced Interdisciplinary Studies, Peking University, \\ 
$^\ddagger$School of CS \& Key Lab of High Confidence Software Technologies (MOE), Peking University,  \\
$^\mathsection$Baichuan Inc. \\
\{keer.lu, zhaoks\}@stu.pku.edu.cn, \{zhangzhuoran, bin.cui, tjwang, wentao.zhang\}@pku.edu.cn \\
\{liangzheng, panda, zhangshusen, wuxin, dongguosheng\}@baichuan-inc.com \\
}

\begin{document}
\maketitle
\begin{abstract}

As demonstrated by the proprietary Large Language Models (LLMs) such as GPT and Claude series, LLMs have the potential to achieve remarkable proficiency across a wide range of domains, including law, medicine, finance, science, code, etc., all within a single model. 
These capabilities are further augmented during the Supervised Fine-Tuning (SFT) phase. Despite their potential, existing work mainly focuses on domain-specific enhancements during fine-tuning, the challenge of which lies in catastrophic forgetting of knowledge across other domains. In this study, we introduce \textbf{\textit{VersaTune}}, a novel data composition framework designed for enhancing LLMs' overall multi-domain capabilities during training. 
We begin with detecting the distribution of domain-specific knowledge within the base model, followed by the training data composition that aligns with the model's existing knowledge distribution. During the subsequent training process, domain weights are dynamically adjusted based on their learnable potential and forgetting degree. 
Experimental results indicate that VersaTune is effective in multi-domain fostering, with an improvement of 35.21\% in the overall multi-ability performances compared to uniform domain weights. Furthermore, we find that Qwen-2.5-32B + VersaTune even surpasses frontier models, including GPT-4o, Claude3.5-Sonnet and DeepSeek-V3 by 0.86\%, 4.76\% and 4.60\%. Additionally, in scenarios where flexible expansion of a specific domain is required, VersaTune reduces the performance degradation in other domains by 38.77\%, while preserving the training efficacy of the target domain. 
\end{abstract}

\addtocontents{toc}{\protect\setcounter{tocdepth}{0}}

\section{Introduction}
\label{sec:intro}
Large Language Models (LLMs) have become a cornerstone in Artificial Intelligence (AI)~\cite{achiam2023gpt,dwivedi2021artificial,lewkowycz2022solving}, particularly for Natural Language Processing tasks~\cite{brown2020language,devlin2018bert,radford2019language}, 
reshaping AI research and applications in domains such as law~\cite{cui2023chatlaw}, medicine~\cite{singhal2023large,thirunavukarasu2023large}, finance~\cite{li2023large,wu2023bloomberggpt}, science~\cite{beltagy2019scibert,taylor2022galactica} and code~\cite{liu2024your,roziere2023code}. 
The success of LLMs stems from their capabilities to automatically learn and distill hierarchical data representations, making them highly effective for complex tasks~\cite{nie2023angel}.
In order to further enhance such abilities across these areas, LLMs typically undergo the supervised fine-tuning (SFT) stages on domain-specific datasets.


As demonstrated by the robust performances of state-of-the-art LLMs such as GPT-4~\cite{achiam2023gpt} 
and Gemini~\cite{team2023gemini}, \textit{LLMs have the potential to master multiple tasks across all specific domains within a single model}. 
However, most existing research on supervised fine-tuning tends to merely concentrate on a single ability of LLMs~\cite{dong2023abilities,xu2024large}, with the multi-domain performance on composite data of essentially different downstream tasks being less studied. 
We try to enhance the overall multitasking performance of LLMs across various domains by optimizing data mixing ratios during training: 


\begin{quote}
\em
How to design a data composition strategy during SFT stages that could achieve overall multi-domain capabilities? 
\end{quote}


Through analysis, we identified that the challenges associated with data composition strategies stem from the following three key aspects:

\textbf{\textit{\hypertarget{C1}{C1}: Catastrophic Forgetting.}} 
Given the fundamental differences between tasks of various domains, for multi-domain SFT, the sequential training strategy across multiple phases, where each phase exclusively utilizes a single-domain dataset for training, can easily lead to significant performance drop of prior knowledge, which is well-known as Catastrophic Forgetting~\cite{kaushik2021understanding,mccloskey1989catastrophic}, as depicted in \Cref{tab:catastrophic_forgetting} and \Cref{fig:catastrophic_forgetting}.
It hinders the versatile fine-tuning performance of a model across multiple domains~\cite{de2021continual,dong2023abilities,yuan2022hype}. 
Therefore, mixing data from different domains is crucial for mitigating catastrophic forgetting during training, enhancing the overall performance and adaptability.

\textbf{\textit{\hypertarget{C2}{C2}: Low Efficiency.}} 
Existing data composition research during the supervised fine-tuning phase for LLMs is still in its initial stages, with most strategies based on heuristic or manually determined rules~\cite{wang2023data,albalak2024survey,dubey2024llama}. 
One of the common baselines is defining domain weights referring to natural domain sizes, which weights all individual data points equally.
Such approaches struggle to optimally balance different domains, failing to maximize the overall training effectiveness for multiple abilities. 
There lacks a well-defined methodology that efficiently enhances the versatile capabilities of LLMs across multiple domains during the SFT stage.

\textbf{\textit{\hypertarget{C3}{C3}: Low Flexibility in Domain Expansion.}} 
Existing SFT approaches for specific domain abilities typically pre-determine the proportions of different datasets according to prior experience~\cite{azerbayev2023llemma,roziere2023code}. 
Such strategies lack the flexibility to dynamically adjust the data mixing ratios of different domains during the training process, which does not allow for real-time feedback from LLMs to inform and optimize the data composition. 
This static approach hinders the minimization of performance loss in other domains as LLMs undergo specialized training.

\begin{figure*}[ht]
    \begin{center}
    \centerline{\includegraphics[width=1\textwidth]{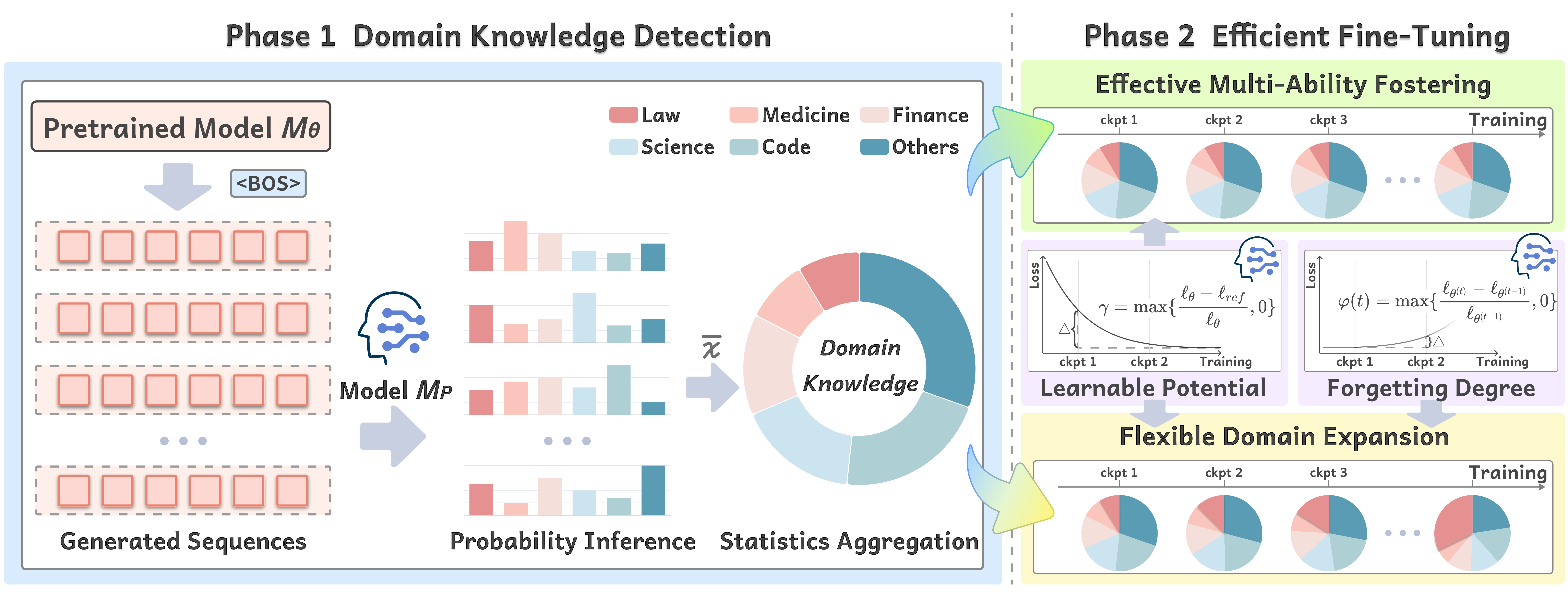}}
    \caption{Overview of \textbf{\textit{VersaTune}}. 
    We begin by probing the knowledge distribution within the base model $M_\theta$, utilizing a proprietary model $M_P$ to estimate the probability of sequences generated by $M_\theta$ belonging to various domains. 
    Throughout the efficient fine-tuning process, we dynamically adjust the data domain ratios in response to $M_\theta$'s real-time performance feedback, with learnable potential and forgetting degree serving as evaluative metrics. }
    \label{fig:pipeline}
    \end{center}
    \vskip -0.4in
\end{figure*}

To address these challenges, we introduce \textbf{\textit{VersaTune}}, a novel data composition framework to enhance models' overall performances across different domains during supervised fine-tuning. 
We first detect the proportion distribution of domain knowledge within the target model (\Cref{subsec:knowledge_distribution_detection}), followed by data composition based on the existing distribution for multi-ability enhancement (\Cref{subsec:effective_multi_ability_fostering}) and flexible domain expansion (\Cref{subsec:flexible_domain_expansion}). 
Our contributions are as follows: 

\begin{itemize}[leftmargin=*]
    \item \underline{\textit{Knowledge Consistency Training.}} 
    We introduce the concept of \textit{knowledge consistency training} for LLMs' multi-capability development, which enables the model to continue learning from datasets that possess a knowledge distribution aligned with its pre-existing knowledge feature. 
    \item \underline{\textit{Multi-Capability Data Composition Framework.}} We propose VersaTune, a novel data composition framework that leverages the model's intrinsic domain knowledge distribution to optimize the training data proportion. \ours is designed to enhance the overall performance across multiple domains (\Cref{subsec:effective_multi_ability_fostering}), as well as to provide flexible expansion for specific domains while minimizing the performance degradation in other domains (\Cref{subsec:flexible_domain_expansion}).
    \item \underline{\textit{Performance and Effectiveness.}} 
    Our extensive evaluations across domains demonstrate that \ours can achieve an improvement of 35.21\% in versatile fine-tuning for multiple domains. Notably, we find that our Qwen-2.5-32B + \ours even outperforms frontier models including GPT-4o, Claude3.5-Sonnet and DeepSeek-V3 by 0.86\%, 4.76\% and 4.60\%. Furthermore, when focusing on specific-domain expansion, \ours maintains training effectiveness in the target domain while reducing performance degradation in other non-target domains by 38.77\%.
\end{itemize}

\section{VersaTune}
\label{sec:versa_tune}

In this section, we introduce VersaTune, a data composition framework designed for multi-capability training, aiming to effectively compose data from multiple domains and optimize the data proportion during training. \Cref{fig:pipeline} presents the workflow of VersaTune, which generally contains two phases.

\subsection{Phase 1: Domain Knowledge Detection}
\label{subsec:knowledge_distribution_detection}

Here, we first present a domain mixing strategy for fine-tuning a LLM that possesses a comprehensive multitask capability (\Cref{subsec:knowledge_consistency_training}). This approach is designed to align with the inherent domain knowledge distribution within the base model waiting for subsequent training. Following this, we describe the method for detecting domain knowledge proportion of the base model, which is crucial for informing the fine-tuning process (\Cref{subsec:knowledge_distribution_detection}).

\subsubsection{Knowledge Consistency Training}
\label{subsec:knowledge_consistency_training}

Previous research on data mixing ratios during the SFT phase for LLMs has predominantly focused on enhancing capabilities within a specific domain, often utilizing only data from that domain or employing heuristic, experience-based data proportions. We argue that such strategies can significantly impair the LLM's abilities in other domains. In the fine-tuning stage, maintaining a robust overall capability across various domains is crucial. 

What data mixing strategy effectively boosts the versatile performance of LLMs across domains during the SFT phase? We propose the statement:

\hypertarget{statement1}{\textbf{\textit{Statement 1}}}
\textit{An LLM fine-tuned with domain-specific data proportions $P_{SFT}(x)$ that align with its pretrained output distributions $P_{knowledge}(x)$ will exhibit enhanced and balanced performance across these domains, compared to a model fine-tuned with a non-matching data distribution. Formally, the relationship can be represented as:}

\begin{equation}
\label{eq:statement1}
    P_{SFT}(x) \approx P_{knowledge}(x), \forall x \in \chi
\end{equation}
\textit{where $\chi$ denotes the set of all possible data points.}




The rationale behind this statement is rooted in the observation that during the pretraining phase, LLMs develop a general understanding of language features and domain-specific knowledge. By maintaining the same distribution of knowledge during fine-tuning, the model can build upon this pre-existing knowledge, thereby enhancing learning efficiency and robustness.


\subsubsection{Knowledge Distribution Detection}
\label{subsubsec:knowledge_distribution_detection}

Drawing on prior research into knowledge identification methods~\cite{gekhman2024does,zhao2023knowing} and training data inference strategies for LLMs~\cite{CVPR_distribution_estimation,hayase2024data}, we propose a structured approach to efficiently detect domain knowledge based on statistics. 
The method involves the generation of textual outputs from the base model $M_{\theta}$ waiting for fine-tuning, followed by classification into predefined domains referring to a proprietary model $M_P$. The process is repeated multiple times to ensure statistical robustness.

Assuming the data corpus contains $k$ distinct domains, 
as shown in \Cref{alg:domain_detection}, we first prompt the base model poised for fine-tuning $M_{\theta}$ with the 
\hyperlink{BOS}{Beginning of Sequence (\textless BOS\textgreater)} 
tokens to generate a set of $N_S$ data entries 
$\mathcal{S}=\{s_i\}_{i=1}^{N_S}$ (Line 3). 
Subsequently, we employ a proprietary model $M_P$ to infer probabilities that these $N_S$ entries belong to each domain (Line 5-7). We then calculate a weighted average of the probability distributions for all data across these domains, thereby deriving the domain knowledge distribution of the current base model $M_{\theta}$ (Line 9). To ensure statistical robustness, the process is iteratively conducted $T$ times, and we use the mean of these $T$ iterations as the estimated result for knowledge distribution.

\subsection{Phase 2: Fine-Tuning Multi-Ability LLMs Efficiently}
\label{subsec:fine_tuning_multi_ability_llm}

Having detected the distribution of domain knowledge within the base model, we will now utilize these findings to guide our multi-ability SFT process. The approaches are designed to enhance the overall performance of the fine-tuned model across a spectrum of multi-domain tasks (\Cref{subsec:effective_multi_ability_fostering}), as well as to facilitate the flexible expansion of capabilities in specific domains (\Cref{subsec:flexible_domain_expansion}).

\textbf{Setting.} 
Our goal is to construct a composite dataset covering $k$ specific domains, which can be 
denoted as 
$\mathcal{D}_{train} = \{ (D_{train}^j, P_j)\}_{j=1}^{k}$ 
with each tuple representing a specific domain and its corresponding proportion, 
such that training a model on dataset $\mathcal{D}_{train}$ could achieve overall lower loss on a uniformly distributed composite target validation dataset 
$\mathcal{D}_{val} = \{ (D_{val}^j, 1 / k)\}_{j=1}^{k}$ or meet the flexible domain expansion while preserving the performances in other domains. 
The specialized capabilities of LLMs are measured using downstream tasks related to different domains (e.g., FinBen~\cite{xie2024finben} for financial performances).

\begin{algorithm}[tb]
\caption{Knowledge Distribution Detection}
\label{alg:domain_detection}
\begin{flushleft}
\textbf{Input}: Base model $M_{\theta}$, Proprietary model $M_P$, Hyperparameters: sample number $N_S$, maximum iterations $T$\\
\textbf{Parameter}: Data samples $\mathcal{S}$ generated from $M_{\theta}$ \\
\textbf{Output}: Domain distribution $\vec{P}$ \\
Define $\vec{p}$: domain probability distribution of data sample $s$ \\
\end{flushleft}
\begin{algorithmic}[1]
\FOR{$t = 1, 2, \ldots, T$}
    \STATE /* \textit{Step 1: Data Generation} */
    \STATE Generate data samples from the base model: $\mathcal{S}=\{s_i\}_{i=1}^{N_S}$ where $s_i = M_{\theta}(\textless BOS \textgreater)$
    \STATE /* \textit{Step 2: Domain Probability Inference} */
    \FOR{each data sample $s_i$ in $\mathcal{S}$}
        \STATE Provide domain probability of $s_i$ referring to the proprietary model $M_P$: \\ \quad \quad $\vec{p_i} = (p_{ij})_{j=1}^{k} \gets M_P(s_i)$
    \ENDFOR
    \STATE /* \textit{Step 3: Statistics Aggregation} */
    \STATE Estimate domain knowledge distribution: \\ \mbox{$\vec{P}^{(t)} = (P_j^{(t)})_{j=1}^k$ where $P_j^{(t)} = \frac{1}{N_S}\sum_{i=1}^{N_S}p_{ij}$}
\ENDFOR
\STATE \mbox{\textbf{Return} $\vec{P} = (P_j)_{j=1}^k$ where $P_j = \frac{1}{T}\sum_{t=1}^{T}P_j^{(t)}$}
\end{algorithmic}
\end{algorithm}

\subsubsection{Preliminary: Learnable Potential and Forgetting Degree of Knowledge}

Before formally introducing the effective multi-task fine-tuning and flexible domain expansion data composing strategies, we will first provide an overview of the evaluation metrics used for the following algorithms in this subsection.

\textbf{\textit{Mastery Ceiling.}} \quad 
We first fine-tuned the small \textit{reference model} $M_{ref}$ for $T_{ref}$ epochs on each domain separately, and identified the epoch with the lowest average loss during this process as the lower bound on the minimum loss attainable by the target model $M_\theta$ for the given domain. This value represents the highest level of domain knowledge mastery that the model can achieve in the context of the current specific domain under given conditions.

\textbf{\textit{Learnable Potential.}} \quad 
We can observe whether a domain could be effectively learned by the model through comparing the difference between the loss of the target model $M_\theta$ and the minimum loss that the reference model $M_{ref}$ can achieve. Based on these principles, we propose \Cref{eq:loss_potential} to score the learnable potential of domain $j$:

\begin{equation}
\label{eq:loss_potential}
    \gamma_j = \max \{\frac{\ell_{\theta}^{j} - \ell_{ref}^{j}}{\ell_{\theta}^{j}}, 0 \}
\end{equation}
where $\ell_{\theta}^{j}$ denotes the loss associated with the target model $M_\theta$ for the $j$-th domain, while $\ell_{ref}^{j}$ signifies the corresponding loss for the reference model $M_{ref}$ within the same domain. 
To mitigate the impact of inherent loss variations across different domains for the model, we have introduced a normalization term into the formula.

\textbf{\textit{Forgetting Degree.}} \quad 
When focusing on expanding a model to a specific domain, our objective is to minimize the loss of the model's knowledge regarding other domains. Here we segment the fine-tuning stage into $T$ distinct checkpoints. We quantify the degree of knowledge loss, or the forgetting of the current domain, by measuring the difference in loss between the $t$-th and $(t - 1)$-th training steps. This difference reflects the model's mastery loss for the tasks associated with the current domain. Based on this principle, we introduce \Cref{eq:forgetting_degree} to assess the model's forgetting degree for domain $j$ at the $t$-th training step.

\begin{equation}
\label{eq:forgetting_degree}
    \varphi_j^{(t)} = \max \{\frac{\ell_{\theta^{(t)}}^{j} - \ell_{\theta^{(t - 1)}}^{j}}{\ell_{\theta^{(t - 1)}}^{j}}, 0 \}
\end{equation}
where $\ell_{\theta^{(t)}}^{j}$ represents the loss at the $t$-th training step associated with the target model $M_\theta$ for the $j$-th domain, while $\ell_{\theta^{(t - 1)}}^{j}$ denotes the loss at the preceding $(t - 1)$-th iteration for the same domain. 
We also incorporated a normalization factor into the equation to counteract the effects of inherent loss disparities among domains.

\subsubsection{Effective Multi-Ability Fostering}
\label{subsec:effective_multi_ability_fostering}

\begin{algorithm}[tb]
\caption{VersaTune Multi-Ability Fine-Tuning (for Domain Robustness)}
\label{alg:multi_ability}
\begin{flushleft}
\textbf{Input}: Base model to be fine-tuned $M_{\theta}^{(0)}$, Domain reference loss $\{\ell_{ref}^{j}\}_{j=1}^{k}$, Hyperparameters: adjustment magnitude $\sigma$, training step number $T$\\
\mbox{\textbf{Parameter}: Data proportion $\{P_{j}\}_{j=1}^{k}$ of dataset} \\
\textbf{Output}: Fine-tuned multi-ability model $M_{\theta}^{(T)}$ \\
Define $\gamma$: learnable potential  of the current domain \\
\end{flushleft}
\begin{algorithmic}[1]
\STATE Initialize domain proportion $\{P_{j}^{(0)}\}_{j=1}^{k}$ according to \Cref{eq:statement1} and \Cref{alg:domain_detection}
\FOR{$t = 1, 2, \ldots, T$}
    \FOR{$j = 1, 2, \ldots, k$}
        \STATE Learnable potential for the $j$-th domain: \\ \quad \quad $\gamma_j^{(t)} = \max \{\frac{\ell_{\theta^{(t)}}^{j} - \ell_{ref}^{j}}{\ell_{\theta^{(t)}}^{j}}, 0 \}$
        \STATE Update domain weights: \\ \quad \quad $P_j^{(t)'} = P_j^{(t-1)}(1 + \sigma 
 \gamma_j^{(t)})$
    \ENDFOR
    \STATE Renormalize domain weights: \\ \quad $P_j^{(t)} = \frac{P_j^{(t)'}}{\sum_{i=1}^{k}P_i^{(t)'}}, \quad \forall j \in \{1, 2, ..., k\}$
    \STATE \mbox{Update parameters of fine-tuned model $M_{\theta}^{(t)}$}
\ENDFOR
\STATE \textbf{Return} Fine-tuned model $M_{\theta}^{(T)}$
\end{algorithmic}
\end{algorithm}

To cultivate the multi-tasking capabilities of a LLM during the fine-tuning phase, we have aligned the initial domain distribution in the SFT stage with the knowledge detection results of the base model as stated in \Cref{eq:statement1}. Furthermore, we dynamically make minor adjustments in the composition ratios of various domains based on the model's real-time feedback at different SFT stages. 

As detailed in \Cref{alg:multi_ability}, in the pursuit of balanced domain expertise enhancement, we first blended the domain proportions in accordance with the base model's intrinsic domain knowledge distribution detected by \Cref{alg:domain_detection} (Line 1). Then at each training step $t$, we assigned a learnable potential score to each domain based on the methodology outlined in \cref{eq:loss_potential}. These scores were then utilized to fine-tune the representation of each domain within the composite SFT dataset, ensuring a balanced development of competencies across all domains throughout the training process (Line 3-7). 
At the same time, the parameters of model $M_{\theta}$ are updated based on the gradients computed through backpropagation (Line 8).
This adaptive approach is imperative to harmonize the progression of capabilities in different domains and to optimize the model's performance on multiple tasks. 

\subsubsection{Flexible Domain Expansion}
\label{subsec:flexible_domain_expansion}

When conducting fine-tuning on a pretrained model, there are instances where we aim to particularly enhance models' performance on specific domain tasks. Consequently, our algorithmic framework ought to possess the flexibility to accommodate domain expansion and generalize effectively. Building upon \hyperlink{statement1}{Statement 1}, we present the following statement tailored for domain expansion:

\paragraph{\textbf{Statement 2}} 
\label{para:statement2}
\textit{When fine-tuning a LLM for a specific capability, increasing the volume of data from a particular domain $D_e$ while adjusting other domains $(j = 1,2, ..., k, j \neq e)$ according to the knowledge distribution of the base model, facilitates a flexible strategy for domain expansion. Formally, the relationship can be represented as:}

\begin{equation}
\label{eq:statement2}
    P_{SFT}^{'}(x) \approx \sum_{j=1}^{k} A(D_j) P_{SFT}(x | D_j), j = 1, 2, ..., k
\end{equation}
\textit{where $P_{SFT}(x | D_j)$ is the data distribution in the domain $D_j$, and $A(D_j)$ is the adjustment factor.}

Here $A(D_j)$ is determined based on the knowledge distribution of the pre-trained domain. In particular, when $D_e$ increases, the other domains $\{D_j\}_{j = 1, j \neq e}^{k}$ shrink proportionally as a whole, which can be expressed as:

\begin{equation}
\label{eq:statement2_cases}
    A(D_j) =
    \begin{cases}  
        \alpha, \text{if} \quad j = e & \\  
        \beta \frac{1}{\sum_{j=1, j \neq e}^{k} A(D_j)} , \text{others} &
    \end{cases} 
\end{equation}
where $\alpha$ is the increased adjustment factor, and $\beta$ is the original ratio of other domain knowledge relative to $D_e$. 
Algorithm implementation details and hyper-parameter settings are provided in \Cref{subsec:appendix_domain_expansion} and \Cref{subsec:appendix_hyper_parameter}.

\section{Experiments and Results}
\label{sec:experiment}


In this section, we describe details of our experimental setup (\Cref{subsec:experimental_setup}), the baseline methods we use for comparison (\Cref{subsec:baselines}), and experimental results (\Cref{subsec:results}).

\begin{table*}[tb]
    \centering
    \resizebox{\textwidth}{!}{
        \begin{tabular}{cccccccccccccc}
            \toprule
                \multirow{2}{*}{Model} & \multirow{2}{*}{Method} & \multicolumn{2}{|c}{Law} & \multicolumn{2}{|c}{Medical} & \multicolumn{2}{|c}{Finance} & \multicolumn{2}{|c}{Science} & \multicolumn{2}{|c}{Code} & \multicolumn{2}{|c}{General} \\ 
            \cline{3-14}
             & & \multicolumn{1}{|c}{LegalBench} & LawBench & \multicolumn{1}{|c}{MedQA} & MedMCQA & \multicolumn{1}{|c}{FinEval} & FinanceIQ & \multicolumn{1}{|c}{SciEval} & MMLU-Sci & \multicolumn{1}{|c}{HumanEval} & MBPP & \multicolumn{1}{|c}{AGIEval} & \multicolumn{1}{c}{HellaSwag} \\
            \midrule
            \multicolumn{14}{c}{\cellcolor[HTML]{EFEFEF}\textbf{Frontier Models}} \\
            \midrule
            GPT-4o & -- & \multicolumn{1}{|c}{79.00} & 57.41 & \multicolumn{1}{|c}{81.92} & 74.60 & \multicolumn{1}{|c}{64.58} & 66.25 & \multicolumn{1}{|c}{72.54} & 85.47 & \multicolumn{1}{|c}{88.40} & 75.50 & \multicolumn{1}{|c}{71.82} & 90.56 \\
            Claude3.5-Sonnet & -- & \multicolumn{1}{|c}{77.60} & 40.73 & \multicolumn{1}{|c}{76.38} & 68.80 & \multicolumn{1}{|c}{65.90} & 62.55 & \multicolumn{1}{|c}{68.72} & 84.24 & \multicolumn{1}{|c}{84.07} & 80.48 & \multicolumn{1}{|c}{75.63} & 89.12 \\
            DeepSeek-V3 & -- & \multicolumn{1}{|c}{65.46} & 52.25 & \multicolumn{1}{|c}{78.82} & 74.30 & \multicolumn{1}{|c}{68.15} & 75.03 & \multicolumn{1}{|c}{69.58} & 82.90 & \multicolumn{1}{|c}{65.20} & 75.40 & \multicolumn{1}{|c}{79.60} & 88.90 \\
            \midrule
            \multicolumn{14}{c}{\cellcolor[HTML]{EFEFEF}\textbf{Open-Sourced Base Models}} \\
            \midrule
             \multirow{3}{*}{LLaMA-2-7B} & Uniform Distribution & \multicolumn{1}{|c}{15.71} & 30.72 & \multicolumn{1}{|c}{23.45} & 27.57 & \multicolumn{1}{|c}{33.50} & 2.71 & \multicolumn{1}{|c}{9.30} & 42.89 & \multicolumn{1}{|c}{5.67} & 3.44 & \multicolumn{1}{|c}{20.16} & 71.40 \\
              &  Inverse Distribution & \multicolumn{1}{|c}{13.23$^{\color{red} \downarrow}$} & 26.94$^{\color{red} \downarrow}$ & \multicolumn{1}{|c}{21.38$^{\color{red} \downarrow}$} & 26.52$^{\color{red} \downarrow }$ & \multicolumn{1}{|c}{32.96$^{\color{red} \downarrow }$} & 2.53$^{\color{red} \downarrow}$ & \multicolumn{1}{|c}{8.98$^{\color{red} \downarrow}$} & 39.67$^{\color{red} \downarrow }$ & \multicolumn{1}{|c}{3.47$^{\color{red} \downarrow}$} & 2.42$^{\color{red} \downarrow}$ & \multicolumn{1}{|c}{18.83$^{\color{red} \downarrow}$} & 71.33$^{\color{red} \downarrow }$ \\
              & \ours & \multicolumn{1}{|c}{\textbf{23.18}$^{\color{green} \uparrow}$} & \textbf{36.31}$^{\color{green} \uparrow}$ & \multicolumn{1}{|c}{\textbf{35.04}$^{\color{green} \uparrow }$} & \textbf{40.75}$^{\color{green} \uparrow}$ & \multicolumn{1}{|c}{\textbf{36.27}$^{\color{green} \uparrow}$} & \textbf{29.04}$^{\color{green} \uparrow}$ & \multicolumn{1}{|c}{\textbf{56.75}$^{\color{green} \uparrow}$} & \textbf{50.06}$^{\color{green} \uparrow}$ & \multicolumn{1}{|c}{\textbf{15.62}$^{\color{green} \uparrow}$} & \textbf{15.68}$^{\color{green} \uparrow}$ & \multicolumn{1}{|c}{\textbf{24.67}$^{\color{green} \uparrow}$} & 71.76$^{\color{green} \uparrow}$ \\
        \midrule
        \multirow{3}{*}{Qwen-2-7B} & Uniform Distribution & \multicolumn{1}{|c}{39.05} & 31.99 & \multicolumn{1}{|c}{35.07} & 17.73 & \multicolumn{1}{|c}{59.49} & 14.62 & \multicolumn{1}{|c}{25.30} & 62.73 & \multicolumn{1}{|c}{53.26} & 37.82 & \multicolumn{1}{|c}{47.31} & 73.60 \\
         &  Inverse Distribution & \multicolumn{1}{|c}{34.01$^{\color{red} \downarrow }$} & 27.81$^{\color{red} \downarrow }$ & \multicolumn{1}{|c}{23.90$^{\color{red} \downarrow }$} & 16.31$^{\color{red} \downarrow }$ & \multicolumn{1}{|c}{56.53$^{\color{red} \downarrow }$} & 11.30$^{\color{red} \downarrow }$ & \multicolumn{1}{|c}{18.57$^{\color{red} \downarrow }$} & 58.25$^{\color{red} \downarrow }$ & \multicolumn{1}{|c}{50.65$^{\color{red} \downarrow }$} & 33.63$^{\color{red} \downarrow }$ & \multicolumn{1}{|c}{45.74$^{\color{red} \downarrow }$} & 73.52$^{\color{red} \downarrow }$ \\
          & \ours & \multicolumn{1}{|c}{\textbf{50.56}$^{\color{green} \uparrow }$} & \textbf{35.54}$^{\color{green} \uparrow }$ & \multicolumn{1}{|c}{\textbf{45.48}$^{\color{green} \uparrow }$} & \textbf{41.24}$^{\color{green} \uparrow }$ & \multicolumn{1}{|c}{\textbf{60.95}$^{\color{green} \uparrow }$} & \textbf{68.39}$^{\color{green} \uparrow }$ & \multicolumn{1}{|c}{\textbf{51.58}$^{\color{green} \uparrow }$} & \textbf{70.42}$^{\color{green} \uparrow }$ & \multicolumn{1}{|c}{\textbf{58.15}$^{\color{green} \uparrow }$} & \textbf{47.64}$^{\color{green} \uparrow }$ & \multicolumn{1}{|c}{48.02$^{\color{green} \uparrow }$} & \textbf{73.67}$^{\color{green} \uparrow }$ \\
        \midrule
           \multirow{3}{*}{Qwen-2.5-7B} & Uniform Distribution & \multicolumn{1}{|c}{40.11} & 31.48 & \multicolumn{1}{|c}{25.17} & 25.84 & \multicolumn{1}{|c}{59.58} & 31.66 & \multicolumn{1}{|c}{19.88} & 65.84 & \multicolumn{1}{|c}{55.64} & 46.86 & \multicolumn{1}{|c}{45.42} & 73.69 \\
          &  Inverse Distribution & \multicolumn{1}{|c}{36.36$^{\color{red} \downarrow }$} & 26.98$^{\color{red} \downarrow }$  & \multicolumn{1}{|c}{24.16$^{\color{red} \downarrow }$} & 19.35$^{\color{red} \downarrow }$ & \multicolumn{1}{|c}{57.07$^{\color{red} \downarrow }$} & 29.25$^{\color{red} \downarrow }$ & \multicolumn{1}{|c}{16.68$^{\color{red} \downarrow }$} & 62.78$^{\color{red} \downarrow }$ & \multicolumn{1}{|c}{52.97$^{\color{red} \downarrow }$} & 44.63$^{\color{red} \downarrow }$ & \multicolumn{1}{|c}{45.67$^{\color{green} \uparrow }$} & 72.92$^{\color{red} \downarrow }$ \\
          & \ours & \multicolumn{1}{|c}{\textbf{51.65}$^{\color{green} \uparrow }$} & \textbf{36.75}$^{\color{green} \uparrow }$ & \multicolumn{1}{|c}{\textbf{34.28}$^{\color{green} \uparrow }$} & \textbf{52.09}$^{\color{green} \uparrow }$ & \multicolumn{1}{|c}{62.48$^{\color{green} \uparrow }$} & \textbf{69.09}$^{\color{green} \uparrow }$ & \multicolumn{1}{|c}{\textbf{68.14}$^{\color{green} \uparrow }$} & \textbf{74.16}$^{\color{green} \uparrow }$ & \multicolumn{1}{|c}{\textbf{60.68}$^{\color{green} \uparrow }$} & \textbf{61.25}$^{\color{green} \uparrow }$ & \multicolumn{1}{|c}{\textbf{49.73}$^{\color{green} \uparrow }$} & \textbf{73.90}$^{\color{green} \uparrow }$ \\
          \midrule
          \multirow{3}{*}{LLaMA-3-8B} & Uniform Distribution & \multicolumn{1}{|c}{33.52} & 31.16 & \multicolumn{1}{|c}{31.03} & 10.26 & \multicolumn{1}{|c}{34.83} & 4.97 & \multicolumn{1}{|c}{6.51} & 50.17 & \multicolumn{1}{|c}{22.94} & 28.85 & \multicolumn{1}{|c}{23.87} & 73.26 \\
        &  Inverse Distribution & \multicolumn{1}{|c}{27.83$^{\color{red} \downarrow }$} & 27.48$^{\color{red} \downarrow }$ & \multicolumn{1}{|c}{25.51$^{\color{red} \downarrow }$} & 8.77$^{\color{red} \downarrow }$ & \multicolumn{1}{|c}{33.71$^{\color{red} \downarrow }$} & 3.31$^{\color{red} \downarrow }$ & \multicolumn{1}{|c}{6.09$^{\color{red} \downarrow }$} & 46.62$^{\color{red} \downarrow }$ & \multicolumn{1}{|c}{19.67$^{\color{red} \downarrow }$} & 24.34$^{\color{red} \downarrow }$ & \multicolumn{1}{|c}{23.45$^{\color{red} \downarrow }$} & 72.40$^{\color{red} \downarrow }$ \\
        & \ours & \multicolumn{1}{|c}{\textbf{49.67}$^{\color{green} \uparrow }$} & \textbf{37.87}$^{\color{green} \uparrow }$ & \multicolumn{1}{|c}{\textbf{42.21}$^{\color{green} \uparrow }$} & \textbf{45.72}$^{\color{green} \uparrow }$ & \multicolumn{1}{|c}{\textbf{38.80}$^{\color{green} \uparrow }$} & \textbf{43.58}$^{\color{green} \uparrow }$ & \multicolumn{1}{|c}{\textbf{56.67}$^{\color{green} \uparrow }$} & \textbf{60.61}$^{\color{green} \uparrow }$ & \multicolumn{1}{|c}{28.91$^{\color{green} \uparrow }$} & \textbf{35.65}$^{\color{green} \uparrow }$ & \multicolumn{1}{|c}{\textbf{28.78}$^{\color{green} \uparrow }$} & \textbf{73.62}$^{\color{green} \uparrow }$ \\
        \midrule
           \multirow{3}{*}{LLaMA-2-13B} & Uniform Distribution & \multicolumn{1}{|c}{47.66} & 34.85 & \multicolumn{1}{|c}{32.98} & 36.54 & \multicolumn{1}{|c}{37.54} & 32.85 & \multicolumn{1}{|c}{45.72} & 50.77 & \multicolumn{1}{|c}{36.54} & 38.55 & \multicolumn{1}{|c}{36.89} & 73.50 \\
          &  Inverse Distribution & \multicolumn{1}{|c}{40.12$^{\color{red} \downarrow }$} & 30.67$^{\color{red} \downarrow }$ & \multicolumn{1}{|c}{26.27$^{\color{red} \downarrow }$} & 28.78$^{\color{red} \downarrow }$ & \multicolumn{1}{|c}{36.67$^{\color{red} \downarrow }$} & 26.76$^{\color{red} \downarrow }$ & \multicolumn{1}{|c}{38.96$^{\color{red} \downarrow }$} & 48.68$^{\color{red} \downarrow }$ & \multicolumn{1}{|c}{28.78$^{\color{red} \downarrow }$} & 35.83$^{\color{red} \downarrow }$ & \multicolumn{1}{|c}{36.67$^{\color{red} \downarrow }$} & 73.11$^{\color{red} \downarrow }$ \\
          & \ours & \multicolumn{1}{|c}{\textbf{55.87}$^{\color{green} \uparrow }$} & \textbf{40.14}$^{\color{green} \uparrow }$ & \multicolumn{1}{|c}{\textbf{45.78}$^{\color{green} \uparrow }$} & \textbf{47.67}$^{\color{green} \uparrow }$ & \multicolumn{1}{|c}{\textbf{39.48}$^{\color{green} \uparrow }$} & \textbf{55.12}$^{\color{green} \uparrow }$ & \multicolumn{1}{|c}{\textbf{63.87}$^{\color{green} \uparrow }$} & \textbf{62.84}$^{\color{green} \uparrow }$ & \multicolumn{1}{|c}{\textbf{47.67}$^{\color{green} \uparrow }$} & \textbf{44.62}$^{\color{green} \uparrow }$ & \multicolumn{1}{|c}{\textbf{39.64}$^{\color{green} \uparrow }$} & \textbf{74.63}$^{\color{green} \uparrow }$ \\
          \midrule
           \multirow{3}{*}{Qwen-2.5-14B} & Uniform Distribution & \multicolumn{1}{|c}{50.73} & 39.49 & \multicolumn{1}{|c}{47.85} & 38.71 & \multicolumn{1}{|c}{64.72} & 64.39 & \multicolumn{1}{|c}{39.74} & 73.45 & \multicolumn{1}{|c}{68.75} & 72.14 & \multicolumn{1}{|c}{54.92} & 75.88 \\
          &  Inverse Distribution & \multicolumn{1}{|c}{46.08$^{\color{red} \downarrow }$} & 35.36$^{\color{red} \downarrow }$ & \multicolumn{1}{|c}{45.75$^{\color{red} \downarrow }$} & 32.56$^{\color{red} \downarrow }$ & \multicolumn{1}{|c}{64.88$^{\color{green} \uparrow }$} & 60.53$^{\color{red} \downarrow }$ & \multicolumn{1}{|c}{27.68$^{\color{red} \downarrow }$} & 68.22$^{\color{red} \downarrow }$ & \multicolumn{1}{|c}{63.36$^{\color{red} \downarrow }$} & 68.49$^{\color{red} \downarrow }$ & \multicolumn{1}{|c}{54.87$^{\color{red} \downarrow }$} & 75.42$^{\color{red} \downarrow }$ \\
          & \ours & \multicolumn{1}{|c}{\textbf{60.59}$^{\color{green} \uparrow }$} & \textbf{46.58}$^{\color{green} \uparrow }$ & \multicolumn{1}{|c}{\textbf{50.24}$^{\color{green} \uparrow }$} & \textbf{45.15}$^{\color{green} \uparrow }$ & \multicolumn{1}{|c}{\textbf{65.84}$^{\color{green} \uparrow }$} & \textbf{78.68}$^{\color{green} \uparrow }$ & \multicolumn{1}{|c}{\textbf{62.89}$^{\color{green} \uparrow }$} & \textbf{82.86}$^{\color{green} \uparrow }$ & \multicolumn{1}{|c}{\textbf{82.64}$^{\color{green} \uparrow }$} & \textbf{81.48}$^{\color{green} \uparrow }$ & \multicolumn{1}{|c}{\textbf{55.52}$^{\color{green} \uparrow }$} & 75.98$^{\color{green} \uparrow }$ \\
          \midrule
           \multirow{3}{*}{Qwen-2.5-32B} & Uniform Distribution & \multicolumn{1}{|c}{68.86} & 45.28 & \multicolumn{1}{|c}{72.34} & 68.18 & \multicolumn{1}{|c}{68.03} & 75.14 & \multicolumn{1}{|c}{58.30} & 80.17 & \multicolumn{1}{|c}{78.59} & 71.04 & \multicolumn{1}{|c}{75.26} & 84.40 \\
          &  Inverse Distribution & \multicolumn{1}{|c}{62.93$^{\color{red} \downarrow }$} & 42.05$^{\color{red} \downarrow }$ & \multicolumn{1}{|c}{68.80$^{\color{red} \downarrow }$} & 66.09$^{\color{red} \downarrow }$ & \multicolumn{1}{|c}{66.80$^{\color{red} \downarrow }$} & 73.93$^{\color{red} \downarrow }$ & \multicolumn{1}{|c}{52.94$^{\color{red} \downarrow }$} & 79.31$^{\color{red} \downarrow }$ & \multicolumn{1}{|c}{74.44$^{\color{red} \downarrow }$} & 70.71$^{\color{red} \downarrow }$ & \multicolumn{1}{|c}{75.00$^{\color{red} \downarrow }$} & 83.80$^{\color{red} \downarrow }$ \\
          & \cellcolor[HTML]{FFFDC6} \ours & \multicolumn{1}{|c}{\cellcolor[HTML]{FFFDC6}\textbf{75.67}$^{\color{green} \uparrow }$} & \cellcolor[HTML]{FFFDC6}\textbf{56.76}$^{\color{green} \uparrow }$ & \multicolumn{1}{|c}{\cellcolor[HTML]{FFFDC6}\textbf{78.72}$^{\color{green} \uparrow }$} & \cellcolor[HTML]{FFFDC6}\textbf{72.36}$^{\color{green} \uparrow }$ & \multicolumn{1}{|c}{\cellcolor[HTML]{FFFDC6}\textbf{70.50}$^{\color{green} \uparrow }$} & \cellcolor[HTML]{FFFDC6}\textbf{78.80}$^{\color{green} \uparrow }$ & \multicolumn{1}{|c}{\cellcolor[HTML]{FFFDC6}\textbf{70.77}$^{\color{green} \uparrow }$} & \cellcolor[HTML]{FFFDC6}\textbf{85.23}$^{\color{green} \uparrow }$ & \multicolumn{1}{|c}{\cellcolor[HTML]{FFFDC6}\textbf{86.60}$^{\color{green} \uparrow }$} & \cellcolor[HTML]{FFFDC6}\textbf{79.89}$^{\color{green} \uparrow }$ & \multicolumn{1}{|c}{\cellcolor[HTML]{FFFDC6}\textbf{75.81}$^{\color{green} \uparrow }$} & \cellcolor[HTML]{FFFDC6}\textbf{84.75}$^{\color{green} \uparrow }$ \\
        \bottomrule
        \end{tabular}
    }
    \caption{\label{tab:multi_ability_total}
    Results of \ours on multi-ability fostering, we compare the performances of several methods across different models. For each domain, we evaluate the models using two relevant benchmarks. The best results are in \textbf{bold}. ${\color{green} \uparrow}$ and ${\color{red} \downarrow}$ indicate an increase or decrease in downstream scores comparing to the \textit{uniform distribution} strategy.
} 
\vskip -0.2in
\end{table*}

\subsection{Experimental Setup}
\label{subsec:experimental_setup}

\textbf{Datasets.} 
For \textit{training}, we have collected datasets spanning 6 domains for supervised fine-tuning, including Lawyer-Instruct\footnote{https://huggingface.co/datasets/Alignment-Lab-AI/Lawyer-Instruct}, the training portion of MedQA~\cite{jin2020disease}, Finance Alpaca\footnote{https://huggingface.co/datasets/gbharti/finance-alpaca}, Sonnet3.5 Science Conversations\footnote{https://huggingface.co/datasets/jeffmeloy/sonnet3.5\_science\_conversations}, Code Alpaca\footnote{https://github.com/sahil280114/codealpaca} and Alpaca~\cite{taori2023stanford}, denoted as $\mathcal{D}_{train} = \{ (D_{train}^j, P_j)\}_{j=1}^{6}$, to represent SFT datasets with respect to law, medicine, finance, science, code as well as general capabilities. 
In order to prevent domain overlap, we curated the Alpaca dataset by excluding data pertaining to the other specific five domains, keeping only the general domain instances unrelated to them. 
More details can be found in \Cref{sec:appendix_training}. 
For \textit{evaluation}, we assess the model performances on downstream tasks across various domains, using two relevant benchmarks for each domain, with details provided in \Cref{sec:appendix_evaluation}.

\textbf{Models and Implementation.} 
We employ LLaMA~\cite{dubey2024llama,touvron2023llama,touvron2023llama2} and Qwen~\cite{bai2023qwen,yang2024qwen2} series as our pretrained language models $M_\theta$. 
During the fine-tuning procedure, we utilized a learning rate scheduler featuring linear warm-up and cosine decay, peaking at a learning rate of 2e\mbox{-}5, alongside a warmup ratio of 0.03, a weight decay of 0.0 and a batch size of 128 for 4 epochs.
To maintain consistency, the total volume of training data across domains was controlled to 60,000 per epoch.
We conducted all fine-tuning and evaluation experiments on NVIDIA RTX H800. 
Details of the experimental settings can be found in \Cref{sec:appendix_experimental_detail}.

\subsection{Baselines}
\label{subsec:baselines}

We compare \ours with the following baselines. For the scenario of 
\textit{effective multi-ability fostering}:
(1) The simplest baseline is \textbf{uniform distribution}, where each domain has an equal weight proportion. 
(2) \textbf{Inverse distribution} assigns the proportionate weights to each domain in an inverse manner to the detected knowledge distribution. 
(3) \textbf{Frontier models} contain 
GPT-4o~\cite{hurst2024gpt}, 
Claude3.5-Sonnet~\cite{claude_anthropic_2024} and DeepSeek-V3~\cite{liu2024deepseek}. 
Under the case of 
\textit{flexible domain expansion}: 
(1) \textbf{100\% specific domain} strategy is a common practice to employ datasets consisting exclusively of data from a single domain during the fine-tuning stage. 
(2) \textbf{Domain increase with uniform distribution of remainder} elevates the proportion of a specific domain, while the remaining domains receive the balance of the distribution in an evenly distributed manner.

\subsection{Results}
\label{subsec:results}

We conduct evaluations to validate the efficiency of \ours across different open-source models in scenarios that encompass both effective multi-ability fostering and flexible domain expansion.
We summarize the observations below.

\textbf{\ours is efficient across different models in both scenarios.} 
For the scenario of multi-capability fostering, 
\Cref{tab:multi_ability_total} 
shows that \ours consistently outperforms other baseline methods across different models in terms of domain-specific capabilities. Compared to the \textit{uniform distribution} of data across domains, \ours enhances downstream task performances by $35.21\%$, which further underscores the effectiveness of our data composition strategy for enhancing the model's overall multi-domain capabilities during the supervised fine-tuning phase. 
Moreover, \colorbox[HTML]{FFFDC6}{\textbf{Qwen-2.5-32B + VersaTune}} has the potential to surpass frontier models under medical scenarios, achieving average improvements over GPT-4o, Claude3.5-Sonnet and DeepSeek-V3 by 0.86\%, 4.76\% and 4.60\%. 
Since we have not conducted domain-specific refinement for domains outside the current five specific domains, the models' performance gains on general benchmarks are not as noticeable. 
For domain expansion scenarios, \ours has nearly maintained training efficiency while reducing the model's loss of competencies in other domains by $38.77\%$ comparing to \textit{100\% specific domain fine-tuning}, as depicted in \Cref{tab:flexible_domain_expansion}, 
where we averaged the experimental results from Qwen-2.5-7B and Qwen-2.5-14B. 
Detailed results and analysis can be found in \Cref{sec:appendix_experiment_result}.

\textbf{Knowledge consistency training boosts performance.} 
In \Cref{tab:multi_ability_total}, we present the experimental results of data composition strategies that allocate domain data in a manner inversely proportional to the pre-existing knowledge distribution detected within each domain. 
As expected, the \textit{inverse distribution} strategy yielded even lower$\color{red} \downarrow $ performance compared to the simplest approach of \textit{uniform distribution}, which evenly distributes data across all domains. 
This finding underscores the importance of aligning domain data ratios with the inherent knowledge distribution of the model during training, which proves the efficacy of knowledge consistency training stated in \Cref{subsec:knowledge_consistency_training}.

\section{Ablations and Analysis}
\label{sec:ablation_study}

Previously in \Cref{sec:experiment}, we have demonstrated the effectiveness of \ours in enhancing multiple abilities and enabling flexible domain expansion during the SFT phase. 
In this section, we perform an in-depth analysis of VersaTune, where we ablate the components of \textit{\textbf{\underline{(1)}}} dynamic adaptation in \Cref{alg:multi_ability}, and \textit{\textbf{\underline{(2)}}} the criteria for determining the upper limit of domain expansion in \Cref{alg:domain_expansion}.

\begin{figure}[ht]
    \begin{center}
    \centerline{\includegraphics[width=\linewidth]{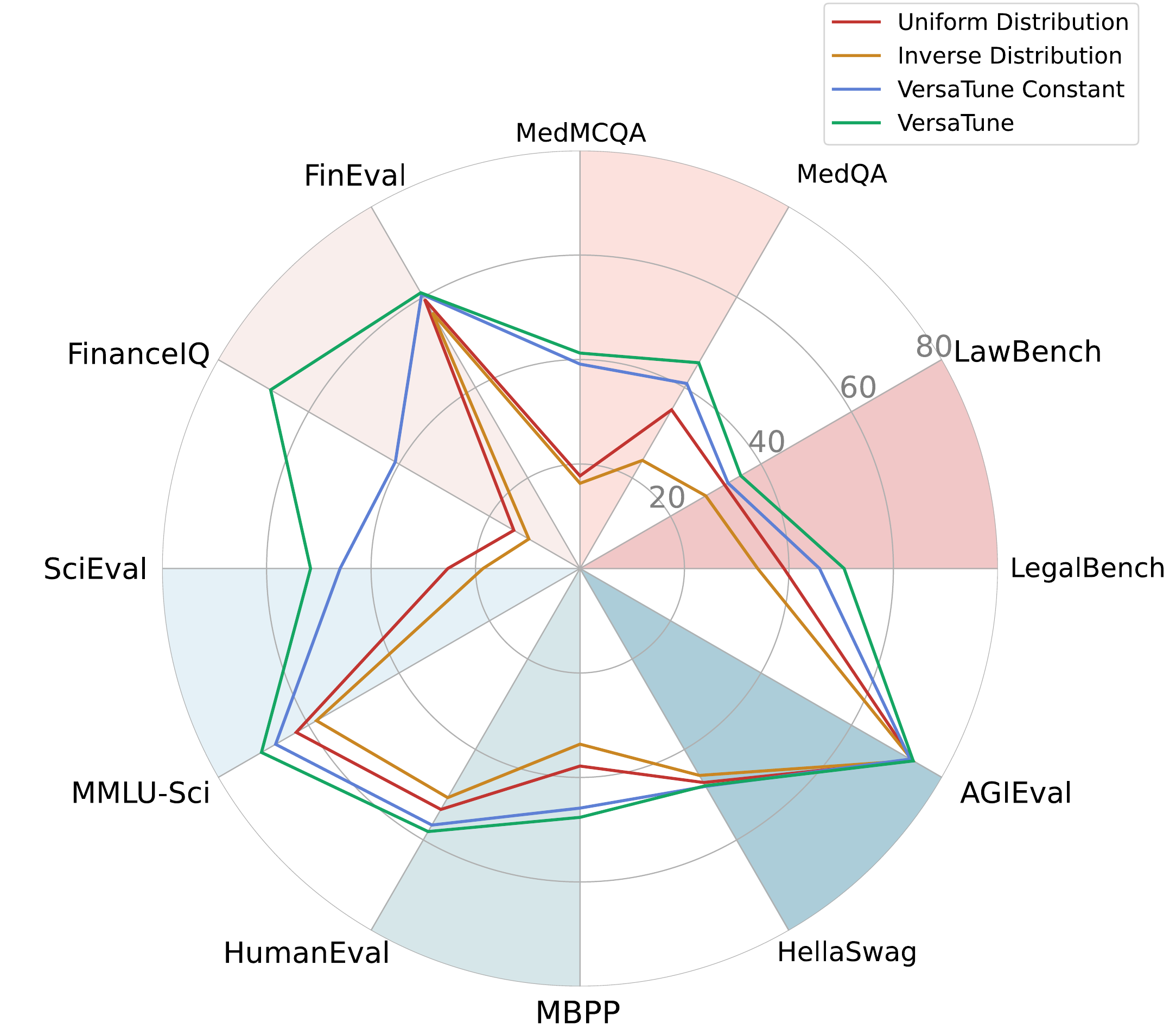}}
    \caption{Performances of \textbf{\textit{Qwen-2-7B}} on versatile tasks across different domains for multi-ability fostering. 
    } 
    \label{fig:qwen2-7b_multi_ability}
    \end{center}
    \vskip -0.4in
\end{figure}


\textbf{Dynamic adjustment enhances the robustness.} 
During the process of cultivating multiple capabilities, we compared \ours with \textit{fixed domain weights} referring to the knowledge distribution obtained from probing the target model $M_\theta$ prior to supervised fine-tuning, namely \textit{\textbf{\ours Constant}}, to ablate the component of dynamic adaptation in \Cref{alg:multi_ability}. 
\Cref{tab:multi_ability_total_appendix}, \Cref{fig:qwen2-7b_multi_ability}, and \Cref{fig:domain_expansion_total_appendix} demonstrated the high robustness of VersaTune, which dynamically adjusts domain weights throughout the training process by continuously monitoring the learnable potential within each domain. 
In contrast, training with fixed domain weights exhibits certain fluctuations. A key reason for this phenomenon is the distribution of domain knowledge mastered by the model changes during training, and the learning efficiency varies among domains. Therefore, dynamically adjusting domain data weights based on the model's feedback at different stages of training is crucial. 
More experimental results can be found in \Cref{subsec:appendix_multi_ability_results}.

\begin{figure*}[ht]
    \begin{center}
    \centerline{\includegraphics[width=\linewidth]{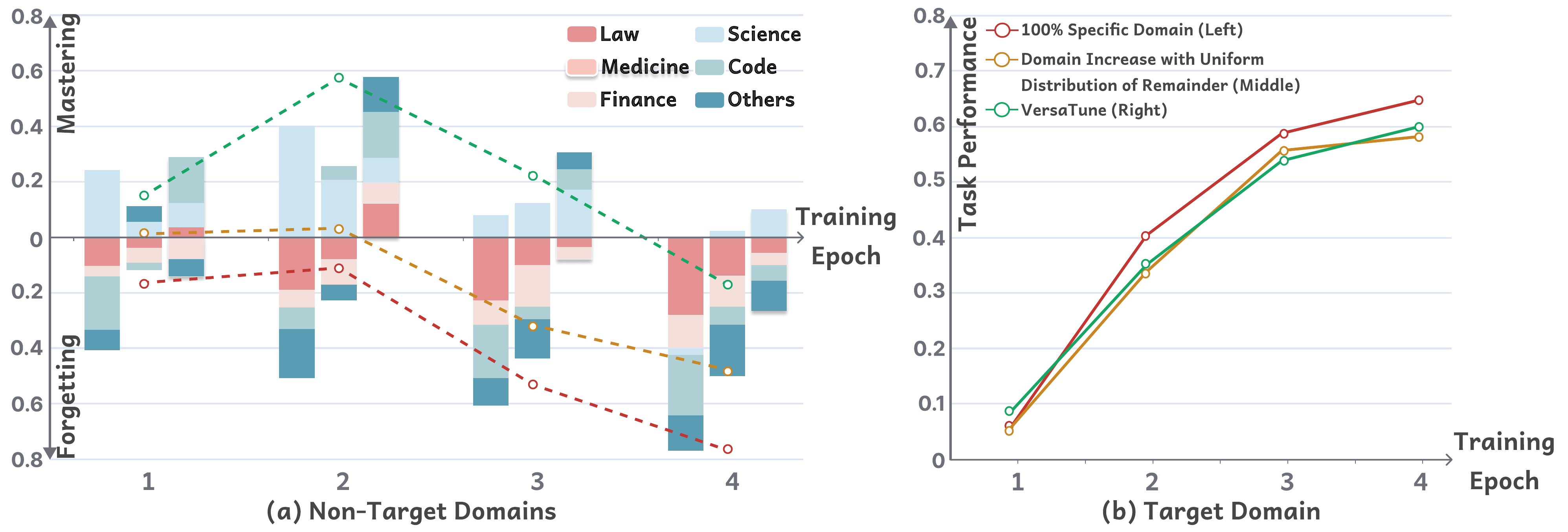}}
    \caption{
    Domain expansion for \textit{medicine} domain. 
    We evaluated checkpoints from each epoch. 
    \textbf{Left (a)} presents the grouped stacked bar chart showing the growth or loss of capabilities in non-target domains compared to the pre-fine-tuning state. Within each group, the left, center, and right bars represent: (1) 100\% specific domain fine-tuning, (2) domain increase with uniform distribution of remainder, and (3) \ours implementation based on \Cref{alg:domain_expansion}. \textbf{Right (b)} features the line chart depicting the enhancement of the medicine domain's capabilities.}
    \label{fig:domain_expansion_medicine}
    \end{center}
    \vskip -0.4in
\end{figure*}

\begin{figure}[ht]
    \begin{center}
    \centerline{\includegraphics[width=\linewidth]{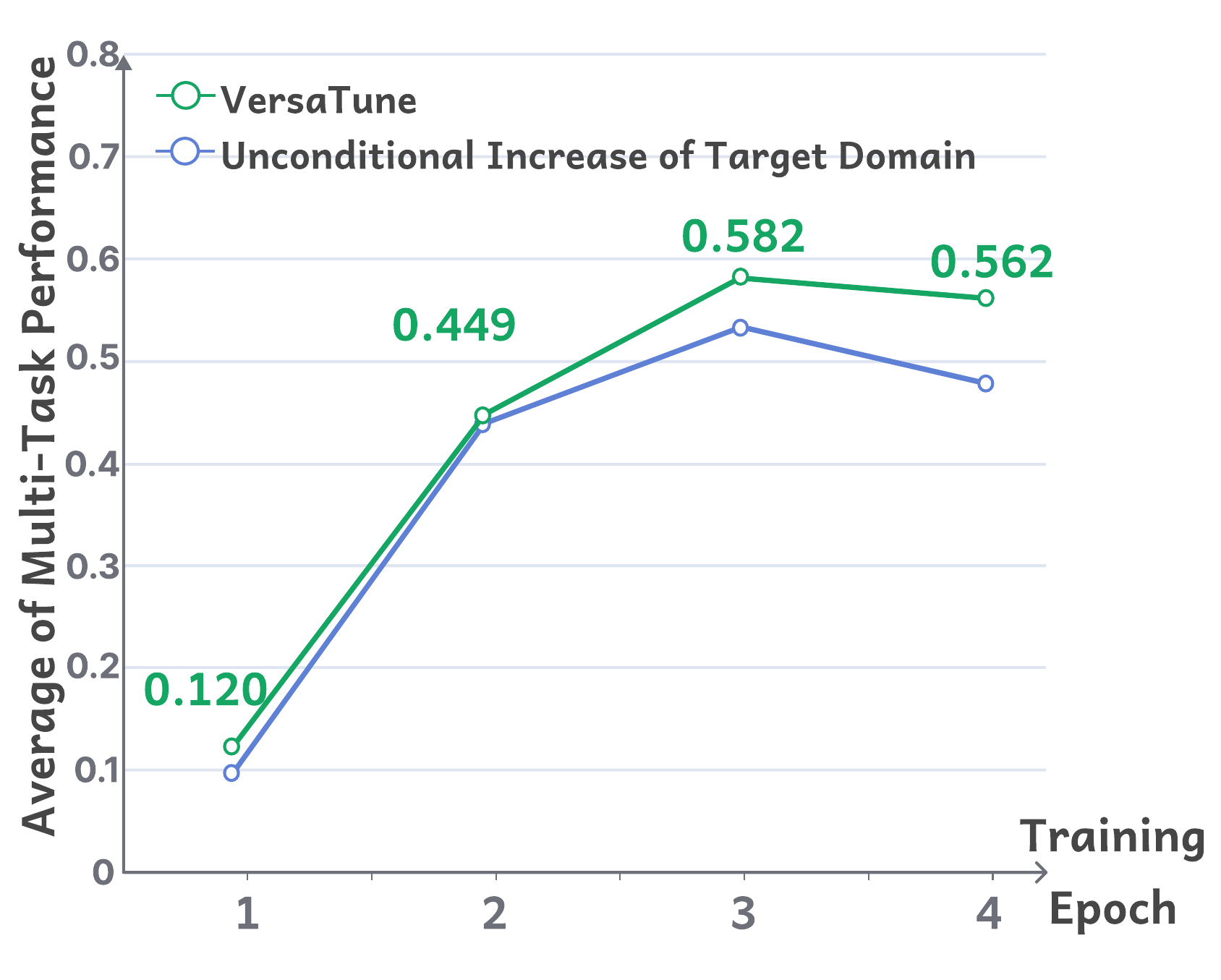}}
    \caption{The average scores of models' performances across domains during the domain expansion process, with detailed domain variations provided in \Cref{fig:domain_expansion_ablation}.}
    \label{fig:domain_expansion_ablation_avg}
    \end{center}
    \vskip -0.4in
\end{figure}

\textbf{Establishing proportion thresholds for specific domains counts during domain expansion.}
We consider conducting a comparative analysis between the outcomes of \ours and those implementing an \textit{unconditional dynamic increase of the specific domain}, where we remove the implementation of Line 8 in \Cref{alg:domain_expansion}. 
\Cref{fig:domain_expansion_ablation_avg}
shows that the criteria for determining the upper limit on the proportion of a specific domain during domain expansion, 
has mitigated the loss of capabilities in other domains experienced by the target model $M_\theta$ during the fine-tuning process. Concurrently, it ensures gains in the capacity for the current domain of interest. 
We speculate that it is because by the later stages of fine-tuning, models' proficiency in the target domain approaches saturation. Further increasing the proportion of the current domain provides diminishing returns and can lead to a significant loss of performance in other domains. 
Detailed analysis are provided in \Cref{subsubsec:appendix_proportion_threshold}.

\section{Related Work}
\label{sec:related_work}

\textbf{Data Reweighting for LLM Training.}
Data reweighting maintains full access to the entire dataset while adjusts the relative importance of each instance for various target tasks, which is essential for both pretraining and fine-tuning stages of LLMs~\cite{wang2023data}. 
During the pretraining stage, DoReMi~\cite{xie2024doremi} and DoGE~\cite{fan2023doge} employ lightweight proxy models to estimate weights for different data sources, which are subsequently applied to the formal training of LLMs. 
Furthermore, Sheared LLaMA~\cite{xia2023sheared} implements an online variant of DoReMi. 
As for the SFT phase, Dong et al.~\cite{dong2023abilities} focus on enhancing the model's math reasoning, coding, and human-aligning abilities through a dual-stage mixed fine-tuning strategy. However, the mixing ratios for different domains rely heavily on empirical methods, and the covered domains are not holistic. 
We provide a comprehensive overview of the model's capabilities across domains during the SFT stage and proposes appropriate and holistic multi-ability fine-tuning methods.

\textbf{Knowledge Detection in LLMs.} 
Investigating the knowledge contained in LLMs is essential for guiding their subsequent training~\cite{chang2024survey}. The knowledge encompasses multiple dimensions, such as different domain sources and task attributes. 
Existing work on LLM knowledge detection primarily focuses on prompting and calibration. Directly prompting the model to generate sequences and extracting confidence scores from the model~\cite{gekhman2024does,kadavath2022language,kuhn2023semantic,manakul2023selfcheckgpt} is a common strategy. However, such approaches highly depend on prompt design and task selection, introducing bias into the assessment. 
Other studies have attempted to infer the training data mixtures used in previous training stages~\cite{antoniades2024generalization,hayase2024data,hu2022membership,ye2022enhanced}. The essence of these studies is to evaluate the current knowledge state of the models and provide targeted strategies for data organization and management in subsequent training phases.


\section{Conclusion}
\label{sec:conclusion}

In this work, we introduce VersaTune, a noval data composition framework designed to enhance the multi-domain capabilities of models during the supervised fine-tuning phase of LLMs, which is based on the domain knowledge distribution of the target model. 
Experimental results have
demonstrated that \ours achieves excellent training outcomes in both scenarios of overall multi-domain enhancement and flexible domain expansion.

\section{Limitations}
\label{sec:limitations}

There are some limitations in our work. 
Firstly, the classification framework of vertical domains may not be comprehensive in scope, and since the classifier relies on advanced language models, it cannot guarantee absolute accuracy in classification. 
Additionally, when computing the learnable potential and forgetting degree of knowledge, to balance the computational cost and effectiveness, we employ a lightweight proxy model to for calculation, yet it does not fully represent the performance tendencies of the target model during actual evaluating. 

\section{Ethical Considerations}
\label{sec:ethical_considerations}


Integrating Large Language Models (LLMs) into domain reweighting settings holds potential for improving multi-domain capabilities of models, while it also brings several ethical considerations that must be addressed to ensure responsible and beneficial use.
\ours dynamically adjusts data distribution based on the model's existing knowledge to ensure fairness and avoid biases that could arise from the data composition process. 
Additionally, \ours adhere to privacy standards by merely utilizing open-sourced datasets, ensuring that personal data used in the training process is anonymized and securely handled to protect individual privacy rights.


\bibliography{versatune2025}

\begin{thebibliography}{68}
\providecommand{\natexlab}[1]{#1}

\bibitem[{Achiam et~al.(2023)Achiam, Adler, Agarwal, Ahmad, Akkaya, Aleman, Almeida, Altenschmidt, Altman, Anadkat et~al.}]{achiam2023gpt}
Josh Achiam, Steven Adler, Sandhini Agarwal, Lama Ahmad, Ilge Akkaya, Florencia~Leoni Aleman, Diogo Almeida, Janko Altenschmidt, Sam Altman, Shyamal Anadkat, and 1 others. 2023.
\newblock Gpt-4 technical report.
\newblock \emph{arXiv preprint arXiv:2303.08774}.

\bibitem[{Albalak et~al.(2024)Albalak, Elazar, Xie, Longpre, Lambert, Wang, Muennighoff, Hou, Pan, Jeong et~al.}]{albalak2024survey}
Alon Albalak, Yanai Elazar, Sang~Michael Xie, Shayne Longpre, Nathan Lambert, Xinyi Wang, Niklas Muennighoff, Bairu Hou, Liangming Pan, Haewon Jeong, and 1 others. 2024.
\newblock A survey on data selection for language models.
\newblock \emph{arXiv preprint arXiv:2402.16827}.

\bibitem[{Anthropic(2024)}]{claude_anthropic_2024}
Anthropic. 2024.
\newblock \href {https://www.anthropic.com/} {Claude}.
\newblock Accessed: 2024-06-27.

\bibitem[{Antoniades et~al.(2024)Antoniades, Wang, Elazar, Amayuelas, Albalak, Zhang, and Wang}]{antoniades2024generalization}
Antonis Antoniades, Xinyi Wang, Yanai Elazar, Alfonso Amayuelas, Alon Albalak, Kexun Zhang, and William~Yang Wang. 2024.
\newblock Generalization vs memorization: Tracing language models' capabilities back to pretraining data.
\newblock \emph{arXiv preprint arXiv:2407.14985}.

\bibitem[{Austin et~al.(2021)Austin, Odena, Nye, Bosma, Michalewski, Dohan, Jiang, Cai, Terry, Le et~al.}]{austin2021program}
Jacob Austin, Augustus Odena, Maxwell Nye, Maarten Bosma, Henryk Michalewski, David Dohan, Ellen Jiang, Carrie Cai, Michael Terry, Quoc Le, and 1 others. 2021.
\newblock Program synthesis with large language models.
\newblock \emph{arXiv preprint arXiv:2108.07732}.

\bibitem[{Azerbayev et~al.(2023)Azerbayev, Schoelkopf, Paster, Santos, McAleer, Jiang, Deng, Biderman, and Welleck}]{azerbayev2023llemma}
Zhangir Azerbayev, Hailey Schoelkopf, Keiran Paster, Marco~Dos Santos, Stephen McAleer, Albert~Q Jiang, Jia Deng, Stella Biderman, and Sean Welleck. 2023.
\newblock Llemma: An open language model for mathematics.
\newblock \emph{arXiv preprint arXiv:2310.10631}.

\bibitem[{Bai et~al.(2023)Bai, Bai, Chu, Cui, Dang, Deng, Fan, Ge, Han, Huang et~al.}]{bai2023qwen}
Jinze Bai, Shuai Bai, Yunfei Chu, Zeyu Cui, Kai Dang, Xiaodong Deng, Yang Fan, Wenbin Ge, Yu~Han, Fei Huang, and 1 others. 2023.
\newblock Qwen technical report.
\newblock \emph{arXiv preprint arXiv:2309.16609}.

\bibitem[{Beltagy et~al.(2019)Beltagy, Lo, and Cohan}]{beltagy2019scibert}
Iz~Beltagy, Kyle Lo, and Arman Cohan. 2019.
\newblock Scibert: A pretrained language model for scientific text.
\newblock \emph{arXiv preprint arXiv:1903.10676}.

\bibitem[{Brown et~al.(2020)Brown, Mann, Ryder, Subbiah, Kaplan, Dhariwal, Neelakantan, Shyam, Sastry, Askell et~al.}]{brown2020language}
Tom Brown, Benjamin Mann, Nick Ryder, Melanie Subbiah, Jared~D Kaplan, Prafulla Dhariwal, Arvind Neelakantan, Pranav Shyam, Girish Sastry, Amanda Askell, and 1 others. 2020.
\newblock Language models are few-shot learners.
\newblock \emph{Advances in neural information processing systems}, 33:1877--1901.

\bibitem[{Chang et~al.(2024)Chang, Wang, Wang, Wu, Yang, Zhu, Chen, Yi, Wang, Wang et~al.}]{chang2024survey}
Yupeng Chang, Xu~Wang, Jindong Wang, Yuan Wu, Linyi Yang, Kaijie Zhu, Hao Chen, Xiaoyuan Yi, Cunxiang Wang, Yidong Wang, and 1 others. 2024.
\newblock A survey on evaluation of large language models.
\newblock \emph{ACM Transactions on Intelligent Systems and Technology}, 15(3):1--45.

\bibitem[{Chen et~al.(2021)Chen, Tworek, Jun, Yuan, Pinto, Kaplan, Edwards, Burda, Joseph, Brockman et~al.}]{chen2021evaluating}
Mark Chen, Jerry Tworek, Heewoo Jun, Qiming Yuan, Henrique Ponde De~Oliveira Pinto, Jared Kaplan, Harri Edwards, Yuri Burda, Nicholas Joseph, Greg Brockman, and 1 others. 2021.
\newblock Evaluating large language models trained on code.
\newblock \emph{arXiv preprint arXiv:2107.03374}.

\bibitem[{Cui et~al.(2023)Cui, Li, Yan, Chen, and Yuan}]{cui2023chatlaw}
Jiaxi Cui, Zongjian Li, Yang Yan, Bohua Chen, and Li~Yuan. 2023.
\newblock Chatlaw: Open-source legal large language model with integrated external knowledge bases.
\newblock \emph{arXiv preprint arXiv:2306.16092}.

\bibitem[{De~Lange et~al.(2021)De~Lange, Aljundi, Masana, Parisot, Jia, Leonardis, Slabaugh, and Tuytelaars}]{de2021continual}
Matthias De~Lange, Rahaf Aljundi, Marc Masana, Sarah Parisot, Xu~Jia, Ale{\v{s}} Leonardis, Gregory Slabaugh, and Tinne Tuytelaars. 2021.
\newblock A continual learning survey: Defying forgetting in classification tasks.
\newblock \emph{IEEE transactions on pattern analysis and machine intelligence}, 44(7):3366--3385.

\bibitem[{Devlin(2018)}]{devlin2018bert}
Jacob Devlin. 2018.
\newblock Bert: Pre-training of deep bidirectional transformers for language understanding.
\newblock \emph{arXiv preprint arXiv:1810.04805}.

\bibitem[{Ding et~al.(2022)Ding, Xu, Tang, Xu, Wang, and Tao}]{CVPR_distribution_estimation}
Ning Ding, Yixing Xu, Yehui Tang, Chao Xu, Yunhe Wang, and Dacheng Tao. 2022.
\newblock \href {https://doi.org/10.1109/CVPR52688.2022.00707} {Source-free domain adaptation via distribution estimation}.
\newblock In \emph{2022 IEEE/CVF Conference on Computer Vision and Pattern Recognition (CVPR)}, pages 7202--7212.

\bibitem[{Dong et~al.(2023)Dong, Yuan, Lu, Li, Xue, Liu, Wang, Yuan, Zhou, and Zhou}]{dong2023abilities}
Guanting Dong, Hongyi Yuan, Keming Lu, Chengpeng Li, Mingfeng Xue, Dayiheng Liu, Wei Wang, Zheng Yuan, Chang Zhou, and Jingren Zhou. 2023.
\newblock How abilities in large language models are affected by supervised fine-tuning data composition.
\newblock \emph{arXiv preprint arXiv:2310.05492}.

\bibitem[{Dubey et~al.(2024)Dubey, Jauhri, Pandey, Kadian, Al-Dahle, Letman, Mathur, Schelten, Yang, Fan et~al.}]{dubey2024llama}
Abhimanyu Dubey, Abhinav Jauhri, Abhinav Pandey, Abhishek Kadian, Ahmad Al-Dahle, Aiesha Letman, Akhil Mathur, Alan Schelten, Amy Yang, Angela Fan, and 1 others. 2024.
\newblock The llama 3 herd of models.
\newblock \emph{arXiv preprint arXiv:2407.21783}.

\bibitem[{Dwivedi et~al.(2021)Dwivedi, Hughes, Ismagilova, Aarts, Coombs, Crick, Duan, Dwivedi, Edwards, Eirug et~al.}]{dwivedi2021artificial}
Yogesh~K Dwivedi, Laurie Hughes, Elvira Ismagilova, Gert Aarts, Crispin Coombs, Tom Crick, Yanqing Duan, Rohita Dwivedi, John Edwards, Aled Eirug, and 1 others. 2021.
\newblock Artificial intelligence (ai): Multidisciplinary perspectives on emerging challenges, opportunities, and agenda for research, practice and policy.
\newblock \emph{International journal of information management}, 57:101994.

\bibitem[{Fan et~al.(2023)Fan, Pagliardini, and Jaggi}]{fan2023doge}
Simin Fan, Matteo Pagliardini, and Martin Jaggi. 2023.
\newblock Doge: Domain reweighting with generalization estimation.
\newblock \emph{arXiv preprint arXiv:2310.15393}.

\bibitem[{Fei et~al.(2023)Fei, Shen, Zhu, Zhou, Han, Zhang, Chen, Shen, and Ge}]{fei2023lawbench}
Zhiwei Fei, Xiaoyu Shen, Dawei Zhu, Fengzhe Zhou, Zhuo Han, Songyang Zhang, Kai Chen, Zongwen Shen, and Jidong Ge. 2023.
\newblock Lawbench: Benchmarking legal knowledge of large language models.
\newblock \emph{arXiv preprint arXiv:2309.16289}.

\bibitem[{Gekhman et~al.(2024)Gekhman, Yona, Aharoni, Eyal, Feder, Reichart, and Herzig}]{gekhman2024does}
Zorik Gekhman, Gal Yona, Roee Aharoni, Matan Eyal, Amir Feder, Roi Reichart, and Jonathan Herzig. 2024.
\newblock Does fine-tuning llms on new knowledge encourage hallucinations?
\newblock \emph{arXiv preprint arXiv:2405.05904}.

\bibitem[{Guha et~al.(2024)Guha, Nyarko, Ho, R{\'e}, Chilton, Chohlas-Wood, Peters, Waldon, Rockmore, Zambrano et~al.}]{guha2024legalbench}
Neel Guha, Julian Nyarko, Daniel Ho, Christopher R{\'e}, Adam Chilton, Alex Chohlas-Wood, Austin Peters, Brandon Waldon, Daniel Rockmore, Diego Zambrano, and 1 others. 2024.
\newblock Legalbench: A collaboratively built benchmark for measuring legal reasoning in large language models.
\newblock \emph{Advances in Neural Information Processing Systems}, 36.

\bibitem[{Hayase et~al.(2024)Hayase, Liu, Choi, Oh, and Smith}]{hayase2024data}
Jonathan Hayase, Alisa Liu, Yejin Choi, Sewoong Oh, and Noah~A Smith. 2024.
\newblock Data mixture inference: What do bpe tokenizers reveal about their training data?
\newblock \emph{arXiv preprint arXiv:2407.16607}.

\bibitem[{Hendrycks et~al.(2020)Hendrycks, Burns, Basart, Zou, Mazeika, Song, and Steinhardt}]{hendrycks2020measuring}
Dan Hendrycks, Collin Burns, Steven Basart, Andy Zou, Mantas Mazeika, Dawn Song, and Jacob Steinhardt. 2020.
\newblock Measuring massive multitask language understanding.
\newblock \emph{arXiv preprint arXiv:2009.03300}.

\bibitem[{Hu et~al.(2022)Hu, Salcic, Sun, Dobbie, Yu, and Zhang}]{hu2022membership}
Hongsheng Hu, Zoran Salcic, Lichao Sun, Gillian Dobbie, Philip~S Yu, and Xuyun Zhang. 2022.
\newblock Membership inference attacks on machine learning: A survey.
\newblock \emph{ACM Computing Surveys (CSUR)}, 54(11s):1--37.

\bibitem[{Hurst et~al.(2024)Hurst, Lerer, Goucher, Perelman, Ramesh, Clark, Ostrow, Welihinda, Hayes, Radford et~al.}]{hurst2024gpt}
Aaron Hurst, Adam Lerer, Adam~P Goucher, Adam Perelman, Aditya Ramesh, Aidan Clark, AJ~Ostrow, Akila Welihinda, Alan Hayes, Alec Radford, and 1 others. 2024.
\newblock Gpt-4o system card.
\newblock \emph{arXiv preprint arXiv:2410.21276}.

\bibitem[{Jin et~al.(2020)Jin, Pan, Oufattole, Weng, Fang, and Szolovits}]{jin2020disease}
Di~Jin, Eileen Pan, Nassim Oufattole, Wei-Hung Weng, Hanyi Fang, and Peter Szolovits. 2020.
\newblock What disease does this patient have? a large-scale open domain question answering dataset from medical exams.
\newblock \emph{arXiv preprint arXiv:2009.13081}.

\bibitem[{Kadavath et~al.(2022)Kadavath, Conerly, Askell, Henighan, Drain, Perez, Schiefer, Hatfield-Dodds, DasSarma, Tran-Johnson et~al.}]{kadavath2022language}
Saurav Kadavath, Tom Conerly, Amanda Askell, Tom Henighan, Dawn Drain, Ethan Perez, Nicholas Schiefer, Zac Hatfield-Dodds, Nova DasSarma, Eli Tran-Johnson, and 1 others. 2022.
\newblock Language models (mostly) know what they know.
\newblock \emph{arXiv preprint arXiv:2207.05221}.

\bibitem[{Kaushik et~al.(2021)Kaushik, Gain, Kortylewski, and Yuille}]{kaushik2021understanding}
Prakhar Kaushik, Alex Gain, Adam Kortylewski, and Alan Yuille. 2021.
\newblock Understanding catastrophic forgetting and remembering in continual learning with optimal relevance mapping.
\newblock \emph{arXiv preprint arXiv:2102.11343}.

\bibitem[{Kuhn et~al.(2023)Kuhn, Gal, and Farquhar}]{kuhn2023semantic}
Lorenz Kuhn, Yarin Gal, and Sebastian Farquhar. 2023.
\newblock Semantic uncertainty: Linguistic invariances for uncertainty estimation in natural language generation.
\newblock \emph{arXiv preprint arXiv:2302.09664}.

\bibitem[{Lewkowycz et~al.(2022)Lewkowycz, Andreassen, Dohan, Dyer, Michalewski, Ramasesh, Slone, Anil, Schlag, Gutman-Solo et~al.}]{lewkowycz2022solving}
Aitor Lewkowycz, Anders Andreassen, David Dohan, Ethan Dyer, Henryk Michalewski, Vinay Ramasesh, Ambrose Slone, Cem Anil, Imanol Schlag, Theo Gutman-Solo, and 1 others. 2022.
\newblock Solving quantitative reasoning problems with language models.
\newblock \emph{Advances in Neural Information Processing Systems}, 35:3843--3857.

\bibitem[{Li et~al.(2023{\natexlab{a}})Li, Lv, Chen, Cui, Lu, Florencio, Zhang, Li, and Wei}]{li2023trocr}
Minghao Li, Tengchao Lv, Jingye Chen, Lei Cui, Yijuan Lu, Dinei Florencio, Cha Zhang, Zhoujun Li, and Furu Wei. 2023{\natexlab{a}}.
\newblock Trocr: Transformer-based optical character recognition with pre-trained models.
\newblock In \emph{Proceedings of the AAAI Conference on Artificial Intelligence}, volume~37, pages 13094--13102.

\bibitem[{Li et~al.(2023{\natexlab{b}})Li, Wang, Ding, and Chen}]{li2023large}
Yinheng Li, Shaofei Wang, Han Ding, and Hang Chen. 2023{\natexlab{b}}.
\newblock Large language models in finance: A survey.
\newblock In \emph{Proceedings of the fourth ACM international conference on AI in finance}, pages 374--382.

\bibitem[{Liu et~al.(2024{\natexlab{a}})Liu, Feng, Xue, Wang, Wu, Lu, Zhao, Deng, Zhang, Ruan et~al.}]{liu2024deepseek}
Aixin Liu, Bei Feng, Bing Xue, Bingxuan Wang, Bochao Wu, Chengda Lu, Chenggang Zhao, Chengqi Deng, Chenyu Zhang, Chong Ruan, and 1 others. 2024{\natexlab{a}}.
\newblock Deepseek-v3 technical report.
\newblock \emph{arXiv preprint arXiv:2412.19437}.

\bibitem[{Liu et~al.(2024{\natexlab{b}})Liu, Xia, Wang, and Zhang}]{liu2024your}
Jiawei Liu, Chunqiu~Steven Xia, Yuyao Wang, and Lingming Zhang. 2024{\natexlab{b}}.
\newblock Is your code generated by chatgpt really correct? rigorous evaluation of large language models for code generation.
\newblock \emph{Advances in Neural Information Processing Systems}, 36.

\bibitem[{Manakul et~al.(2023)Manakul, Liusie, and Gales}]{manakul2023selfcheckgpt}
Potsawee Manakul, Adian Liusie, and Mark~JF Gales. 2023.
\newblock Selfcheckgpt: Zero-resource black-box hallucination detection for generative large language models.
\newblock \emph{arXiv preprint arXiv:2303.08896}.

\bibitem[{McCloskey and Cohen(1989)}]{mccloskey1989catastrophic}
Michael McCloskey and Neal~J Cohen. 1989.
\newblock Catastrophic interference in connectionist networks: The sequential learning problem.
\newblock In \emph{Psychology of learning and motivation}, volume~24, pages 109--165. Elsevier.

\bibitem[{Nie et~al.(2023)Nie, Liu, Fu, Xue, Jiao, Miao, Tao, and Cui}]{nie2023angel}
Xiaonan Nie, Yi~Liu, Fangcheng Fu, Jinbao Xue, Dian Jiao, Xupeng Miao, Yangyu Tao, and Bin Cui. 2023.
\newblock Angel-ptm: A scalable and economical large-scale pre-training system in tencent.
\newblock \emph{Proceedings of the VLDB Endowment}, 16(12):3781--3794.

\bibitem[{Pal et~al.(2022)Pal, Umapathi, and Sankarasubbu}]{pal2022medmcqa}
Ankit Pal, Logesh~Kumar Umapathi, and Malaikannan Sankarasubbu. 2022.
\newblock Medmcqa: A large-scale multi-subject multi-choice dataset for medical domain question answering.
\newblock In \emph{Conference on health, inference, and learning}, pages 248--260. PMLR.

\bibitem[{Radford et~al.(2019)Radford, Wu, Child, Luan, Amodei, Sutskever et~al.}]{radford2019language}
Alec Radford, Jeffrey Wu, Rewon Child, David Luan, Dario Amodei, Ilya Sutskever, and 1 others. 2019.
\newblock Language models are unsupervised multitask learners.
\newblock \emph{OpenAI blog}, 1(8):9.

\bibitem[{Roziere et~al.(2023)Roziere, Gehring, Gloeckle, Sootla, Gat, Tan, Adi, Liu, Sauvestre, Remez et~al.}]{roziere2023code}
Baptiste Roziere, Jonas Gehring, Fabian Gloeckle, Sten Sootla, Itai Gat, Xiaoqing~Ellen Tan, Yossi Adi, Jingyu Liu, Romain Sauvestre, Tal Remez, and 1 others. 2023.
\newblock Code llama: Open foundation models for code.
\newblock \emph{arXiv preprint arXiv:2308.12950}.

\bibitem[{Sanh et~al.(2021)Sanh, Webson, Raffel, Bach, Sutawika, Alyafeai, Chaffin, Stiegler, Scao, Raja et~al.}]{sanh2021multitask}
Victor Sanh, Albert Webson, Colin Raffel, Stephen~H Bach, Lintang Sutawika, Zaid Alyafeai, Antoine Chaffin, Arnaud Stiegler, Teven~Le Scao, Arun Raja, and 1 others. 2021.
\newblock Multitask prompted training enables zero-shot task generalization.
\newblock \emph{arXiv preprint arXiv:2110.08207}.

\bibitem[{Singhal et~al.(2023)Singhal, Azizi, Tu, Mahdavi, Wei, Chung, Scales, Tanwani, Cole-Lewis, Pfohl et~al.}]{singhal2023large}
Karan Singhal, Shekoofeh Azizi, Tao Tu, S~Sara Mahdavi, Jason Wei, Hyung~Won Chung, Nathan Scales, Ajay Tanwani, Heather Cole-Lewis, Stephen Pfohl, and 1 others. 2023.
\newblock Large language models encode clinical knowledge.
\newblock \emph{Nature}, 620(7972):172--180.

\bibitem[{Sun et~al.(2024)Sun, Han, Zhao, Ma, Shen, Chen, Chen, and Yu}]{sun2024scieval}
Liangtai Sun, Yang Han, Zihan Zhao, Da~Ma, Zhennan Shen, Baocai Chen, Lu~Chen, and Kai Yu. 2024.
\newblock Scieval: A multi-level large language model evaluation benchmark for scientific research.
\newblock In \emph{Proceedings of the AAAI Conference on Artificial Intelligence}, volume~38, pages 19053--19061.

\bibitem[{Taori et~al.(2023)Taori, Gulrajani, Zhang, Dubois, Li, Guestrin, Liang, and Hashimoto}]{taori2023stanford}
Rohan Taori, Ishaan Gulrajani, Tianyi Zhang, Yann Dubois, Xuechen Li, Carlos Guestrin, Percy Liang, and Tatsunori~B Hashimoto. 2023.
\newblock Stanford alpaca: An instruction-following llama model.

\bibitem[{Taylor et~al.(2022)Taylor, Kardas, Cucurull, Scialom, Hartshorn, Saravia, Poulton, Kerkez, and Stojnic}]{taylor2022galactica}
Ross Taylor, Marcin Kardas, Guillem Cucurull, Thomas Scialom, Anthony Hartshorn, Elvis Saravia, Andrew Poulton, Viktor Kerkez, and Robert Stojnic. 2022.
\newblock Galactica: A large language model for science.
\newblock \emph{arXiv preprint arXiv:2211.09085}.

\bibitem[{Team et~al.(2023)Team, Anil, Borgeaud, Wu, Alayrac, Yu, Soricut, Schalkwyk, Dai, Hauth et~al.}]{team2023gemini}
Gemini Team, Rohan Anil, Sebastian Borgeaud, Yonghui Wu, Jean-Baptiste Alayrac, Jiahui Yu, Radu Soricut, Johan Schalkwyk, Andrew~M Dai, Anja Hauth, and 1 others. 2023.
\newblock Gemini: a family of highly capable multimodal models.
\newblock \emph{arXiv preprint arXiv:2312.11805}.

\bibitem[{Thirunavukarasu et~al.(2023)Thirunavukarasu, Ting, Elangovan, Gutierrez, Tan, and Ting}]{thirunavukarasu2023large}
Arun~James Thirunavukarasu, Darren Shu~Jeng Ting, Kabilan Elangovan, Laura Gutierrez, Ting~Fang Tan, and Daniel Shu~Wei Ting. 2023.
\newblock Large language models in medicine.
\newblock \emph{Nature medicine}, 29(8):1930--1940.

\bibitem[{Touvron et~al.(2023{\natexlab{a}})Touvron, Lavril, Izacard, Martinet, Lachaux, Lacroix, Rozi{\`e}re, Goyal, Hambro, Azhar et~al.}]{touvron2023llama}
Hugo Touvron, Thibaut Lavril, Gautier Izacard, Xavier Martinet, Marie-Anne Lachaux, Timoth{\'e}e Lacroix, Baptiste Rozi{\`e}re, Naman Goyal, Eric Hambro, Faisal Azhar, and 1 others. 2023{\natexlab{a}}.
\newblock Llama: Open and efficient foundation language models.
\newblock \emph{arXiv preprint arXiv:2302.13971}.

\bibitem[{Touvron et~al.(2023{\natexlab{b}})Touvron, Martin, Stone, Albert, Almahairi, Babaei, Bashlykov, Batra, Bhargava, Bhosale et~al.}]{touvron2023llama2}
Hugo Touvron, Louis Martin, Kevin Stone, Peter Albert, Amjad Almahairi, Yasmine Babaei, Nikolay Bashlykov, Soumya Batra, Prajjwal Bhargava, Shruti Bhosale, and 1 others. 2023{\natexlab{b}}.
\newblock Llama 2: Open foundation and fine-tuned chat models.
\newblock \emph{arXiv preprint arXiv:2307.09288}.

\bibitem[{Wang et~al.(2022)Wang, Kordi, Mishra, Liu, Smith, Khashabi, and Hajishirzi}]{selfinstruct}
Yizhong Wang, Yeganeh Kordi, Swaroop Mishra, Alisa Liu, Noah~A. Smith, Daniel Khashabi, and Hannaneh Hajishirzi. 2022.
\newblock Self-instruct: Aligning language model with self generated instructions.

\bibitem[{Wang et~al.(2023)Wang, Zhong, Wang, Zhu, Mi, Wang, Shang, Jiang, and Liu}]{wang2023data}
Zige Wang, Wanjun Zhong, Yufei Wang, Qi~Zhu, Fei Mi, Baojun Wang, Lifeng Shang, Xin Jiang, and Qun Liu. 2023.
\newblock Data management for large language models: A survey.
\newblock \emph{arXiv preprint arXiv:2312.01700}.

\bibitem[{Wu et~al.(2025)Wu, Wang, Yang, Gan, Liu, Yuan, and Wang}]{wu2025grit}
Jialian Wu, Jianfeng Wang, Zhengyuan Yang, Zhe Gan, Zicheng Liu, Junsong Yuan, and Lijuan Wang. 2025.
\newblock Grit: A generative region-to-text transformer for object understanding.
\newblock In \emph{European Conference on Computer Vision}, pages 207--224. Springer.

\bibitem[{Wu et~al.(2023)Wu, Irsoy, Lu, Dabravolski, Dredze, Gehrmann, Kambadur, Rosenberg, and Mann}]{wu2023bloomberggpt}
Shijie Wu, Ozan Irsoy, Steven Lu, Vadim Dabravolski, Mark Dredze, Sebastian Gehrmann, Prabhanjan Kambadur, David Rosenberg, and Gideon Mann. 2023.
\newblock Bloomberggpt: A large language model for finance.
\newblock \emph{arXiv preprint arXiv:2303.17564}.

\bibitem[{Xia et~al.(2023)Xia, Gao, Zeng, and Chen}]{xia2023sheared}
Mengzhou Xia, Tianyu Gao, Zhiyuan Zeng, and Danqi Chen. 2023.
\newblock Sheared llama: Accelerating language model pre-training via structured pruning.
\newblock \emph{arXiv preprint arXiv:2310.06694}.

\bibitem[{Xie et~al.(2024{\natexlab{a}})Xie, Han, Chen, Xiang, Zhang, He, Xiao, Li, Dai, Feng et~al.}]{xie2024finben}
Qianqian Xie, Weiguang Han, Zhengyu Chen, Ruoyu Xiang, Xiao Zhang, Yueru He, Mengxi Xiao, Dong Li, Yongfu Dai, Duanyu Feng, and 1 others. 2024{\natexlab{a}}.
\newblock The finben: An holistic financial benchmark for large language models.
\newblock \emph{arXiv preprint arXiv:2402.12659}.

\bibitem[{Xie et~al.(2024{\natexlab{b}})Xie, Pham, Dong, Du, Liu, Lu, Liang, Le, Ma, and Yu}]{xie2024doremi}
Sang~Michael Xie, Hieu Pham, Xuanyi Dong, Nan Du, Hanxiao Liu, Yifeng Lu, Percy~S Liang, Quoc~V Le, Tengyu Ma, and Adams~Wei Yu. 2024{\natexlab{b}}.
\newblock Doremi: Optimizing data mixtures speeds up language model pretraining.
\newblock \emph{Advances in Neural Information Processing Systems}, 36.

\bibitem[{Xu et~al.(2024)Xu, Shi, and Liang}]{xu2024large}
Zhuoyan Xu, Zhenmei Shi, and Yingyu Liang. 2024.
\newblock Do large language models have compositional ability? an investigation into limitations and scalability.
\newblock In \emph{ICLR 2024 Workshop on Mathematical and Empirical Understanding of Foundation Models}.

\bibitem[{Yang et~al.(2024)Yang, Yang, Hui, Zheng, Yu, Zhou, Li, Li, Liu, Huang et~al.}]{yang2024qwen2}
An~Yang, Baosong Yang, Binyuan Hui, Bo~Zheng, Bowen Yu, Chang Zhou, Chengpeng Li, Chengyuan Li, Dayiheng Liu, Fei Huang, and 1 others. 2024.
\newblock Qwen2 technical report.
\newblock \emph{arXiv preprint arXiv:2407.10671}.

\bibitem[{Ye et~al.(2022)Ye, Maddi, Murakonda, Bindschaedler, and Shokri}]{ye2022enhanced}
Jiayuan Ye, Aadyaa Maddi, Sasi~Kumar Murakonda, Vincent Bindschaedler, and Reza Shokri. 2022.
\newblock Enhanced membership inference attacks against machine learning models.
\newblock In \emph{Proceedings of the 2022 ACM SIGSAC Conference on Computer and Communications Security}, pages 3093--3106.

\bibitem[{Yuan et~al.(2022)Yuan, Yuan, Tan, Huang, and Huang}]{yuan2022hype}
Hongyi Yuan, Zheng Yuan, Chuanqi Tan, Fei Huang, and Songfang Huang. 2022.
\newblock Hype: Better pre-trained language model fine-tuning with hidden representation perturbation.
\newblock \emph{arXiv preprint arXiv:2212.08853}.

\bibitem[{Zellers et~al.(2019)Zellers, Holtzman, Bisk, Farhadi, and Choi}]{zellers2019hellaswag}
Rowan Zellers, Ari Holtzman, Yonatan Bisk, Ali Farhadi, and Yejin Choi. 2019.
\newblock Hellaswag: Can a machine really finish your sentence?
\newblock \emph{arXiv preprint arXiv:1905.07830}.

\bibitem[{Zhang et~al.(2024)Zhang, Hu, Zhoubian, Du, Yang, Wang, Yue, Dong, and Tang}]{zhang2024sciglm}
Dan Zhang, Ziniu Hu, Sining Zhoubian, Zhengxiao Du, Kaiyu Yang, Zihan Wang, Yisong Yue, Yuxiao Dong, and Jie Tang. 2024.
\newblock Sciglm: Training scientific language models with self-reflective instruction annotation and tuning.
\newblock \emph{arXiv preprint arXiv:2401.07950}.

\bibitem[{Zhang et~al.(2023)Zhang, Cai, Liu, Yang, Dai, Liao, Qin, Li, Liu, Liu et~al.}]{zhang2023fineval}
Liwen Zhang, Weige Cai, Zhaowei Liu, Zhi Yang, Wei Dai, Yujie Liao, Qianru Qin, Yifei Li, Xingyu Liu, Zhiqiang Liu, and 1 others. 2023.
\newblock Fineval: A chinese financial domain knowledge evaluation benchmark for large language models.
\newblock \emph{arXiv preprint arXiv:2308.09975}.

\bibitem[{Zhang and Yang(2023)}]{zhang2023xuanyuan}
Xuanyu Zhang and Qing Yang. 2023.
\newblock Xuanyuan 2.0: A large chinese financial chat model with hundreds of billions parameters.
\newblock In \emph{Proceedings of the 32nd ACM international conference on information and knowledge management}, pages 4435--4439.

\bibitem[{Zhao et~al.(2023{\natexlab{a}})Zhao, Zhou, Li, Tang, Wang, Hou, Min, Zhang, Zhang, Dong et~al.}]{zhao2023survey}
Wayne~Xin Zhao, Kun Zhou, Junyi Li, Tianyi Tang, Xiaolei Wang, Yupeng Hou, Yingqian Min, Beichen Zhang, Junjie Zhang, Zican Dong, and 1 others. 2023{\natexlab{a}}.
\newblock A survey of large language models.
\newblock \emph{arXiv preprint arXiv:2303.18223}.

\bibitem[{Zhao et~al.(2023{\natexlab{b}})Zhao, Yan, Sun, Xing, Meng, Wang, Cheng, Ren, and Yin}]{zhao2023knowing}
Yukun Zhao, Lingyong Yan, Weiwei Sun, Guoliang Xing, Chong Meng, Shuaiqiang Wang, Zhicong Cheng, Zhaochun Ren, and Dawei Yin. 2023{\natexlab{b}}.
\newblock Knowing what llms do not know: A simple yet effective self-detection method.
\newblock \emph{arXiv preprint arXiv:2310.17918}.

\bibitem[{Zhong et~al.(2023)Zhong, Cui, Guo, Liang, Lu, Wang, Saied, Chen, and Duan}]{zhong2023agieval}
Wanjun Zhong, Ruixiang Cui, Yiduo Guo, Yaobo Liang, Shuai Lu, Yanlin Wang, Amin Saied, Weizhu Chen, and Nan Duan. 2023.
\newblock Agieval: A human-centric benchmark for evaluating foundation models.
\newblock \emph{arXiv preprint arXiv:2304.06364}.

\end{thebibliography}

\newpage
\appendix
\onecolumn

\renewcommand{\contentsname}{Appendix}
\tableofcontents
\addtocontents{toc}{\protect\setcounter{tocdepth}{2}}

\section{Background and Discussion}
\label{sec:appendix_background_and_discussion}

In this section, we provide the background information and design motivation for our VersaTune.

\subsection{Pretraining and Supervised Fine-Tuning}

The training process of Large Language Models (LLMs) generally involves the pretraining and fine-tuning stages. We have outlined several concepts about LLMs training.

\subsubsection{Pretraining}
\label{subsubsec:appendix_pretraining}

Large Language Models (LLMs) establish basic knowledge abilities, including language understanding and text generation, during the pretraining stage~\cite{brown2020language}. 
In this stage, LLMs engage in unsupervised training through the processing of extensive raw text corpora, thereby enhancing their capabilities in language modeling. For a given sequence $\mathrm {\textbf{x}} = \{x_1, x_2, ..., x_n\}$, the typical task for LLMs involves the prediction of the subsequent token $x_i$ given the preceding tokens $\mathrm {\textbf{x}}_{< i}$ as contextual input. The goal is to maximize the likelihood function presented in \Cref{eq:pretraining}:
\begin{gather}
\mathcal{L} _{LLM}^{PT}(\mathrm {\textbf{x}}) = \sum_{i=1}^{n}\log{P(x_i|\mathrm {\textbf{x}}_{< i})}
\label{eq:pretraining}
\end{gather}

\begin{wrapfigure}{r}{.5\textwidth}
    \centering
    \vspace{-1em}

    \includegraphics[width=\linewidth]{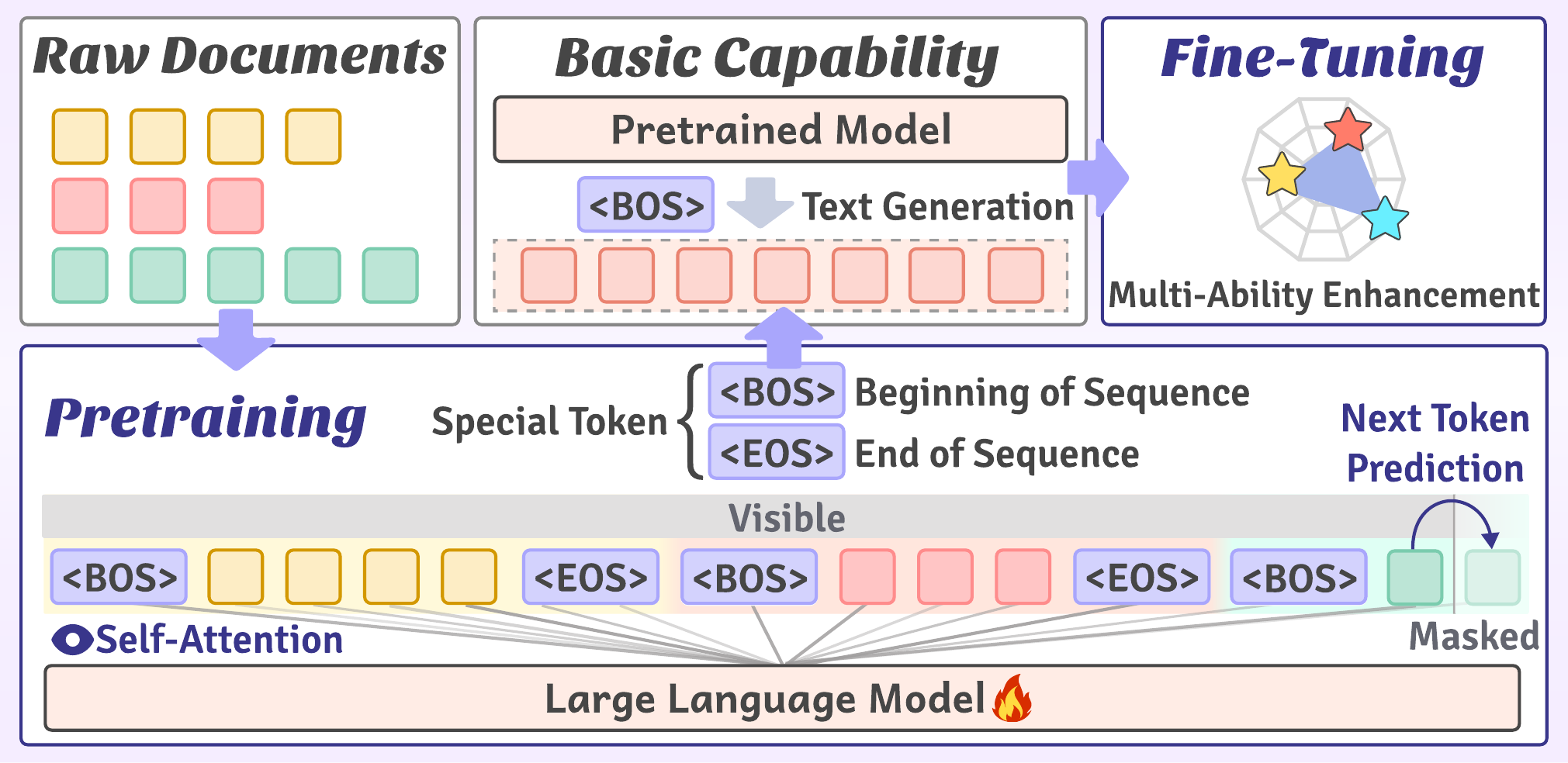}
    \caption{Illustration of the LLMs training workflow. In the pretraining phase, raw documents are concatenated into a sequence using special tokens such as <BOS> (Beginning of Sequence) and <EOS> (End of Sequence), thereby endowing the LLM with fundamental language generation capabilities. In the fine-tuning phase, the model's abilities in various domains are further enhanced.}
   \label{fig:preliminary}
    \vspace{-2em}
\end{wrapfigure}


\hypertarget{BOS}{\textbf{\textit{Beginning of Sequence}}  \textbf{(\textit{<BOS>})}} \quad
During the above process, the Beginning of Sequence (<BOS>) token plays an important role, which serves as a signal to the model that the input sequence is starting~\cite{brown2020language,li2023trocr,wu2025grit}. It can be thought of as a special marker that indicates the start of a new sequence, allowing the model to reset its context and begin processing a new piece of text. 
In the context of pretraining, the <BOS> token is used to initialize the input to the model, and it can be concatenated with the actual text data to form the input sequence. This token helps the model to differentiate between the start of a new input and the continuation of an existing one. It is particularly crucial in tasks where \textit{the model needs to generate text} or understand the beginning of a new sentence or document, which helps the model to learn the boundaries of text sequences and to better model the statistical properties of the language data it is trained on.
The use of <BOS> tokens, along with other special tokens like \textit{<EOS> (End of Sequence)}, helps the model to learn the boundaries of text sequences and to better model the statistical properties of the language data it is trained on.

\subsubsection{Supervised Fine-Tuning}
\label{subsubsec:appendix_sft}

The Supervised Fine-Tuning (SFT) stage of a Large Language Model (LLM) involves further training to refine the model's task-solving capabilities and ensure greater alignment with human instructions~\cite{zhao2023survey}. 
While recent research has delved into exploring fine-tuning methods for multi-task enhancement~\cite{dong2023abilities,sanh2021multitask}, they are still in their early stages. 
However, as shown by proprietary models such as GPT-4~\cite{achiam2023gpt}, Gemini~\cite{team2023gemini}, and DeepSeek series~\cite{liu2024deepseek}, which exhibit outstanding multi-task performance, improving a model's versatile capabilities across various domains during the SFT phase is crucial. Therefore, our work systematically investigates methods to \textit{enhance multi-domain performance} during the SFT stage to bridge this gap.

\subsection{Analysis on Catastrophic Forgetting}
\label{subsec:appendix_catastrophic_forgetting}

\begin{wraptable}{r}{0.5\textwidth}
\centering
\vskip -0.1in
\resizebox{\linewidth}{!}{
\begin{tabular}{ccccccccc}
\toprule
\multirow{2}{*}{\begin{tabular}[c]{@{}c@{}}Target Domain\end{tabular}} & \multicolumn{1}{|c}{Training Step} & \multicolumn{6}{|c|}{Variations in Comprehensive Domains ($\%$)} & \multirow{2}{*}{Sum. ($\%$)} \\
\cline{3-8}
   & \multicolumn{1}{|c}{(Epoch)} & \multicolumn{1}{|c}{Law}  & Medicine  & Finance  & Science  & Code  & \multicolumn{1}{c|}{Other}  & \\
\midrule
\multirow{4}{*}{Law} & \multicolumn{1}{|c|}{1} & - & $\color{red} \downarrow $18.82 & $\color{green} \uparrow $14.71 & $\color{red} \downarrow $11.76 & $\color{red} \downarrow $11.18 & $\color{red} \downarrow $5.00 & \multicolumn{1}{|c}{$\color{red} \downarrow $32.05} \\
   & \multicolumn{1}{|c|}{2} & - & $\color{red} \downarrow $12.65 & $\color{green} \uparrow $30.59 & $\color{red} \downarrow $4.41 & $\color{red} \downarrow $5.29 & $\color{red} \downarrow $11.76 & \multicolumn{1}{|c}{$\color{red} \downarrow $3.52} \\
   & \multicolumn{1}{|c|}{3} & - & $\color{red} \downarrow $17.94 & $\color{green} \uparrow $12.35 & $\color{red} \downarrow $8.82 & $\color{red} \downarrow $23.53 & $\color{red} \downarrow$5.00 & \multicolumn{1}{|c}{$\color{red} \downarrow $42.94} \\
   & \multicolumn{1}{|c|}{4} & - & $\color{red} \downarrow $5.00 & $\color{red} \downarrow $2.06 & $\color{red} \downarrow $31.18 & $\color{red} \downarrow $21.76 & $\color{red} \downarrow $12.65 & \multicolumn{1}{|c}{$\color{red} \downarrow $\textbf{72.65}} \\
\midrule
\multirow{4}{*}{Medicine} & \multicolumn{1}{|c|}{1} & $\color{red} \downarrow $10.29 & - & $\color{red} \downarrow $3.82 & $\color{green} \uparrow $24.12 & $\color{red} \downarrow $19.41 & $\color{red} \downarrow $7.35 & \multicolumn{1}{|c}{$\color{red} \downarrow $16.75} \\
   & \multicolumn{1}{|c|}{2} & $\color{red} \downarrow $18.82 & - & $\color{red} \downarrow $6.47 & $\color{green} \uparrow $40.00 & $\color{red} \downarrow $7.94 & $\color{red} \downarrow $17.65 & \multicolumn{1}{|c}{$\color{red} \downarrow $10.88} \\
   & \multicolumn{1}{|c|}{3} & $\color{red} \downarrow $22.35 & - & $\color{red} \downarrow $8.82 & $\color{green} \uparrow $7.94 & $\color{red} \downarrow $19.12 & $\color{red} \downarrow $10.00 & \multicolumn{1}{|c}{$\color{red} \downarrow $52.35} \\
   & \multicolumn{1}{|c|}{4} & $\color{red} \downarrow $27.94 & - & $\color{red} \downarrow $11.76 & $\color{red} \downarrow $2.35 & $\color{red} \downarrow $21.76 & $\color{red} \downarrow $12.65 & \multicolumn{1}{|c}{$\color{red} \downarrow $\textbf{76.46}} \\
\midrule
\multirow{4}{*}{Finance} & \multicolumn{1}{|c|}{1} & $\color{green} \uparrow $20.59 & $\color{red} \downarrow $7.94 & - & $\color{red} \downarrow $10.29 & $\color{red} \downarrow $12.65 & $\color{red} \downarrow $6.47 & \multicolumn{1}{|c}{$\color{red} \downarrow $16.76} \\
   & \multicolumn{1}{|c|}{2} & $\color{green} \uparrow $18.24 & $\color{red} \downarrow $9.71 & - & $\color{red} \downarrow $9.41 & $\color{green} \uparrow $5.29 & $\color{red} \downarrow $8.82 & \multicolumn{1}{|c}{$\color{red} \downarrow $4.41} \\
   & \multicolumn{1}{|c|}{3} & $\color{green} \uparrow $23.53 & $\color{red} \downarrow $9.41 & - & $\color{red} \downarrow $17.35 & $\color{red} \downarrow $14.71 & $\color{red} \downarrow $7.94 & \multicolumn{1}{|c}{$\color{red} \downarrow $25.88} \\
   & \multicolumn{1}{|c|}{4} & $\color{green} \uparrow $5.00 & $\color{red} \downarrow $9.12 & - & $\color{red} \downarrow $20.29 & $\color{red} \downarrow $12.94 & $\color{red} \downarrow $21.76 & \multicolumn{1}{|c}{$\color{red} \downarrow $\textbf{59.11}} \\
\midrule
\multirow{4}{*}{Science} & \multicolumn{1}{|c|}{1} & $\color{red} \downarrow $10.29 & $\color{green} \uparrow $17.06 & $\color{red} \downarrow $3.82 & - & $\color{red} \downarrow $4.71 & $\color{red} \downarrow $7.35 & \multicolumn{1}{|c}{$\color{red} \downarrow $9.11} \\
   & \multicolumn{1}{|c|}{2} & $\color{red} \downarrow $11.47 & $\color{green} \uparrow $12.35 & $\color{red} \downarrow $4.71 & - & $\color{red} \downarrow $5.88 & $\color{red} \downarrow $12.94 & \multicolumn{1}{|c}{$\color{red} \downarrow $22.65} \\
   & \multicolumn{1}{|c|}{3} & $\color{red} \downarrow $21.76 & $\color{green} \uparrow $7.94 & $\color{red} \downarrow $8.82 & - & $\color{red} \downarrow $4.41 & $\color{red} \downarrow $10.00 & \multicolumn{1}{|c}{$\color{red} \downarrow $37.05} \\
   & \multicolumn{1}{|c|}{4} & $\color{red} \downarrow $27.94 & $\color{green} \uparrow $2.35 & $\color{red} \downarrow $11.47 & - & $\color{red} \downarrow $12.65 & $\color{red} \downarrow $12.59 & \multicolumn{1}{|c}{$\color{red} \downarrow $\textbf{62.30}} \\
\midrule
\multirow{4}{*}{Code} & \multicolumn{1}{|c|}{1} & $\color{red} \downarrow $3.82 & $\color{green} \uparrow $7.35 & $\color{red} \downarrow $17.35 & $\color{green} \uparrow $9.12 & - & $\color{red} \downarrow $7.29 & \multicolumn{1}{|c}{$\color{red} \downarrow $11.99} \\
   & \multicolumn{1}{|c|}{2} & $\color{red} \downarrow $9.71 & $\color{red} \downarrow $6.47 & $\color{red} \downarrow $7.94 & $\color{green} \uparrow $5.29 & - & $\color{red} \downarrow $6.18 & \multicolumn{1}{|c}{$\color{red} \downarrow $25.01} \\
   & \multicolumn{1}{|c|}{3} & $\color{red} \downarrow $22.35 & $\color{red} \downarrow $8.82 & $\color{red} \downarrow $14.12 & $\color{green} \uparrow $7.94 & - & $\color{red} \downarrow $10.02 & \multicolumn{1}{|c}{$\color{red} \downarrow $47.37} \\
   & \multicolumn{1}{|c|}{4} & $\color{red} \downarrow $26.18 & $\color{red} \downarrow $7.06 & $\color{red} \downarrow $8.82 & $\color{red} \downarrow $3.24 & - & $\color{red} \downarrow $22.65 & \multicolumn{1}{|c}{$\color{red} \downarrow $\textbf{67.95}} \\
\bottomrule
\end{tabular}
}
\caption{\label{tab:catastrophic_forgetting} 
Variations in models' performance on non-target domain tasks when trained on single sourced dataset.${\color{green} \uparrow}$ and ${\color{red} \downarrow}$ indicate an increase or decrease in the percentage of scores (\%) compared to the \textit{initial state} before fine-tuning.}
\vskip -0.2in
\end{wraptable}

\begin{figure*}[ht]
    \begin{center}
    \includegraphics[width=1\linewidth]{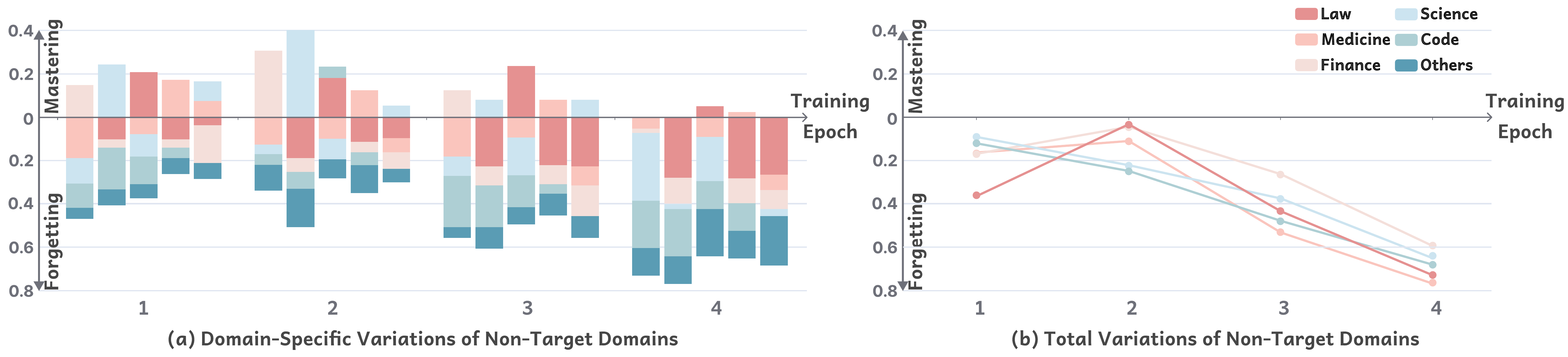}
    \caption{Illustration of variations in models' performance on non-target domain tasks when trained on a single-domain dataset. 
    The grouped stacked bar chart on the \textbf{left (a)} describes the detailed changes in performance across various non-target domains as training progresses. 
    Each group of stacked bars, from left to right, represents the use of training datasets from \textit{law}, \textit{medicine}, \textit{finance}, \textit{science}, and \textit{code}, respectively. 
    The line chart on the \textbf{right (b)} shows the overall performance changes in all non-target domains. The color of each line indicates the domain from which the training dataset was sourced.
   }
    \label{fig:catastrophic_forgetting}
    \end{center}
    \vskip -0.2in
\end{figure*}

During the SFT phase, it is a typical practice to employ datasets specific to a particular domain for the fine-tuning of LLMs, which may lead to a significant performance drop of knowledge in non-target domains, a phenomenon commonly referred to as Catastrophic Forgetting ~\cite{kaushik2021understanding,mccloskey1989catastrophic}. 
We conducted experiments on open-sourced models including LLaMA~\cite{dubey2024llama,touvron2023llama,touvron2023llama2} and Qwen~\cite{bai2023qwen,yang2024qwen2} series to assess how the model's proficiency in other domains changes when fine-tuned with data from a single domain, as depicted in \Cref{tab:catastrophic_forgetting} and \Cref{fig:catastrophic_forgetting}. 
We have regulated the number of training instances per epoch to a fixed count of 10,000. 
More details on training and evaluation settings can be found in \Cref{subsec:experimental_setup} and \Cref{sec:appendix_experimental_detail}. 
Our findings indicate that \textit{when a model is trained exclusively with data from \textbf{a single domain}, its performance on tasks from other domains tends to \textbf{degrade progressively} over the course of training}. 
This experimental outcome has provided significant motivation and direction for our work.

\clearpage

\section{Method Details}
\label{sec:appendix_method}

\subsection{Algorithm for Flexible Domain Expansion}
\label{subsec:appendix_domain_expansion}

Here we provide the detailed algorithm of flexible domain expansion (\Cref{subsec:flexible_domain_expansion}). 
As outlined in \Cref{alg:domain_expansion}, we initially establish the data distribution based on the knowledge detected from the original pretrained model (Line 1). At each training step $t$, we calculate the learnable potential and forgetting degree scores for each domain (Line 4-5), and assign domain weights for the current training phase following the method from \Cref{alg:multi_ability} (Line 6). A trade-off is necessary between the remaining learning margin of the domain that requires focused cultivation and the model's forgetting degree towards other non-target domains. If the improvement benefit of the specific domain exceeds the average forgetting degree of the other domains (ratio greater than $\varepsilon$), we increase the data weight of the current specific domain by $\delta$, and proportionally reduce the weights of the other non-target domains according to \Cref{eq:statement2_cases} (Line 8-9). Otherwise, we maintain the current domain distribution and only perform minor adjustments and renormalization as described in \cref{alg:multi_ability} (Line 10-11). Subsequently, we update the parameters of the target model $M_{\theta}$ (Line 13).

\begin{algorithm}[ht]
\caption{VersaTune Multi-Ability Fine-Tuning (for Domain Expansion)}
\label{alg:domain_expansion}
\begin{flushleft}
\textbf{Input}: Base model to be fine-tuned $M_{\theta}^{(0)}$, Domains that require enhanced cultivation $D_e$, Domain reference loss $\{\ell_{ref}^{j}\}_{j=1}^{k}$, Hyperparameters: number of training steps $T$, magnitude of adjustment $\sigma $, extent of domain adjustment $\delta$, variation threshold $\varepsilon$ \\
\mbox{\textbf{Parameter}: Data proportion $\{P_{j}\}_{j=1}^{k}$ of the SFT dataset} \\
\textbf{Output}: Fine-tuned multi-ability model $M_{\theta}^{(T)}$ \\
Define $\gamma$: learnable potential  of the current domain \\
Define $\varphi$: forgetting degree of the current domain \\
\end{flushleft}
\begin{algorithmic}[1]
\STATE Initialize domain proportion $\{P_{j}^{(0)}\}_{j=1}^{k}$ according to \Cref{eq:statement1} and \Cref{alg:domain_detection}
\FOR{$t = 1, 2, \ldots, T$}
    \FOR{$j = 1, 2, \ldots, k $} 
        \STATE Learnable potential for the $j$-th domain: $\gamma_j^{(t)} = \max \{\frac{\ell_{\theta^{(t)}}^{j} - \ell_{ref}^{j}}{\ell_{\theta^{(t)}}^{j}}, 0 \}$
        \STATE Forgetting degree for the $j$-th domain: $\varphi_j^{(t)} = \max \{\frac{\ell_{\theta^{(t)}}^{j} - \ell_{\theta^{(t - 1)}}^{j}}{\ell_{\theta^{(t - 1)}}^{j}}, 0 \}$
        \STATE Update domain weights: $P_j^{(t)'} = P_j^{(t-1)}(1 + \sigma 
 \gamma_j^{(t)})$
    \ENDFOR
    \IF{$\frac{1}{k} \sum_{j=1, j \neq e}^{k} \varphi_j^{(t)} < \varepsilon \gamma_e^{(t)}$}
        \STATE Update specific domain weight: \\ \centering{$P_j^{(t)} = \begin{cases}  
            P_j^{(t-1)} + \delta, \text{if} \quad j = e & \\  
            \frac{P_j^{(t)'}}{\sum_{i=1, j \neq e}^{k}P_i^{(t)'}} (1 - P_j^{(t-1)} - \delta), \text{others} &
        \end{cases}$}
    \ELSE
        \STATE Renormalize domain weights: $P_j^{(t)} = \frac{P_j^{(t)'}}{\sum_{i=1}^{k}P_i^{(t)'}}, \quad \forall j \in \{1, 2, ..., k\}$
    \ENDIF
    \STATE \mbox{Update parameters of fine-tuned model $M_{\theta}^{(t)}$}
\ENDFOR
\STATE \textbf{Return} Fine-tuned model $M_{\theta}^{(T)}$
\end{algorithmic}
\end{algorithm}

\clearpage

\section{Experiments Details}
\label{sec:appendix_experimental_detail}

\subsection{Knowledge Distribution Detection}
\label{sec:appendix_knowledge_distribution}

\begin{figure*}[ht]
    \begin{center}
    \centerline{\includegraphics[width=\textwidth]{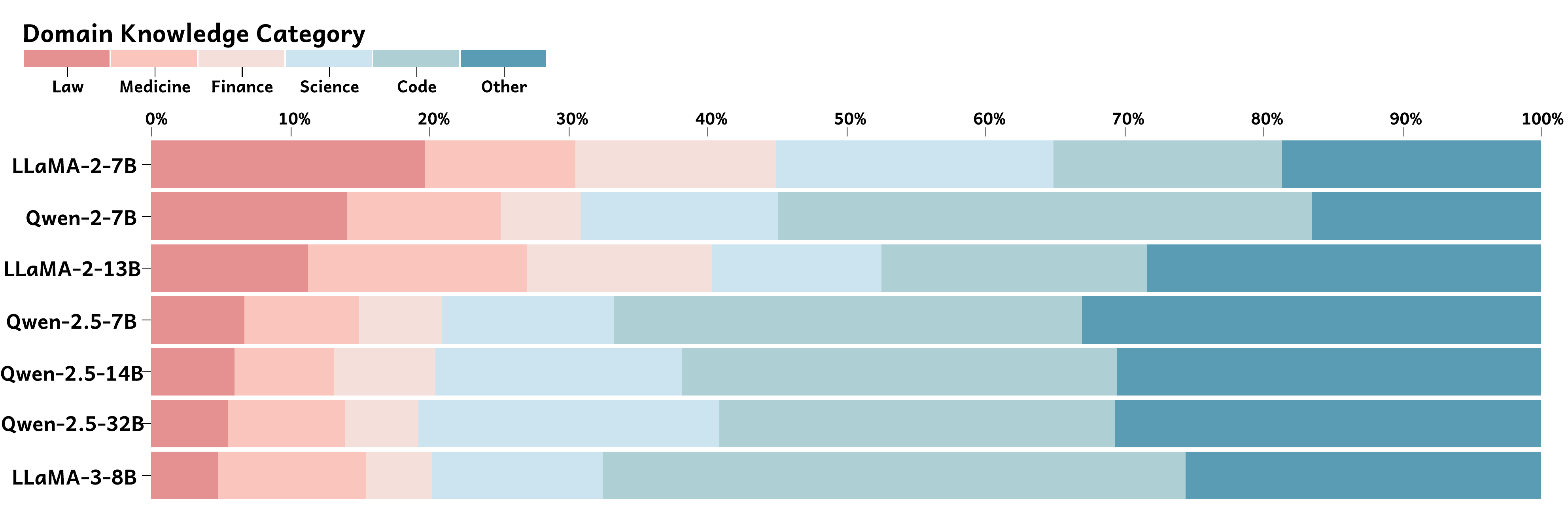}}
    \caption{
    \label{fig:knowledge_detection_stacked_bar_chart}
    An illustration of the domain knowledge distribution among models.}
    \end{center}
\end{figure*}

During the knowledge distribution detection phase for our target models, we have manually annotated 120 samples (20 samples for each domain) to fine-tune Qwen2.5-72B-Instruct\footnote{https://huggingface.co/Qwen/Qwen2.5-72B-Instruct}, and employed the trained model as the proprietary model $M_P$. For each target model $M_\theta$ slated for supervised fine-tuning, we prompted the generation of $40K$ data samples using the Beginning of Sequence ($<BOS>$) token, with the sample number set at $N_S = 40,000$. These samples were subsequently assessed by the proprietary model $M_P$ to ascertain their probabilistic affinity for several domains, including \textit{law}, \textit{medicine}, \textit{finance}, \textit{science}, \textit{code}, and \textit{others}. To ensure the reliability of our statistical outcomes, the entire process was iterated 5 times, with the maximum number of iterations set at $T = 5$. The average knowledge distribution was then computed across these iterations. Empirically, with a dataset of $40K$ samples, the distribution of sequences generated by $M_\theta$ across domains demonstrated a high degree of consistency, with an overall variance not exceeding 1.874\%. The final domain knowledge distribution for each open-source model is depicted in the stacked bar chart presented in \Cref{fig:knowledge_detection_stacked_bar_chart}. 
The pre-existing domain knowledge distribution varies among different models. Therefore, it is essential to develop a data composition strategy that is tailored to the specific model being trained.

\subsection{Training Details}
\label{sec:appendix_training}

\textbf{Models and Implementation.} \quad 
All experiments were conducted based on full-parameter fine-tuning, during which we utilized a learning rate scheduler featuring linear warm-up and cosine decay, peaking at a learning rate of 2e\mbox{-}5, alongside a warmup ratio of 0.03, a weight decay of 0.0 and a batch size of 128 for 4 epochs. 
For scenarios aimed at \textit{fostering multi-ability}, we trained and assessed models including LLaMA-2-7B, LLaMA-3-8B, LLaMA-2-13B, Qwen-2-7B, Qwen-2.5-7B, Qwen-2.5-14B, and Qwen-2.5-32B. In the context of \textit{domain expansion}, the training and evaluations were performed using the Qwen-2.5-7B and Qwen-2.5-14B models. The total number of samples per epoch was set to $60k$, with each domain's samples being downsampled or upsampled according to the corresponding weights during the mixing process. Regarding \textit{\underline{reference models}}, for the LLaMA series, we used the Sheared-LLaMA-1.3B~\cite{xia2023sheared} as a lightweight reference model; as for the Qwen series, we utilized Qwen-2-1.5B and Qwen-2.5-1.5B as our reference models.


\textbf{Training Datasets.} \quad 
For training, we selected representative datasets for each domain, which exhibit significant differences in format, sentence length, and domain-specific content. These differences reflect the heterogeneity of training data across various domains during the fine-tuning stage. Further details about these datasets can be found in \Cref{tab:datasets}. 
Specifically, for the Alpaca dataset, which we utilize for representing the general domain, we have excluded data related to law, medicine, finance, science, and code domains to ensure the precision and authenticity of the actual domain weight.

\begin{table*}[ht]
    \centering
    \resizebox{\textwidth}{!}{
        \begin{tabular}{ccccc}
            \toprule
                Dataset & \# Instance & Source & \# Rounds & Full \\
            \midrule
                Lawyer-Instruct & 9241 & Reformatted from LawyerChat Dataset\footnote{https://huggingface.co/datasets/Alignment-Lab-AI/Lawyer-chat} & 1 & $\surd$ \\
                MedQA & 10178 & Professional Medical Board Exams & 1 & Training Portion \\
                Finance Alpaca & 68912 & Alpaca, FiQA, 1.3k Pairs Generated using GPT3.5 & 1 & $\surd$ \\
                Sonnet3.5 Science Conversations & 8835 & Scientific Conversations with Sonnet3.5 & 11.1 & $\surd$ \\
                Code Alpaca & 20022 & Generate Based on Self-Instruct~\cite{selfinstruct} & 1 & $\surd$ \\
                Alpaca & 49,087 & Generate Based on Self-Instruct~\cite{selfinstruct} & 1 & Excluding Samples of Other Domains \\
            \bottomrule
        \end{tabular}
    }
    \caption{\label{tab:datasets} 
    Details of the training datasets. ``Full'' indicates whether we utilize the entire data samples of the dataset. }  
\end{table*}

\subsubsection{Hyper-Parameters Setting}
\label{subsec:appendix_hyper_parameter}

During the multi-domain task fine-tuning of LLMs, we configured the number of training steps $T$ in \Cref{alg:multi_ability} and \Cref{alg:domain_expansion} to 4 epochs. We experimented with various magnitude of adjustment, specifically $\left[0.1, 0.3, 0.5, 0.8, 1.0 \right]$, and observed consistent weight ordering across domains, which far outperformed our baselines (detailed in \Cref{subsec:baselines}). Based on these experimental outcomes, we set the magnitude of adjustment $\sigma $ to $0.5$. 
Additionally, in the context of domain expansion, we set the increment for the target domain $\delta$ to $10\%$ per training step, considering the overall domain weight distribution across models. The variation threshold, denoted as $\varepsilon$, reflects the trade-off between enhancing specific domain skills and mitigating the loss of capabilities in non-target domains, where we assigned a weight of $1$.



\subsection{Evaluation Details}
\label{sec:appendix_evaluation}

We evaluate the performance of the models on downstream tasks across various domains, using two relevant benchmarks for each domain. 
Details of the datasets are provided in \Cref{tab:domain_benchmark}. 
Specifically, for the MedMCQA dataset, since the standard answers for the test set are not publicly available, we conducted our evaluations using the validation dataset. For the MMLU dataset, we selected 14 sub-tasks to construct the MMLU-Sci subset~\cite{zhang2024sciglm} for testing, aiming to ensure a robust and thorough evaluation.

\begin{table*}[ht]
    \centering
    \label{tab:domain_benchmark}
    \resizebox{\textwidth}{!}{
        \begin{tabular}{ccccc}
            \toprule
                Domain & Benchmark & \# Instance & Language & N-Shot \\ 
            \midrule
                \multirow{2}{*}{Law} & LegalBench~\cite{guha2024legalbench} & 90,394 (164 sub-tasks) & English & 1 \\ 
                  & LawBench~\cite{fei2023lawbench} &  10,000 (20 sub-tasks)  & Chinese & 1 \\ 
            \midrule
            \multirow{2}{*}{Medicine} & MedQA~\cite{jin2020disease} & 1,273 & English & 1 \\ 
                  & MedMCQA~\cite{pal2022medmcqa} & 4,183 & English & 1 \\ 
            \midrule
            \multirow{2}{*}{Finance} & FinEval~\cite{zhang2023fineval} & 4,661 (34 sub-tasks) & Chinese & 1 \\ 
                  & FinanceIQ~\cite{zhang2023xuanyuan} & 7,173 (10 sub-tasks) & Chinese & 5 \\ 
            \midrule
            \multirow{2}{*}{Science} & SciEval~\cite{sun2024scieval} & 15,901 & English & 1 \\ 
                  & MMLU-Sci~\cite{hendrycks2020measuring} & 2,999 (14 sub-tasks) & English & 0 \\ 
            \midrule
            \multirow{2}{*}{Code} & HumanEval~\cite{chen2021evaluating} & 164 & English & 0 \\ 
              & MBPP~\cite{austin2021program} & 974 & English & 0 \\ 
            \midrule
            \multirow{2}{*}{Other (General)} & AGIEval~\cite{zhong2023agieval} & 8,062 (20 sub-tasks) & English, Chinese & 0 \\ 
                  & HellaSwag~\cite{zellers2019hellaswag} & 10,003 & English & 0 \\ 
            \bottomrule
        \end{tabular}
    }
    \caption{\label{tab:domain_benchmark}
    Details of the benchmarks we employed for evaluation. ``N-Shot'' indicates that the model is given N example(s) to understand and perform the task. }
\end{table*}

\clearpage

\section{More Experiment Results}
\label{sec:appendix_experiment_result}
\subsection{Multi-Ability Fostering}
\label{subsec:appendix_multi_ability_results}

\begin{figure*}[ht]
    \begin{center}
    \centerline{\includegraphics[width=\linewidth]{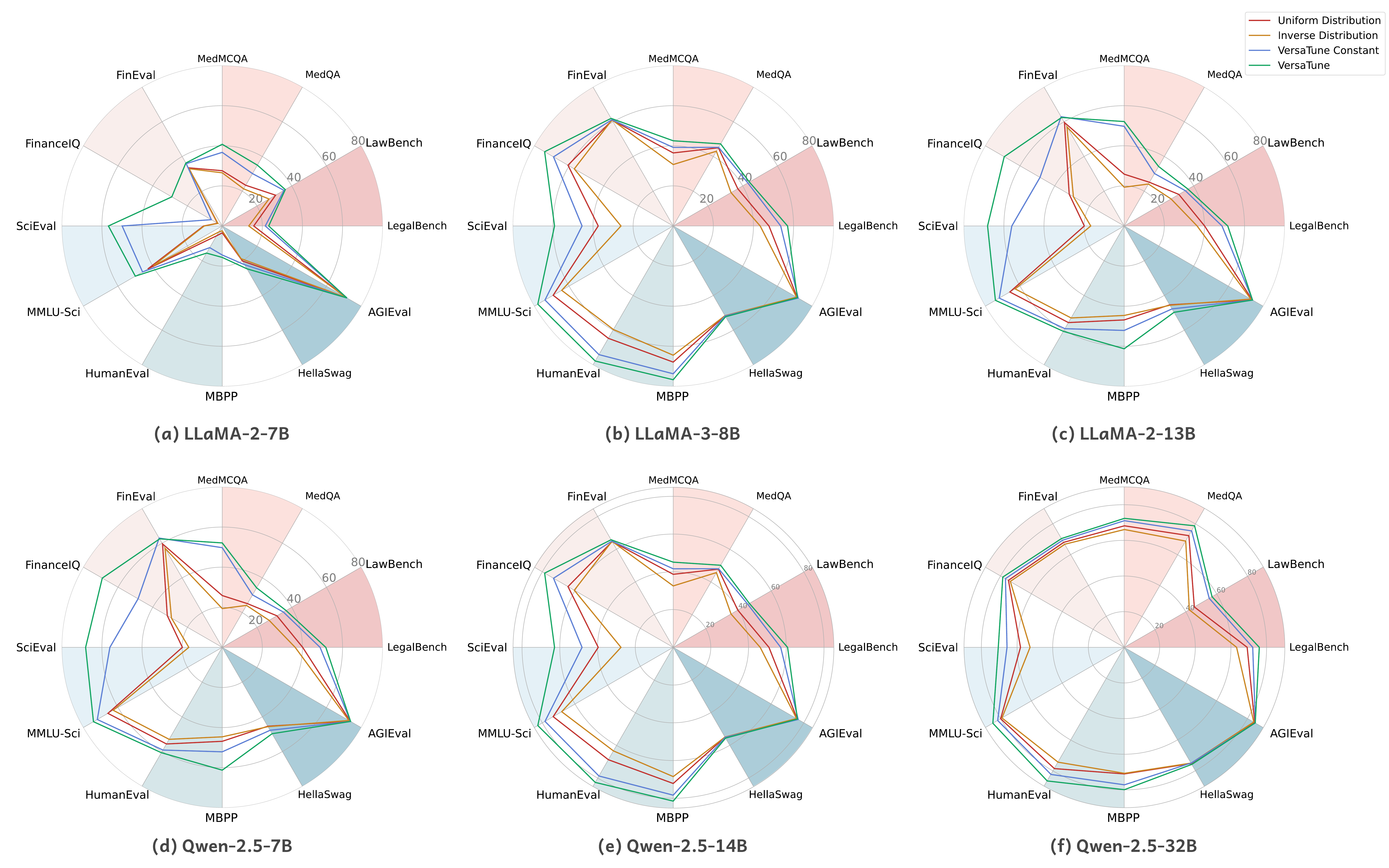}}
    \caption{
    \label{fig:domain_expansion_total_appendix}
    Performances of different models on versatile benchmarks related to various domains under the scenario of effective multi-ability fostering. The background color of the radar chart signifies the domain to which the current benchmark belongs, with reference to the color key provided in \Cref{fig:pipeline}, which includes law, medicine, finance, science, code, and general fields.}
    \end{center}
\end{figure*}

We present the results of the ablation study along with the raw scores from domain-specific benchmarks. 
As shown in \Cref{tab:multi_ability_total_appendix} and \Cref{fig:domain_expansion_total_appendix}, \colorbox[HTML]{DCE6FF}{\textbf{\ours constant}} is implemented with fixed domain weights derived from the knowledge distribution obtained by probing the target model $M_\theta$ prior to fine-tuning, where we ablate the components of dynamic adaptation in \Cref{alg:multi_ability} for an in-depth analysis. 
Additionally, to demonstrate the robustness of the dynamic adjustment more clearly, we compare the domain-level averaged performance of the \textit{\ours constant} and \textit{\ours} strategies, as depicted in \Cref{tab:multi_ability_avg_ablation}.

\begin{table*}[hb]
    \centering
    \resizebox{\textwidth}{!}{
        \begin{tabular}{cccccccccccccc}
            \toprule
                \multirow{2}{*}{Model} & \multirow{2}{*}{Method} & \multicolumn{2}{|c}{Law} & \multicolumn{2}{|c}{Medical} & \multicolumn{2}{|c}{Finance} & \multicolumn{2}{|c}{Science} & \multicolumn{2}{|c}{Code} & \multicolumn{2}{|c}{General} \\ 
            \cline{3-14}
             & & \multicolumn{1}{|c}{LegalBench} & LawBench & \multicolumn{1}{|c}{MedQA} & MedMCQA & \multicolumn{1}{|c}{FinEval} & FinanceIQ & \multicolumn{1}{|c}{SciEval} & MMLU-Sci & \multicolumn{1}{|c}{HumanEval} & MBPP & \multicolumn{1}{|c}{AGIEval} & \multicolumn{1}{c}{HellaSwag} \\
            \midrule
             \multirow{4}{*}{LLaMA-2-7B} & Uniform Distribution & \multicolumn{1}{|c}{15.71} & 30.72 & \multicolumn{1}{|c}{23.45} & 27.57 & \multicolumn{1}{|c}{33.50} & 2.71 & \multicolumn{1}{|c}{9.30} & 42.89 & \multicolumn{1}{|c}{5.67} & 3.44 & \multicolumn{1}{|c}{20.16} & 71.40 \\
              &  Inverse Distribution & \multicolumn{1}{|c}{13.23$^{\color{red} \downarrow}$} & 26.94$^{\color{red} \downarrow}$ & \multicolumn{1}{|c}{21.38$^{\color{red} \downarrow}$} & 26.52$^{\color{red} \downarrow }$ & \multicolumn{1}{|c}{32.96$^{\color{red} \downarrow }$} & 2.53$^{\color{red} \downarrow}$ & \multicolumn{1}{|c}{8.98$^{\color{red} \downarrow}$} & 39.67$^{\color{red} \downarrow }$ & \multicolumn{1}{|c}{3.47$^{\color{red} \downarrow}$} & 2.42$^{\color{red} \downarrow}$ & \multicolumn{1}{|c}{18.83$^{\color{red} \downarrow}$} & 71.33$^{\color{red} \downarrow }$ \\
               & \cellcolor[HTML]{DCE6FF}\ours Constant & \multicolumn{1}{|c}{\cellcolor[HTML]{DCE6FF}\underline{21.47}$^{\color{green} \uparrow}$} & \cellcolor[HTML]{DCE6FF}\underline{35.55}$^{\color{green} \uparrow}$ & \multicolumn{1}{|c}{\cellcolor[HTML]{DCE6FF}\underline{30.17}$^{\color{green} \uparrow}$} & \cellcolor[HTML]{DCE6FF}\underline{36.72}$^{\color{green} \uparrow}$ & \multicolumn{1}{|c}{\cellcolor[HTML]{DCE6FF}\underline{35.89}$^{\color{green} \uparrow}$} & \cellcolor[HTML]{DCE6FF}\underline{6.28}$^{\color{green} \uparrow}$ & \multicolumn{1}{|c}{\cellcolor[HTML]{DCE6FF}\underline{49.91}$^{\color{green} \uparrow}$} & \cellcolor[HTML]{DCE6FF}\underline{45.87}$^{\color{green} \uparrow}$ & \multicolumn{1}{|c}{\cellcolor[HTML]{DCE6FF}\underline{12.47}$^{\color{green} \uparrow}$} & \cellcolor[HTML]{DCE6FF}\underline{14.47}$^{\color{green} \uparrow}$ & \multicolumn{1}{|c}{\cellcolor[HTML]{DCE6FF}\underline{22.31}$^{\color{green} \uparrow}$} & \cellcolor[HTML]{DCE6FF}\textbf{71.89}$^{\color{green} \uparrow}$ \\
              & \ours & \multicolumn{1}{|c}{\textbf{23.18}$^{\color{green} \uparrow}$} & \textbf{36.31}$^{\color{green} \uparrow}$ & \multicolumn{1}{|c}{\textbf{35.04}$^{\color{green} \uparrow }$} & \textbf{40.75}$^{\color{green} \uparrow}$ & \multicolumn{1}{|c}{\textbf{36.27}$^{\color{green} \uparrow}$} & \textbf{29.04}$^{\color{green} \uparrow}$ & \multicolumn{1}{|c}{\textbf{56.75}$^{\color{green} \uparrow}$} & \textbf{50.06}$^{\color{green} \uparrow}$ & \multicolumn{1}{|c}{\textbf{15.62}$^{\color{green} \uparrow}$} & \textbf{15.68}$^{\color{green} \uparrow}$ & \multicolumn{1}{|c}{\textbf{24.67}$^{\color{green} \uparrow}$} & \underline{71.76}$^{\color{green} \uparrow}$ \\
        \midrule
        \multirow{4}{*}{Qwen-2-7B} & Uniform Distribution & \multicolumn{1}{|c}{39.05} & 31.99 & \multicolumn{1}{|c}{35.07} & 17.73 & \multicolumn{1}{|c}{59.49} & 14.62 & \multicolumn{1}{|c}{25.30} & 62.73 & \multicolumn{1}{|c}{53.26} & 37.82 & \multicolumn{1}{|c}{47.31} & \underline{73.60} \\
         &  Inverse Distribution & \multicolumn{1}{|c}{34.01$^{\color{red} \downarrow }$} & 27.81$^{\color{red} \downarrow }$ & \multicolumn{1}{|c}{23.90$^{\color{red} \downarrow }$} & 16.31$^{\color{red} \downarrow }$ & \multicolumn{1}{|c}{56.53$^{\color{red} \downarrow }$} & 11.30$^{\color{red} \downarrow }$ & \multicolumn{1}{|c}{18.57$^{\color{red} \downarrow }$} & 58.25$^{\color{red} \downarrow }$ & \multicolumn{1}{|c}{50.65$^{\color{red} \downarrow }$} & 33.63$^{\color{red} \downarrow }$ & \multicolumn{1}{|c}{45.74$^{\color{red} \downarrow }$} & 73.52$^{\color{red} \downarrow }$ \\
          &  \cellcolor[HTML]{DCE6FF}\ours Constant & \multicolumn{1}{|c}{\cellcolor[HTML]{DCE6FF}\underline{45.86}$^{\color{green} \uparrow }$} & \cellcolor[HTML]{DCE6FF}\underline{32.72}$^{\color{green} \uparrow }$ & \multicolumn{1}{|c}{\cellcolor[HTML]{DCE6FF}\underline{40.89}$^{\color{green} \uparrow }$} & \cellcolor[HTML]{DCE6FF}\underline{39.13}$^{\color{green} \uparrow }$ & \multicolumn{1}{|c}{\cellcolor[HTML]{DCE6FF}\underline{60.63}$^{\color{green} \uparrow }$} & \cellcolor[HTML]{DCE6FF}\underline{40.82}$^{\color{green} \uparrow }$ & \multicolumn{1}{|c}{\cellcolor[HTML]{DCE6FF}\underline{45.93}$^{\color{green} \uparrow }$} & \cellcolor[HTML]{DCE6FF}\underline{67.29}$^{\color{green} \uparrow }$ & \multicolumn{1}{|c}{\cellcolor[HTML]{DCE6FF}\underline{56.71}$^{\color{green} \uparrow }$} & \cellcolor[HTML]{DCE6FF}\underline{45.87}$^{\color{green} \uparrow }$ & \multicolumn{1}{|c}{\cellcolor[HTML]{DCE6FF}\textbf{48.16}$^{\color{green} \uparrow }$} & \cellcolor[HTML]{DCE6FF}72.98$^{\color{red} \downarrow }$ \\
          & \ours & \multicolumn{1}{|c}{\textbf{50.56}$^{\color{green} \uparrow }$} & \textbf{35.54}$^{\color{green} \uparrow }$ & \multicolumn{1}{|c}{\textbf{45.48}$^{\color{green} \uparrow }$} & \textbf{41.24}$^{\color{green} \uparrow }$ & \multicolumn{1}{|c}{\textbf{60.95}$^{\color{green} \uparrow }$} & \textbf{68.39}$^{\color{green} \uparrow }$ & \multicolumn{1}{|c}{\textbf{51.58}$^{\color{green} \uparrow }$} & \textbf{70.42}$^{\color{green} \uparrow }$ & \multicolumn{1}{|c}{\textbf{58.15}$^{\color{green} \uparrow }$} & \textbf{47.64}$^{\color{green} \uparrow }$ & \multicolumn{1}{|c}{\underline{48.02}$^{\color{green} \uparrow }$} & \textbf{73.67}$^{\color{green} \uparrow }$ \\
        \midrule
           \multirow{4}{*}{Qwen-2.5-7B} & Uniform Distribution & \multicolumn{1}{|c}{40.11} & 31.48 & \multicolumn{1}{|c}{25.17} & 25.84 & \multicolumn{1}{|c}{59.58} & 31.66 & \multicolumn{1}{|c}{19.88} & 65.84 & \multicolumn{1}{|c}{55.64} & 46.86 & \multicolumn{1}{|c}{45.42} & 73.69 \\
          &  Inverse Distribution & \multicolumn{1}{|c}{36.36$^{\color{red} \downarrow }$} & 26.98$^{\color{red} \downarrow }$  & \multicolumn{1}{|c}{24.16$^{\color{red} \downarrow }$} & 19.35$^{\color{red} \downarrow }$ & \multicolumn{1}{|c}{57.07$^{\color{red} \downarrow }$} & 29.25$^{\color{red} \downarrow }$ & \multicolumn{1}{|c}{16.68$^{\color{red} \downarrow }$} & 62.78$^{\color{red} \downarrow }$ & \multicolumn{1}{|c}{52.97$^{\color{red} \downarrow }$} & 44.63$^{\color{red} \downarrow }$ & \multicolumn{1}{|c}{45.67$^{\color{green} \uparrow }$} & 72.92$^{\color{red} \downarrow }$ \\
          &  \cellcolor[HTML]{DCE6FF}\ours Constant & \multicolumn{1}{|c}{\cellcolor[HTML]{DCE6FF}\underline{48.78}$^{\color{green} \uparrow }$} & \cellcolor[HTML]{DCE6FF}\underline{35.20}$^{\color{green} \uparrow }$ & \multicolumn{1}{|c}{\cellcolor[HTML]{DCE6FF}\underline{30.20}$^{\color{green} \uparrow }$} & \cellcolor[HTML]{DCE6FF}\underline{49.71}$^{\color{green} \uparrow }$ & \multicolumn{1}{|c}{\cellcolor[HTML]{DCE6FF}\textbf{62.94}$^{\color{green} \uparrow }$} & \cellcolor[HTML]{DCE6FF}\underline{48.47}$^{\color{green} \uparrow }$ & \multicolumn{1}{|c}{\cellcolor[HTML]{DCE6FF}\underline{56.04}$^{\color{green} \uparrow }$} & \cellcolor[HTML]{DCE6FF}\underline{71.96}$^{\color{green} \uparrow }$ & \multicolumn{1}{|c}{\cellcolor[HTML]{DCE6FF}\underline{59.15}$^{\color{green} \uparrow }$} & \cellcolor[HTML]{DCE6FF}\underline{52.10}$^{\color{green} \uparrow }$ & \multicolumn{1}{|c}{\cellcolor[HTML]{DCE6FF}\underline{47.75}$^{\color{green} \uparrow }$} & \cellcolor[HTML]{DCE6FF}\underline{73.88}$^{\color{green} \uparrow }$ \\
          & \ours & \multicolumn{1}{|c}{\textbf{51.65}$^{\color{green} \uparrow }$} & \textbf{36.75}$^{\color{green} \uparrow }$ & \multicolumn{1}{|c}{\textbf{34.28}$^{\color{green} \uparrow }$} & \textbf{52.09}$^{\color{green} \uparrow }$ & \multicolumn{1}{|c}{\underline{62.48}$^{\color{green} \uparrow }$} & \textbf{69.09}$^{\color{green} \uparrow }$ & \multicolumn{1}{|c}{\textbf{68.14}$^{\color{green} \uparrow }$} & \textbf{74.16}$^{\color{green} \uparrow }$ & \multicolumn{1}{|c}{\textbf{60.68}$^{\color{green} \uparrow }$} & \textbf{61.25}$^{\color{green} \uparrow }$ & \multicolumn{1}{|c}{\textbf{49.73}$^{\color{green} \uparrow }$} & \textbf{73.90}$^{\color{green} \uparrow }$ \\
          \midrule
          \multirow{4}{*}{LLaMA-3-8B} & Uniform Distribution & \multicolumn{1}{|c}{33.52} & 31.16 & \multicolumn{1}{|c}{31.03} & 10.26 & \multicolumn{1}{|c}{34.83} & 4.97 & \multicolumn{1}{|c}{6.51} & 50.17 & \multicolumn{1}{|c}{22.94} & 28.85 & \multicolumn{1}{|c}{23.87} & 73.26 \\
        &  Inverse Distribution & \multicolumn{1}{|c}{27.83$^{\color{red} \downarrow }$} & 27.48$^{\color{red} \downarrow }$ & \multicolumn{1}{|c}{25.51$^{\color{red} \downarrow }$} & 8.77$^{\color{red} \downarrow }$ & \multicolumn{1}{|c}{33.71$^{\color{red} \downarrow }$} & 3.31$^{\color{red} \downarrow }$ & \multicolumn{1}{|c}{6.09$^{\color{red} \downarrow }$} & 46.62$^{\color{red} \downarrow }$ & \multicolumn{1}{|c}{19.67$^{\color{red} \downarrow }$} & 24.34$^{\color{red} \downarrow }$ & \multicolumn{1}{|c}{23.45$^{\color{red} \downarrow }$} & 72.40$^{\color{red} \downarrow }$ \\
        & \cellcolor[HTML]{DCE6FF}\ours Constant & \multicolumn{1}{|c}{\cellcolor[HTML]{DCE6FF}\underline{47.85}$^{\color{green} \uparrow }$} & \cellcolor[HTML]{DCE6FF}\underline{37.75}$^{\color{green} \uparrow }$ & \multicolumn{1}{|c}{\cellcolor[HTML]{DCE6FF}\underline{37.33}$^{\color{green} \uparrow }$} & \cellcolor[HTML]{DCE6FF}\underline{30.15}$^{\color{green} \uparrow }$ & \multicolumn{1}{|c}{\cellcolor[HTML]{DCE6FF}\underline{37.93}$^{\color{green} \uparrow }$} & \cellcolor[HTML]{DCE6FF}\underline{25.27}$^{\color{green} \uparrow }$ & \multicolumn{1}{|c}{\cellcolor[HTML]{DCE6FF}\underline{54.77}$^{\color{green} \uparrow }$} & \cellcolor[HTML]{DCE6FF}\underline{56.04}$^{\color{green} \uparrow }$ & \multicolumn{1}{|c}{\cellcolor[HTML]{DCE6FF}\textbf{29.88}$^{\color{green} \uparrow }$} & \cellcolor[HTML]{DCE6FF}\underline{33.22}$^{\color{green} \uparrow }$ & \multicolumn{1}{|c}{\cellcolor[HTML]{DCE6FF}\underline{25.62}$^{\color{green} \uparrow }$} & \cellcolor[HTML]{DCE6FF}\underline{73.33}$^{\color{green} \uparrow }$ \\
        & \ours & \multicolumn{1}{|c}{\textbf{49.67}$^{\color{green} \uparrow }$} & \textbf{37.87}$^{\color{green} \uparrow }$ & \multicolumn{1}{|c}{\textbf{42.21}$^{\color{green} \uparrow }$} & \textbf{45.72}$^{\color{green} \uparrow }$ & \multicolumn{1}{|c}{\textbf{38.80}$^{\color{green} \uparrow }$} & \textbf{43.58}$^{\color{green} \uparrow }$ & \multicolumn{1}{|c}{\textbf{56.67}$^{\color{green} \uparrow }$} & \textbf{60.61}$^{\color{green} \uparrow }$ & \multicolumn{1}{|c}{\underline{28.91}$^{\color{green} \uparrow }$} & \textbf{35.65}$^{\color{green} \uparrow }$ & \multicolumn{1}{|c}{\textbf{28.78}$^{\color{green} \uparrow }$} & \textbf{73.62}$^{\color{green} \uparrow }$ \\
        \midrule
           \multirow{4}{*}{LLaMA-2-13B} & Uniform Distribution & \multicolumn{1}{|c}{47.66} & 34.85 & \multicolumn{1}{|c}{32.98} & 36.54 & \multicolumn{1}{|c}{37.54} & 32.85 & \multicolumn{1}{|c}{45.72} & 50.77 & \multicolumn{1}{|c}{36.54} & 38.55 & \multicolumn{1}{|c}{36.89} & 73.50 \\
          &  Inverse Distribution & \multicolumn{1}{|c}{40.12$^{\color{red} \downarrow }$} & 30.67$^{\color{red} \downarrow }$ & \multicolumn{1}{|c}{26.27$^{\color{red} \downarrow }$} & 28.78$^{\color{red} \downarrow }$ & \multicolumn{1}{|c}{36.67$^{\color{red} \downarrow }$} & 26.76$^{\color{red} \downarrow }$ & \multicolumn{1}{|c}{38.96$^{\color{red} \downarrow }$} & 48.68$^{\color{red} \downarrow }$ & \multicolumn{1}{|c}{28.78$^{\color{red} \downarrow }$} & 35.83$^{\color{red} \downarrow }$ & \multicolumn{1}{|c}{36.67$^{\color{red} \downarrow }$} & 73.11$^{\color{red} \downarrow }$ \\
          &  \cellcolor[HTML]{DCE6FF}\ours Constant & \multicolumn{1}{|c}{\cellcolor[HTML]{DCE6FF}\underline{53.79}$^{\color{green} \uparrow }$} & \cellcolor[HTML]{DCE6FF}\underline{38.73}$^{\color{green} \uparrow }$ & \multicolumn{1}{|c}{\cellcolor[HTML]{DCE6FF}\underline{40.69}$^{\color{green} \uparrow }$} & \cellcolor[HTML]{DCE6FF}\underline{42.74}$^{\color{green} \uparrow }$ & \multicolumn{1}{|c}{\cellcolor[HTML]{DCE6FF}\underline{39.33}$^{\color{green} \uparrow }$} & \cellcolor[HTML]{DCE6FF}\underline{38.47}$^{\color{green} \uparrow }$ & \multicolumn{1}{|c}{\cellcolor[HTML]{DCE6FF}\underline{57.13}$^{\color{green} \uparrow }$} & \cellcolor[HTML]{DCE6FF}\underline{55.10}$^{\color{green} \uparrow }$ & \multicolumn{1}{|c}{\cellcolor[HTML]{DCE6FF}\underline{42.74}$^{\color{green} \uparrow }$} & \cellcolor[HTML]{DCE6FF}\underline{42.76}$^{\color{green} \uparrow }$ & \multicolumn{1}{|c}{\cellcolor[HTML]{DCE6FF}\underline{37.91}$^{\color{green} \uparrow }$} & \cellcolor[HTML]{DCE6FF}\underline{74.27}$^{\color{green} \uparrow }$ \\
          & \ours & \multicolumn{1}{|c}{\textbf{55.87}$^{\color{green} \uparrow }$} & \textbf{40.14}$^{\color{green} \uparrow }$ & \multicolumn{1}{|c}{\textbf{45.78}$^{\color{green} \uparrow }$} & \textbf{47.67}$^{\color{green} \uparrow }$ & \multicolumn{1}{|c}{\textbf{39.48}$^{\color{green} \uparrow }$} & \textbf{55.12}$^{\color{green} \uparrow }$ & \multicolumn{1}{|c}{\textbf{63.87}$^{\color{green} \uparrow }$} & \textbf{62.84}$^{\color{green} \uparrow }$ & \multicolumn{1}{|c}{\textbf{47.67}$^{\color{green} \uparrow }$} & \textbf{44.62}$^{\color{green} \uparrow }$ & \multicolumn{1}{|c}{\textbf{39.64}$^{\color{green} \uparrow }$} & \textbf{74.63}$^{\color{green} \uparrow }$ \\
          \midrule
           \multirow{4}{*}{Qwen-2.5-14B} & Uniform Distribution & \multicolumn{1}{|c}{50.73} & 39.49 & \multicolumn{1}{|c}{47.85} & 38.71 & \multicolumn{1}{|c}{64.72} & 64.39 & \multicolumn{1}{|c}{39.74} & 73.45 & \multicolumn{1}{|c}{68.75} & 72.14 & \multicolumn{1}{|c}{54.92} & 75.88 \\
          &  Inverse Distribution & \multicolumn{1}{|c}{46.08$^{\color{red} \downarrow }$} & 35.36$^{\color{red} \downarrow }$ & \multicolumn{1}{|c}{45.75$^{\color{red} \downarrow }$} & 32.56$^{\color{red} \downarrow }$ & \multicolumn{1}{|c}{64.88$^{\color{green} \uparrow }$} & 60.53$^{\color{red} \downarrow }$ & \multicolumn{1}{|c}{27.68$^{\color{red} \downarrow }$} & 68.22$^{\color{red} \downarrow }$ & \multicolumn{1}{|c}{63.36$^{\color{red} \downarrow }$} & 68.49$^{\color{red} \downarrow }$ & \multicolumn{1}{|c}{54.87$^{\color{red} \downarrow }$} & 75.42$^{\color{red} \downarrow }$ \\
          &  \cellcolor[HTML]{DCE6FF}\ours Constant & \multicolumn{1}{|c}{\cellcolor[HTML]{DCE6FF}\underline{56.94}$^{\color{green} \uparrow }$} & \cellcolor[HTML]{DCE6FF}\underline{45.64}$^{\color{green} \uparrow }$ & \multicolumn{1}{|c}{\cellcolor[HTML]{DCE6FF}\underline{48.11}$^{\color{green} \uparrow }$} & \cellcolor[HTML]{DCE6FF}\underline{41.64}$^{\color{green} \uparrow }$ & \multicolumn{1}{|c}{\cellcolor[HTML]{DCE6FF}\underline{65.03}$^{\color{green} \uparrow }$} & \cellcolor[HTML]{DCE6FF}\underline{73.24}$^{\color{green} \uparrow }$ & \multicolumn{1}{|c}{\cellcolor[HTML]{DCE6FF}\underline{48.31}$^{\color{green} \uparrow }$} & \cellcolor[HTML]{DCE6FF}\underline{78.46}$^{\color{green} \uparrow }$ & \multicolumn{1}{|c}{\cellcolor[HTML]{DCE6FF}\underline{78.72}$^{\color{green} \uparrow }$} & \cellcolor[HTML]{DCE6FF}\underline{78.33}$^{\color{green} \uparrow }$ & \multicolumn{1}{|c}{\cellcolor[HTML]{DCE6FF}\underline{55.04}$^{\color{green} \uparrow }$} & \cellcolor[HTML]{DCE6FF}\textbf{76.45}$^{\color{green} \uparrow }$ \\
          & \ours & \multicolumn{1}{|c}{\textbf{60.59}$^{\color{green} \uparrow }$} & \textbf{46.58}$^{\color{green} \uparrow }$ & \multicolumn{1}{|c}{\textbf{50.24}$^{\color{green} \uparrow }$} & \textbf{45.15}$^{\color{green} \uparrow }$ & \multicolumn{1}{|c}{\textbf{65.84}$^{\color{green} \uparrow }$} & \textbf{78.68}$^{\color{green} \uparrow }$ & \multicolumn{1}{|c}{\textbf{62.89}$^{\color{green} \uparrow }$} & \textbf{82.86}$^{\color{green} \uparrow }$ & \multicolumn{1}{|c}{\textbf{82.64}$^{\color{green} \uparrow }$} & \textbf{81.48}$^{\color{green} \uparrow }$ & \multicolumn{1}{|c}{\textbf{55.52}$^{\color{green} \uparrow }$} & \underline{75.98}$^{\color{green} \uparrow }$ \\
          \midrule
          \multirow{4}{*}{Qwen-2.5-32B} & Uniform Distribution & \multicolumn{1}{|c}{68.86} & 45.28 & \multicolumn{1}{|c}{72.34} & 68.18 & \multicolumn{1}{|c}{68.03} & 75.14 & \multicolumn{1}{|c}{58.30} & 80.17 & \multicolumn{1}{|c}{78.59} & 71.04 & \multicolumn{1}{|c}{\underline{75.26}} & 84.40 \\
          &  Inverse Distribution & \multicolumn{1}{|c}{62.93$^{\color{red} \downarrow }$} & 42.05$^{\color{red} \downarrow }$ & \multicolumn{1}{|c}{68.80$^{\color{red} \downarrow }$} & 66.09$^{\color{red} \downarrow }$ & \multicolumn{1}{|c}{66.80$^{\color{red} \downarrow }$} & 73.93$^{\color{red} \downarrow }$ & \multicolumn{1}{|c}{52.94$^{\color{red} \downarrow }$} & 79.31$^{\color{red} \downarrow }$ & \multicolumn{1}{|c}{74.44$^{\color{red} \downarrow }$} & 70.71$^{\color{red} \downarrow }$ & \multicolumn{1}{|c}{75.00$^{\color{red} \downarrow }$} & 83.80$^{\color{red} \downarrow }$ \\
          &  \cellcolor[HTML]{DCE6FF}\ours Constant & \multicolumn{1}{|c}{\cellcolor[HTML]{DCE6FF}\underline{71.98}$^{\color{green} \uparrow }$} & \cellcolor[HTML]{DCE6FF}\underline{54.93}$^{\color{green} \uparrow }$ & \multicolumn{1}{|c}{\cellcolor[HTML]{DCE6FF}\underline{75.42}$^{\color{green} \uparrow }$} & \cellcolor[HTML]{DCE6FF}\underline{71.05}$^{\color{green} \uparrow }$ & \multicolumn{1}{|c}{\cellcolor[HTML]{DCE6FF}\underline{69.38}$^{\color{green} \uparrow }$} & \cellcolor[HTML]{DCE6FF}\underline{77.07}$^{\color{green} \uparrow }$ & \multicolumn{1}{|c}{\cellcolor[HTML]{DCE6FF}\underline{65.87}$^{\color{green} \uparrow }$} & \cellcolor[HTML]{DCE6FF}\underline{82.11}$^{\color{green} \uparrow }$ & \multicolumn{1}{|c}{\cellcolor[HTML]{DCE6FF}\underline{82.38}$^{\color{green} \uparrow }$} & \cellcolor[HTML]{DCE6FF}\underline{77.16}$^{\color{green} \uparrow }$ & \multicolumn{1}{|c}{\cellcolor[HTML]{DCE6FF}74.97$^{\color{red} \downarrow }$} & \cellcolor[HTML]{DCE6FF}\textbf{84.82}$^{\color{green} \uparrow }$ \\
          & \ours & \multicolumn{1}{|c}{\textbf{75.67}$^{\color{green} \uparrow }$} & \textbf{56.76}$^{\color{green} \uparrow }$ & \multicolumn{1}{|c}{\textbf{78.72}$^{\color{green} \uparrow }$} & \textbf{72.36}$^{\color{green} \uparrow }$ & \multicolumn{1}{|c}{\textbf{70.50}$^{\color{green} \uparrow }$} & \textbf{78.80}$^{\color{green} \uparrow }$ & \multicolumn{1}{|c}{\textbf{70.77}$^{\color{green} \uparrow }$} & \textbf{85.23}$^{\color{green} \uparrow }$ & \multicolumn{1}{|c}{\textbf{86.60}$^{\color{green} \uparrow }$} & \textbf{79.89}$^{\color{green} \uparrow }$ & \multicolumn{1}{|c}{\textbf{75.81}$^{\color{green} \uparrow }$} & \underline{84.75}$^{\color{green} \uparrow }$ \\
        \bottomrule
        \end{tabular}
    }
    \caption{\label{tab:multi_ability_total_appendix}
    Experimental results of \ours on multi-ability fostering, we compare the performances of several methods across different models. For each domain, we evaluate the models using two relevant benchmarks. The best and second best results are in \textbf{bold} and \underline{underlined}. Symbols ${\color{green} \uparrow}$ and ${\color{red} \downarrow}$ indicate an increase or decrease in downstream scores comparing to the \textit{uniform distribution} strategy.
} 
\end{table*}

\begin{table*}[ht]
\vskip -0.2in
    \centering
    \resizebox{0.81\textwidth}{!}{
        \begin{tabular}{ccccccccc}
            \toprule
                Model & Method & \multicolumn{1}{|c}{Law} & Medical & Finance & Science & Code & \multicolumn{1}{c|}{General} & Avg. \\
            \midrule
             \multirow{2}{*}{LLaMA-2-7B} & \ours Constant & \multicolumn{1}{|c}{28.51$^{\color{red} \downarrow }$} & 33.45$^{\color{red} \downarrow }$ & 21.09$^{\color{red} \downarrow }$ & 47.89$^{\color{red} \downarrow }$ & 13.47$^{\color{red} \downarrow }$ & 47.10$^{\color{red} \downarrow }$ & \multicolumn{1}{|c}{31.92$^{\color{red} \downarrow }$} \\
              & \ours & \multicolumn{1}{|c}{29.75} & 37.90 & 32.66& 53.41& 15.65& 48.22& \multicolumn{1}{|c}{36.27} \\
              \midrule
              \multirow{2}{*}{Qwen-2-7B} & \ours Constant & \multicolumn{1}{|c}{39.29$^{\color{red} \downarrow }$} & 40.01$^{\color{red} \downarrow }$ & 50.73$^{\color{red} \downarrow }$ & 56.61$^{\color{red} \downarrow }$ & 51.29$^{\color{red} \downarrow }$ & 60.57$^{\color{red} \downarrow }$ & \multicolumn{1}{|c}{49.75$^{\color{red} \downarrow }$} \\
              & \ours & \multicolumn{1}{|c}{43.05} & 43.36& 64.67& 61.00& 52.90& 60.85& \multicolumn{1}{|c}{54.31} \\
              \midrule
             \multirow{2}{*}{Qwen-2.5-7B} & \ours Constant & \multicolumn{1}{|c}{41.99$^{\color{red} \downarrow }$} & 39.96$^{\color{red} \downarrow }$ & 55.71$^{\color{red} \downarrow }$ & 64.00$^{\color{red} \downarrow }$ & 55.63$^{\color{red} \downarrow }$ & 60.82$^{\color{red} \downarrow }$ & \multicolumn{1}{|c}{53.02$^{\color{red} \downarrow }$} \\
              & \ours & \multicolumn{1}{|c}{44.20} & 43.19& 65.79& 71.15& 60.97& 61.82& \multicolumn{1}{|c}{57.85} \\
              \midrule
              \multirow{2}{*}{LLaMA-3-8B} & \ours Constant & \multicolumn{1}{|c}{42.80$^{\color{red} \downarrow }$} & 33.74$^{\color{red} \downarrow }$ & 31.60$^{\color{red} \downarrow }$ & 55.41$^{\color{red} \downarrow }$ & 31.55$^{\color{red} \downarrow }$ & 49.48$^{\color{red} \downarrow }$ & \multicolumn{1}{|c}{40.76$^{\color{red} \downarrow }$} \\
              & \ours & \multicolumn{1}{|c}{43.77} & 43.97& 41.19& 58.64& 32.28& 51.20& \multicolumn{1}{|c}{45.18} \\
              \midrule
              \multirow{2}{*}{LLaMA-2-13B} & \ours Constant & \multicolumn{1}{|c}{46.26$^{\color{red} \downarrow }$} & 41.72$^{\color{red} \downarrow }$ & 38.90$^{\color{red} \downarrow }$ & 56.12$^{\color{red} \downarrow }$ & 42.75$^{\color{red} \downarrow }$ & 56.09$^{\color{red} \downarrow }$ & \multicolumn{1}{|c}{46.97$^{\color{red} \downarrow }$} \\
              & \ours & \multicolumn{1}{|c}{48.01} & 46.73 & 47.30 & 63.36 & 46.15 & 57.14 & \multicolumn{1}{|c}{51.45} \\
              \midrule
              \multirow{2}{*}{Qwen-2.5-14B} & \ours Constant & \multicolumn{1}{|c}{51.29$^{\color{red} \downarrow }$} & 44.88$^{\color{red} \downarrow }$ & 34.14$^{\color{red} \downarrow }$ & 63.39$^{\color{red} \downarrow }$ & 78.53$^{\color{red} \downarrow }$ & 65.75$^{\color{red} \downarrow }$ & \multicolumn{1}{|c}{56.33$^{\color{red} \downarrow }$} \\
              & \ours & \multicolumn{1}{|c}{53.59} & 47.70 & 72.26 & 72.88 & 82.06 & 65.75 & \multicolumn{1}{|c}{65.71} \\
              \midrule
              \multirow{2}{*}{Qwen-2.5-32B} & \ours Constant & \multicolumn{1}{|c}{63.46$^{\color{red} \downarrow }$} & 73.24$^{\color{red} \downarrow }$ & 73.23$^{\color{red} \downarrow }$ & 73.99$^{\color{red} \downarrow }$ & 79.77$^{\color{red} \downarrow }$ & 79.90$^{\color{red} \downarrow }$ & \multicolumn{1}{|c}{73.93$^{\color{red} \downarrow }$} \\
              & \ours & \multicolumn{1}{|c}{66.22} & 75.54 & 74.65 & 78.00 & 83.25 & 80.28 & \multicolumn{1}{|c}{76.32} \\
            \bottomrule
        \end{tabular}
    }
    \caption{\label{tab:multi_ability_avg_ablation} 
    Ablation studies on multi-ability fostering, we compare the performances of \textit{\ours} and \textit{\ours Constant} across different models. 
    The domain performance scores were calculated as the \underline{arithmetic mean} of the respective benchmark scores obtained for each domain. 
    ``Avg'' denotes the average performance across all domain-specific tasks. ${\color{green} \uparrow}$ and ${\color{red} \downarrow}$ indicate an increase or decrease in scores comparing to the \textit{\ours} strategy.}
    \vskip -0.2in
\end{table*}

\clearpage

\begin{wrapfigure}{r}{.6\textwidth}
    \centering
    \vspace{-1.5em}
    \includegraphics[width=\linewidth]{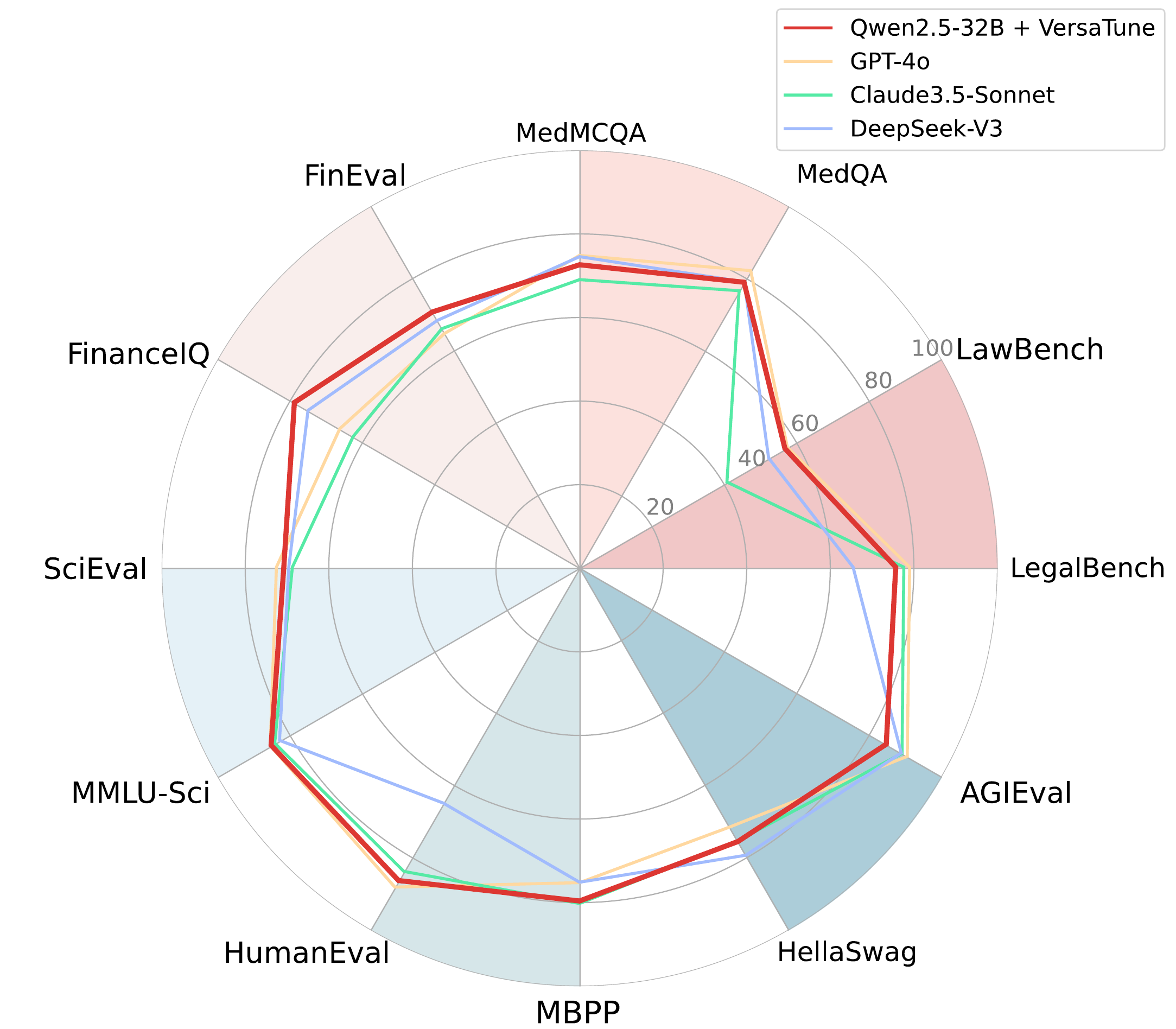}
    \caption{\label{fig:sota_comparison_radar}
    Performance comparison between \textit{Qwen-2.5-32B + \ours} and \textit{frontier models} acorss multiple domains.}
    \vspace{-2em}
\end{wrapfigure}

\textbf{Comparison with Frontier Models.}  
Furthermore, to demonstrate the efficacy of \ours across diverse domain tasks, we conducted a comparative analysis between \textit{Qwen-2.5-32B + VersaTune} and \textit{frontier models} across various domain-specific tasks, with results visualized in \Cref{fig:sota_comparison_radar}. Such experimental results indicate that Qwen-2.5-32B equipped with \ours enhances multi-domain performance to the state-of-the-art levels, which even outperforms frontier models like GPT-4o, Claude3.5-Sonnet and DeepSeek-V3 by 0.86\%, 4.76\% and 4.60\% on the overall domain capabilities, respectively. This comparison underscores the superior performance of our \ours in advancing multi-domain capabilities.

\clearpage

\subsection{Flexible Domain Expansion}
\label{subsec:appendix_domain_expansion_results}

\begin{table*}[ht]
\centering
\resizebox{\linewidth}{!}{
\begin{tabular}{ccccccccccc}
\toprule
\multirow{2}{*}{\begin{tabular}[c]{@{}c@{}}Target Domain\end{tabular}} & \multicolumn{1}{|c|}{Training Step} & \multirow{2}{*}{Method} & \multicolumn{6}{|c|}{Variations in Comprehensive Domains ($\%$)} & \multicolumn{2}{c}{Sum. ($\%$)} \\
\cline{4-11}
   & \multicolumn{1}{|c|}{(Epoch)} & & \multicolumn{1}{|c}{Law}  & Medicine  & Finance  & Science  & Code  & \multicolumn{1}{c|}{Other} & \multicolumn{1}{c|}{Target} & Non-Target \\
\midrule
\multirow{12}{*}{Law} & \multicolumn{1}{|c|}{ } & \multicolumn{1}{|c|}{100\% Specific Domain} & \cellcolor[HTML]{E59191}$\color{green} \uparrow $5.89 & $\color{red} \downarrow $18.82 & $\color{green} \uparrow $14.71 & $\color{red} \downarrow $11.76 & $\color{red} \downarrow $11.18 & $\color{red} \downarrow $5.00 & \multicolumn{1}{|c}{$\color{green} \uparrow $5.89} & \multicolumn{1}{|c}{$\color{red} \downarrow $32.05} \\
 & \multicolumn{1}{|c|}{1} & \multicolumn{1}{|c|}{Uniform Distribution of Non-Target Domains} & \cellcolor[HTML]{E59191}$\color{green} \uparrow $5.38 & $\color{red} \downarrow $7.35 & $\color{green} \uparrow $6.47 & $\color{red} \downarrow $6.58 & $\color{red} \downarrow $0.05 & $\color{green} \uparrow $13.99 & \multicolumn{1}{|c}{$\color{green} \uparrow $5.38} & \multicolumn{1}{|c}{$\color{red} \downarrow $6.48} \\
  & \multicolumn{1}{|c|}{ } & \multicolumn{1}{|c|}{\ours} & \cellcolor[HTML]{E59191}$\color{green} \uparrow $8.25 & $\color{green} \uparrow $6.18 & $\color{green} \uparrow $12.06 & $\color{red} \downarrow $7.65 & $\color{red} \downarrow $6.22 & $\color{green} \uparrow $17.59 & \multicolumn{1}{|c}{$\color{green} \uparrow $8.25} & \multicolumn{1}{|c}{$\color{green} \uparrow $21.96} \\
  \cline{2-11}
   & \multicolumn{1}{|c|}{ } & \multicolumn{1}{|c|}{100\% Specific Domain} & \cellcolor[HTML]{E59191}$\color{green} \uparrow $35.51 & $\color{red} \downarrow $12.65 & $\color{green} \uparrow $30.59 & $\color{red} \downarrow $4.41 & $\color{red} \downarrow $5.29 & $\color{red} \downarrow $11.76 & \multicolumn{1}{|c}{$\color{green} \uparrow $35.51} & \multicolumn{1}{|c}{$\color{red} \downarrow $3.52} \\
   & \multicolumn{1}{|c|}{2} & \multicolumn{1}{|c|}{Uniform Distribution of Non-Target Domains} & \cellcolor[HTML]{E59191}$\color{green} \uparrow $33.89 & $\color{red} \downarrow $7.94 & $\color{green} \uparrow $26.47 & $\color{red} \downarrow $12.65 & $\color{red} \downarrow $7.65 & $\color{green} \uparrow $5.00 & \multicolumn{1}{|c}{$\color{green} \uparrow $33.89} & \multicolumn{1}{|c}{$\color{green} \uparrow $3.23} \\
  & \multicolumn{1}{|c|}{ } & \multicolumn{1}{|c|}{\ours} & \cellcolor[HTML]{E59191}$\color{green} \uparrow $35.14 & $\color{green} \uparrow $3.53 & $\color{green} \uparrow $18.82 & $\color{red} \downarrow $6.89 & $\color{green} \uparrow $17.06 & $\color{green} \uparrow $13.24 & \multicolumn{1}{|c}{$\color{green} \uparrow $35.14} & \multicolumn{1}{|c}{$\color{green} \uparrow $45.76} \\
  \cline{2-11}
   & \multicolumn{1}{|c|}{ } & \multicolumn{1}{|c|}{100\% Specific Domain} & \cellcolor[HTML]{E59191}$\color{green} \uparrow $55.84 & $\color{red} \downarrow $17.94 & $\color{green} \uparrow $12.35 & $\color{red} \downarrow $8.82 & $\color{red} \downarrow $23.53 & $\color{red} \downarrow$5.00 & \multicolumn{1}{|c}{$\color{green} \uparrow $55.84} & \multicolumn{1}{|c}{$\color{red} \downarrow $42.94} \\
    & \multicolumn{1}{|c|}{3} & \multicolumn{1}{|c|}{Uniform Distribution of Non-Target Domains} & \cellcolor[HTML]{E59191}$\color{green} \uparrow $52.89 & $\color{red} \downarrow $8.24 & $\color{green} \uparrow $15.00 & $\color{red} \downarrow $12.06 & $\color{red} \downarrow $10.17 & $\color{red} \downarrow $5.12 & \multicolumn{1}{|c}{$\color{green} \uparrow $52.89} & \multicolumn{1}{|c}{$\color{red} \downarrow $20.59} \\
  & \multicolumn{1}{|c|}{ } & \multicolumn{1}{|c|}{\ours} & \cellcolor[HTML]{E59191}$\color{green} \uparrow $51.71 & $\color{red} \downarrow $4.41 & $\color{green} \uparrow $24.71 & $\color{red} \downarrow $5.29 & $\color{green} \uparrow $4.87 & $\color{green} \uparrow $8.42 & \multicolumn{1}{|c}{$\color{green} \uparrow $51.71} & \multicolumn{1}{|c}{$\color{green} \uparrow $29.30} \\
  \cline{2-11}
   & \multicolumn{1}{|c|}{ } & \multicolumn{1}{|c|}{100\% Specific Domain} & \cellcolor[HTML]{E59191}$\color{green} \uparrow $62.76 & $\color{red} \downarrow $5.00 & $\color{red} \downarrow $2.06 & $\color{red} \downarrow $31.18 & $\color{red} \downarrow $21.76 & $\color{red} \downarrow $12.65 & \multicolumn{1}{|c}{$\color{green} \uparrow $62.76} & \multicolumn{1}{|c}{$\color{red} \downarrow $72.65} \\
    & \multicolumn{1}{|c|}{4} & \multicolumn{1}{|c|}{Uniform Distribution of Non-Target Domains} & \cellcolor[HTML]{E59191}$\color{green} \uparrow $58.12 & $\color{red} \downarrow $9.71 & $\color{green} \uparrow $5.59 & $\color{red} \downarrow $9.41 & $\color{red} \downarrow $10.14 & $\color{red} \downarrow $12.05 & \multicolumn{1}{|c}{$\color{green} \uparrow $58.12} & \multicolumn{1}{|c}{$\color{red} \downarrow $35.72} \\
  & \multicolumn{1}{|c|}{ } & \multicolumn{1}{|c|}{\ours} & \cellcolor[HTML]{E59191}$\color{green} \uparrow $59.08 & $\color{red} \downarrow $5.59 & $\color{green} \uparrow $13.82 & $\color{red} \downarrow $8.82 & $\color{red} \downarrow $5.61 & $\color{red} \downarrow $6.17 & \multicolumn{1}{|c}{$\color{green} \uparrow $59.08} & \multicolumn{1}{|c}{$\color{red} \downarrow $12.37} \\
\midrule
\multirow{12}{*}{Medicine} & \multicolumn{1}{|c|}{ } & \multicolumn{1}{|c|}{100\% Specific Domain} & $\color{red} \downarrow $10.29 & \cellcolor[HTML]{FAC5BD}$\color{green} \uparrow $5.87 & $\color{red} \downarrow $3.82 & $\color{green} \uparrow $24.12 & $\color{red} \downarrow $19.41 & $\color{red} \downarrow $7.35 & \multicolumn{1}{|c}{$\color{green} \uparrow $5.87} & \multicolumn{1}{|c}{$\color{red} \downarrow $16.75} \\
& \multicolumn{1}{|c|}{1} & \multicolumn{1}{|c|}{Uniform Distribution of Non-Target Domains} & $\color{red} \downarrow $3.82 & \cellcolor[HTML]{FAC5BD}$\color{green} \uparrow $5.36 & $\color{red} \downarrow $5.29 & $\color{green} \uparrow $5.59 & $\color{red} \downarrow $2.65 & $\color{green} \uparrow $6.53 & \multicolumn{1}{|c}{$\color{green} \uparrow $5.36} & \multicolumn{1}{|c}{$\color{green} \uparrow $0.36} \\
  & \multicolumn{1}{|c|}{ } & \multicolumn{1}{|c|}{\ours} & $\color{green} \uparrow $3.53 & \cellcolor[HTML]{FAC5BD}$\color{green} \uparrow $8.17 & $\color{red} \downarrow $7.65 & $\color{green} \uparrow $8.82 & $\color{green} \uparrow $16.47 & $\color{red} \downarrow $6.17 & \multicolumn{1}{|c}{$\color{green} \uparrow $8.17} & \multicolumn{1}{|c}{$\color{green} \uparrow $15.00} \\
\cline{2-11}
   & \multicolumn{1}{|c|}{ } & \multicolumn{1}{|c|}{100\% Specific Domain} & $\color{red} \downarrow $18.82 & \cellcolor[HTML]{FAC5BD}$\color{green} \uparrow $40.44 & $\color{red} \downarrow $6.47 & $\color{green} \uparrow $40.00 & $\color{red} \downarrow $7.94 & $\color{red} \downarrow $17.65 & \multicolumn{1}{|c}{$\color{green} \uparrow $40.44} & \multicolumn{1}{|c}{$\color{red} \downarrow $10.88} \\
   & \multicolumn{1}{|c|}{2} & \multicolumn{1}{|c|}{Uniform Distribution of Non-Target Domains} & $\color{red} \downarrow $7.94 & \cellcolor[HTML]{FAC5BD}$\color{green} \uparrow $33.68 & $\color{red} \downarrow $9.12 & $\color{green} \uparrow $20.59 & $\color{green} \uparrow $4.98 & $\color{red} \downarrow $5.64 & \multicolumn{1}{|c}{$\color{green} \uparrow $33.68} & \multicolumn{1}{|c}{$\color{green} \uparrow $2.87} \\
  & \multicolumn{1}{|c|}{ } & \multicolumn{1}{|c|}{\ours} & $\color{green} \uparrow $12.35 & \cellcolor[HTML]{FAC5BD}$\color{green} \uparrow $35.21 & $\color{green} \uparrow $7.65 & $\color{green} \uparrow $8.84 & $\color{green} \uparrow $16.52 & $\color{green} \uparrow $12.65 & \multicolumn{1}{|c}{$\color{green} \uparrow $35.21} & \multicolumn{1}{|c}{$\color{green} \uparrow $58.01} \\
\cline{2-11}
& \multicolumn{1}{|c|}{ } & \multicolumn{1}{|c|}{100\% Specific Domain} & $\color{red} \downarrow $22.35 & \cellcolor[HTML]{FAC5BD}$\color{green} \uparrow $58.78 & $\color{red} \downarrow $8.82 & $\color{green} \uparrow $7.94 & $\color{red} \downarrow $19.12 & $\color{red} \downarrow $10.00 & \multicolumn{1}{|c}{$\color{green} \uparrow $58.78} & \multicolumn{1}{|c}{$\color{red} \downarrow $52.35} \\
   & \multicolumn{1}{|c|}{3} & \multicolumn{1}{|c|}{Uniform Distribution of Non-Target Domains} & $\color{red} \downarrow $10.46 & \cellcolor[HTML]{FAC5BD}$\color{green} \uparrow $55.69 & $\color{red} \downarrow $14.98 & $\color{green} \uparrow $12.35 & $\color{red} \downarrow $4.41 & $\color{red} \downarrow $14.18 & \multicolumn{1}{|c}{$\color{green} \uparrow $55.69} & \multicolumn{1}{|c}{$\color{red} \downarrow $31.68} \\
  & \multicolumn{1}{|c|}{ } & \multicolumn{1}{|c|}{\ours} & $\color{red} \downarrow $3.53 & \cellcolor[HTML]{FAC5BD}$\color{green} \uparrow $53.85 & $\color{red} \downarrow $4.52 & $\color{green} \uparrow $17.06 & $\color{green} \uparrow $7.64 & $\color{green} \uparrow $6.03 & \multicolumn{1}{|c}{$\color{green} \uparrow $53.85} & \multicolumn{1}{|c}{$\color{green} \uparrow $22.68} \\
\cline{2-11}
& \multicolumn{1}{|c|}{ } & \multicolumn{1}{|c|}{100\% Specific Domain} & $\color{red} \downarrow $27.94 & \cellcolor[HTML]{FAC5BD}$\color{green} \uparrow $64.61 & $\color{red} \downarrow $11.76 & $\color{red} \downarrow $2.35 & $\color{red} \downarrow $21.76 & $\color{red} \downarrow $12.65 & \multicolumn{1}{|c}{$\color{green} \uparrow $64.61} & \multicolumn{1}{|c}{$\color{red} \downarrow $76.46} \\
   & \multicolumn{1}{|c|}{4} & \multicolumn{1}{|c|}{Uniform Distribution of Non-Target Domains} & $\color{red} \downarrow $13.82 & \cellcolor[HTML]{FAC5BD}$\color{green} \uparrow $58.07 & $\color{red} \downarrow $11.18 & $\color{green} \uparrow $2.35 & $\color{red} \downarrow $6.47 & $\color{red} \downarrow $18.53 & \multicolumn{1}{|c}{$\color{green} \uparrow $58.07} & \multicolumn{1}{|c}{$\color{red} \downarrow $47.65} \\
  & \multicolumn{1}{|c|}{ } & \multicolumn{1}{|c|}{\ours} & $\color{red} \downarrow $5.59 & \cellcolor[HTML]{FAC5BD}$\color{green} \uparrow $59.81 & $\color{red} \downarrow $4.27 & $\color{green} \uparrow $10.68 & $\color{red} \downarrow $5.94 & $\color{red} \downarrow $10.26 & \multicolumn{1}{|c}{$\color{green} \uparrow $59.81} & \multicolumn{1}{|c}{$\color{red} \downarrow $15.38} \\
\midrule
\multirow{12}{*}{Finance} & \multicolumn{1}{|c|}{ } & \multicolumn{1}{|c|}{100\% Specific Domain} & $\color{green} \uparrow $20.59 & $\color{red} \downarrow $7.94 & \cellcolor[HTML]{F4DFDA}$\color{green} \uparrow $5.45 & $\color{red} \downarrow $10.29 & $\color{red} \downarrow $12.65 & $\color{red} \downarrow $6.47 & \multicolumn{1}{|c}{$\color{green} \uparrow $5.45} & \multicolumn{1}{|c}{$\color{red} \downarrow $16.76} \\
& \multicolumn{1}{|c|}{1} & \multicolumn{1}{|c|}{Uniform Distribution of Non-Target Domains} & $\color{green} \uparrow $12.05 & $\color{red} \downarrow $7.36 & \cellcolor[HTML]{F4DFDA}$\color{green} \uparrow $8.21 & $\color{red} \downarrow $6.74 & $\color{green} \uparrow $19.41 & $\color{red} \downarrow $8.53 & \multicolumn{1}{|c}{$\color{green} \uparrow $8.21} & \multicolumn{1}{|c}{$\color{green} \uparrow $8.83} \\
  & \multicolumn{1}{|c|}{ } & \multicolumn{1}{|c|}{\ours} & $\color{green} \uparrow $15.07 & $\color{red} \downarrow $4.12 & \cellcolor[HTML]{F4DFDA}$\color{green} \uparrow $10.46 & $\color{green} \uparrow $3.24 & $\color{green} \uparrow $17.06 & $\color{red} \downarrow $8.55 & \multicolumn{1}{|c}{$\color{green} \uparrow $10.46} & \multicolumn{1}{|c}{$\color{green} \uparrow $22.70} \\
\cline{2-11}
   & \multicolumn{1}{|c|}{ } & \multicolumn{1}{|c|}{100\% Specific Domain} & $\color{green} \uparrow $18.24 & $\color{red} \downarrow $9.71 & \cellcolor[HTML]{F4DFDA}$\color{green} \uparrow $34.97 & $\color{red} \downarrow $9.41 & $\color{green} \uparrow $5.29 & $\color{red} \downarrow $8.82 & \multicolumn{1}{|c}{$\color{green} \uparrow $34.97} & \multicolumn{1}{|c}{$\color{red} \downarrow $4.41} \\
   & \multicolumn{1}{|c|}{2} & \multicolumn{1}{|c|}{Uniform Distribution of Non-Target Domains} & $\color{green} \uparrow $20.59 & $\color{red} \downarrow $7.94 & \cellcolor[HTML]{F4DFDA}$\color{green} \uparrow $31.08 & $\color{red} \downarrow $4.71 & $\color{red} \downarrow $7.69 & $\color{green} \uparrow $4.98 & \multicolumn{1}{|c}{$\color{green} \uparrow $31.08} & \multicolumn{1}{|c}{$\color{green} \uparrow $5.23} \\
  & \multicolumn{1}{|c|}{ } & \multicolumn{1}{|c|}{\ours} & $\color{green} \uparrow $24.70 & $\color{green} \uparrow $7.35 & \cellcolor[HTML]{F4DFDA}$\color{green} \uparrow $33.92 & $\color{red} \downarrow $5.02 & $\color{green} \uparrow $7.08 & $\color{green} \uparrow $10.58 & \multicolumn{1}{|c}{$\color{green} \uparrow $33.92} & \multicolumn{1}{|c}{$\color{green} \uparrow $44.69} \\
\cline{2-11}
& \multicolumn{1}{|c|}{ } & \multicolumn{1}{|c|}{100\% Specific Domain} & $\color{green} \uparrow $23.53 & $\color{red} \downarrow $9.41 & \cellcolor[HTML]{F4DFDA}$\color{green} \uparrow $55.87 & $\color{red} \downarrow $17.35 & $\color{red} \downarrow $14.71 & $\color{red} \downarrow $7.94 & \multicolumn{1}{|c}{$\color{green} \uparrow $55.87} & \multicolumn{1}{|c}{$\color{red} \downarrow $25.88} \\
   & \multicolumn{1}{|c|}{3} & \multicolumn{1}{|c|}{Uniform Distribution of Non-Target Domains} & $\color{green} \uparrow $15.02 & $\color{red} \downarrow $8.24 & \cellcolor[HTML]{F4DFDA}$\color{green} \uparrow $52.41 & $\color{red} \downarrow $12.06 & $\color{red} \downarrow $10.30 & $\color{red} \downarrow $4.98 & \multicolumn{1}{|c}{$\color{green} \uparrow $52.41} & \multicolumn{1}{|c}{$\color{red} \downarrow $20.56} \\
  & \multicolumn{1}{|c|}{ } & \multicolumn{1}{|c|}{\ours} & $\color{green} \uparrow $24.71 & $\color{red} \downarrow $11.18 & \cellcolor[HTML]{F4DFDA}$\color{green} \uparrow $53.04 & $\color{green} \uparrow $5.29 & $\color{green} \uparrow $4.44 & $\color{green} \uparrow $8.83 & \multicolumn{1}{|c}{$\color{green} \uparrow $53.04} & \multicolumn{1}{|c}{$\color{green} \uparrow $32.09} \\
\cline{2-11}
 & \multicolumn{1}{|c|}{ } & \multicolumn{1}{|c|}{100\% Specific Domain} & $\color{green} \uparrow $5.00 & $\color{red} \downarrow $9.12 & \cellcolor[HTML]{F4DFDA}$\color{green} \uparrow $62.89 & $\color{red} \downarrow $20.29 & $\color{red} \downarrow $12.94 & $\color{red} \downarrow $21.76 & \multicolumn{1}{|c}{$\color{green} \uparrow $62.89} & \multicolumn{1}{|c}{$\color{red} \downarrow $59.11} \\
   & \multicolumn{1}{|c|}{4} & \multicolumn{1}{|c|}{Uniform Distribution of Non-Target Domains} & $\color{green} \uparrow $5.88 & $\color{red} \downarrow $5.29 & \cellcolor[HTML]{F4DFDA}$\color{green} \uparrow $56.13 & $\color{red} \downarrow $14.09 & $\color{red} \downarrow $13.23 & $\color{red} \downarrow $20.87 & \multicolumn{1}{|c}{$\color{green} \uparrow $56.13} & \multicolumn{1}{|c}{$\color{red} \downarrow $47.60} \\
  & \multicolumn{1}{|c|}{ } & \multicolumn{1}{|c|}{\ours} & $\color{green} \uparrow $14.19 & $\color{red} \downarrow $8.24 & \cellcolor[HTML]{F4DFDA}$\color{green} \uparrow $58.47 & $\color{red} \downarrow $13.24 & $\color{red} \downarrow $8.52 & $\color{red} \downarrow $9.41 & \multicolumn{1}{|c}{$\color{green} \uparrow $58.47} & \multicolumn{1}{|c}{$\color{red} \downarrow $25.22} \\
\midrule
\multirow{12}{*}{Science} & \multicolumn{1}{|c|}{ } & \multicolumn{1}{|c|}{100\% Specific Domain} & $\color{red} \downarrow $10.29 & $\color{green} \uparrow $17.06 & $\color{red} \downarrow $3.82 & \cellcolor[HTML]{CCE4F0}$\color{green} \uparrow $6.78 & $\color{red} \downarrow $4.71 & $\color{red} \downarrow $7.35 & \multicolumn{1}{|c}{$\color{green} \uparrow $6.78} & \multicolumn{1}{|c}{$\color{red} \downarrow $9.11} \\
& \multicolumn{1}{|c|}{1} & \multicolumn{1}{|c|}{Uniform Distribution of Non-Target Domains} & $\color{red} \downarrow $6.76 & $\color{green} \uparrow $17.64 & $\color{red} \downarrow $9.18 & \cellcolor[HTML]{CCE4F0}$\color{green} \uparrow $5.37 & $\color{green} \uparrow $7.05 & $\color{green} \uparrow $4.73 & \multicolumn{1}{|c}{$\color{green} \uparrow $5.37} & \multicolumn{1}{|c}{$\color{green} \uparrow $13.48} \\
  & \multicolumn{1}{|c|}{ } & \multicolumn{1}{|c|}{\ours} & $\color{red} \downarrow $3.53 & $\color{green} \uparrow $17.35 & $\color{red} \downarrow $7.64 & \cellcolor[HTML]{CCE4F0}$\color{green} \uparrow $8.35 & $\color{green} \uparrow $16.67 & $\color{green} \uparrow $6.98 & \multicolumn{1}{|c}{$\color{green} \uparrow $8.35} & \multicolumn{1}{|c}{$\color{green} \uparrow $29.83} \\
\cline{2-11}
   & \multicolumn{1}{|c|}{ } & \multicolumn{1}{|c|}{100\% Specific Domain} & $\color{red} \downarrow $11.47 & $\color{green} \uparrow $12.35 & $\color{red} \downarrow $4.71 & \cellcolor[HTML]{CCE4F0}$\color{green} \uparrow $40.84 & $\color{red} \downarrow $5.88 & $\color{red} \downarrow $12.94 & \multicolumn{1}{|c}{$\color{green} \uparrow $40.84} & \multicolumn{1}{|c}{$\color{red} \downarrow $22.65} \\
   & \multicolumn{1}{|c|}{2} & \multicolumn{1}{|c|}{Uniform Distribution of Non-Target Domains} & $\color{red} \downarrow $8.24 & $\color{green} \uparrow $20.59 & $\color{green} \uparrow $5.68 & \cellcolor[HTML]{CCE4F0}$\color{green} \uparrow $32.78 & $\color{green} \uparrow $9.12 & $\color{red} \downarrow $5.85 & \multicolumn{1}{|c}{$\color{green} \uparrow $32.78} & \multicolumn{1}{|c}{$\color{green} \uparrow $21.30} \\
  & \multicolumn{1}{|c|}{ } & \multicolumn{1}{|c|}{\ours} & $\color{red} \downarrow $12.36 & $\color{green} \uparrow $16.17 & $\color{green} \uparrow $9.43 & \cellcolor[HTML]{CCE4F0}$\color{green} \uparrow $36.97 & $\color{green} \uparrow $16.89 & $\color{green} \uparrow $12.65 & \multicolumn{1}{|c}{$\color{green} \uparrow $36.97} & \multicolumn{1}{|c}{$\color{green} \uparrow $42.78} \\
\cline{2-11}
& \multicolumn{1}{|c|}{ } & \multicolumn{1}{|c|}{100\% Specific Domain} & $\color{red} \downarrow $21.76 & $\color{green} \uparrow $7.94 & $\color{red} \downarrow $8.82 & \cellcolor[HTML]{CCE4F0}$\color{green} \uparrow $63.20 & $\color{red} \downarrow $4.41 & $\color{red} \downarrow $10.00 & \multicolumn{1}{|c}{$\color{green} \uparrow $63.20} & \multicolumn{1}{|c}{$\color{red} \downarrow $37.05} \\
   & \multicolumn{1}{|c|}{3} & \multicolumn{1}{|c|}{Uniform Distribution of Non-Target Domains} & $\color{red} \downarrow $9.98 & $\color{green} \uparrow $12.06 & $\color{red} \downarrow $15.01 & \cellcolor[HTML]{CCE4F0}$\color{green} \uparrow $55.78 & $\color{red} \downarrow $4.11 & $\color{red} \downarrow $14.13 & \multicolumn{1}{|c}{$\color{green} \uparrow $55.78} & \multicolumn{1}{|c}{$\color{red} \downarrow $31.17} \\
  & \multicolumn{1}{|c|}{ } & \multicolumn{1}{|c|}{\ours} & $\color{red} \downarrow $11.47 & $\color{green} \uparrow $11.18 & $\color{red} \downarrow $6.76 & \cellcolor[HTML]{CCE4F0}$\color{green} \uparrow $55.40 & $\color{green} \uparrow $16.49 & $\color{green} \uparrow $6.81 & \multicolumn{1}{|c}{$\color{green} \uparrow $55.40} & \multicolumn{1}{|c}{$\color{green} \uparrow $16.25} \\
\cline{2-11}
 & \multicolumn{1}{|c|}{ } & \multicolumn{1}{|c|}{100\% Specific Domain} & $\color{red} \downarrow $27.94 & $\color{green} \uparrow $2.35 & $\color{red} \downarrow $11.47 & \cellcolor[HTML]{CCE4F0}$\color{green} \uparrow $66.15 & $\color{red} \downarrow $12.65 & $\color{red} \downarrow $12.59 & \multicolumn{1}{|c}{$\color{green} \uparrow $66.15} & \multicolumn{1}{|c}{$\color{red} \downarrow $62.30} \\
   & \multicolumn{1}{|c|}{4} & \multicolumn{1}{|c|}{Uniform Distribution of Non-Target Domains} & $\color{red} \downarrow $13.82 & $\color{green} \uparrow $13.53 & $\color{red} \downarrow $11.18 & \cellcolor[HTML]{CCE4F0}$\color{green} \uparrow $58.46 & $\color{red} \downarrow $6.57 & $\color{red} \downarrow $6.47 & \multicolumn{1}{|c}{$\color{green} \uparrow $58.46} & \multicolumn{1}{|c}{$\color{red} \downarrow $24.51} \\
  & \multicolumn{1}{|c|}{ } & \multicolumn{1}{|c|}{\ours} & $\color{red} \downarrow $10.12 & $\color{green} \uparrow $10.00 & $\color{red} \downarrow $4.21  & \cellcolor[HTML]{CCE4F0}$\color{green} \uparrow $61.30 & $\color{red} \downarrow $5.30 & $\color{red} \downarrow $6.74 & \multicolumn{1}{|c}{$\color{green} \uparrow $61.30} & \multicolumn{1}{|c}{$\color{red} \downarrow $16.37} \\
\midrule
\multirow{12}{*}{Code} & \multicolumn{1}{|c|}{ } & \multicolumn{1}{|c|}{100\% Specific Domain} & $\color{red} \downarrow $3.82 & $\color{green} \uparrow $7.35 & $\color{red} \downarrow $17.35 & $\color{green} \uparrow $9.12 & \cellcolor[HTML]{AECFD4}$\color{green} \uparrow $10.46 & $\color{red} \downarrow $7.29 & \multicolumn{1}{|c}{$\color{green} \uparrow $10.46} & \multicolumn{1}{|c}{$\color{red} \downarrow $11.99} \\
& \multicolumn{1}{|c|}{1} & \multicolumn{1}{|c|}{Uniform Distribution of Non-Target Domains} & $\color{red} \downarrow $3.76 & $\color{green} \uparrow $5.09 & $\color{red} \downarrow $7.80 & $\color{green} \uparrow $7.68 & \cellcolor[HTML]{AECFD4}$\color{green} \uparrow $5.23 & $\color{green} \uparrow $5.65 & \multicolumn{1}{|c}{$\color{green} \uparrow $5.23} & \multicolumn{1}{|c}{$\color{green} \uparrow $6.86} \\
  & \multicolumn{1}{|c|}{ } & \multicolumn{1}{|c|}{\ours} & $\color{red} \downarrow $5.06 & $\color{green} \uparrow $11.82 & $\color{red} \downarrow $8.76 & $\color{green} \uparrow $8.96 & \cellcolor[HTML]{AECFD4}$\color{green} \uparrow $5.98 & $\color{green} \uparrow $8.19 & \multicolumn{1}{|c}{$\color{green} \uparrow $5.98} & \multicolumn{1}{|c}{$\color{green} \uparrow $15.15} \\
\cline{2-11}
   & \multicolumn{1}{|c|}{ } & \multicolumn{1}{|c|}{100\% Specific Domain} & $\color{red} \downarrow $9.71 & $\color{red} \downarrow $6.47 & $\color{red} \downarrow $7.94 & $\color{green} \uparrow $5.29 & \cellcolor[HTML]{AECFD4}$\color{green} \uparrow $47.28 & $\color{red} \downarrow $6.18 & \multicolumn{1}{|c}{$\color{green} \uparrow $47.28} & \multicolumn{1}{|c}{$\color{red} \downarrow $25.01} \\
   & \multicolumn{1}{|c|}{2} & \multicolumn{1}{|c|}{Uniform Distribution of Non-Target Domains} & $\color{red} \downarrow $7.90 & $\color{green} \uparrow $13.49 & $\color{red} \downarrow $9.05 & $\color{green} \uparrow $12.03 & \cellcolor[HTML]{AECFD4}$\color{green} \uparrow $38.31 & $\color{red} \downarrow $17.76 & \multicolumn{1}{|c}{$\color{green} \uparrow $38.31} & \multicolumn{1}{|c}{$\color{red} \downarrow $9.19} \\
  & \multicolumn{1}{|c|}{ } & \multicolumn{1}{|c|}{\ours} & $\color{red} \downarrow $12.22 & $\color{green} \uparrow $15.04 & $\color{green} \uparrow $6.78 & $\color{green} \uparrow $16.28 & \cellcolor[HTML]{AECFD4}$\color{green} \uparrow $39.77 & $\color{green} \uparrow $6.14 & \multicolumn{1}{|c}{$\color{green} \uparrow $39.77} & \multicolumn{1}{|c}{$\color{green} \uparrow $32.02} \\
\cline{2-11}
& \multicolumn{1}{|c|}{ } & \multicolumn{1}{|c|}{100\% Specific Domain} & $\color{red} \downarrow $22.35 & $\color{red} \downarrow $8.82 & $\color{red} \downarrow $14.12 & $\color{green} \uparrow $7.94 & \cellcolor[HTML]{AECFD4}$\color{green} \uparrow $61.95 & $\color{red} \downarrow $10.02 & \multicolumn{1}{|c}{$\color{green} \uparrow $61.95} & \multicolumn{1}{|c}{$\color{red} \downarrow $47.37} \\
   & \multicolumn{1}{|c|}{3} & \multicolumn{1}{|c|}{Uniform Distribution of Non-Target Domains} & $\color{red} \downarrow $9.96 & $\color{red} \downarrow $15.07 & $\color{red} \downarrow $8.46 & $\color{green} \uparrow $5.33 & \cellcolor[HTML]{AECFD4}$\color{green} \uparrow $55.62 & $\color{red} \downarrow $10.04 & \multicolumn{1}{|c}{$\color{green} \uparrow $55.62} & \multicolumn{1}{|c}{$\color{red} \downarrow $38.20} \\
  & \multicolumn{1}{|c|}{ } & \multicolumn{1}{|c|}{\ours} & $\color{red} \downarrow $17.39 & $\color{green} \uparrow $17.25 & $\color{red} \downarrow $9.09 & $\color{green} \uparrow $12.31 & \cellcolor[HTML]{AECFD4}$\color{green} \uparrow $56.12 & $\color{green} \uparrow $6.19 & \multicolumn{1}{|c}{$\color{green} \uparrow $56.12} & \multicolumn{1}{|c}{$\color{green} \uparrow $9.27} \\
\cline{2-11}
 & \multicolumn{1}{|c|}{ } & \multicolumn{1}{|c|}{100\% Specific Domain} & $\color{red} \downarrow $26.18 & $\color{red} \downarrow $7.06 & $\color{red} \downarrow $8.82 & $\color{red} \downarrow $3.24 & \cellcolor[HTML]{AECFD4}$\color{green} \uparrow $64.76 & $\color{red} \downarrow $22.65 & \multicolumn{1}{|c}{$\color{green} \uparrow $64.76} & \multicolumn{1}{|c}{$\color{red} \downarrow $67.95} \\
   & \multicolumn{1}{|c|}{4} & \multicolumn{1}{|c|}{Uniform Distribution of Non-Target Domains} & $\color{red} \downarrow $14.01 & $\color{red} \downarrow $13.86 & $\color{red} \downarrow $6.57 & $\color{green} \uparrow $6.66 & \cellcolor[HTML]{AECFD4}$\color{green} \uparrow $58.06 & $\color{red} \downarrow $13.93 & \multicolumn{1}{|c}{$\color{green} \uparrow $58.06} & \multicolumn{1}{|c}{$\color{red} \downarrow $41.71} \\
  & \multicolumn{1}{|c|}{ } & \multicolumn{1}{|c|}{\ours} & $\color{red} \downarrow $5.66 & $\color{green} \uparrow $4.42 & $\color{red} \downarrow $10.10  & $\color{green} \uparrow $9.95 & \cellcolor[HTML]{AECFD4}$\color{green} \uparrow $59.71 & $\color{red} \downarrow $13.57 & \multicolumn{1}{|c}{$\color{green} \uparrow $59.71} & \multicolumn{1}{|c}{$\color{red} \downarrow $14.96} \\
\bottomrule
\end{tabular}
}
\caption{\label{tab:flexible_domain_expansion}
Results of \ours on flexible domain expansion, 
we computed the average percentage change across various models for each method.
``Sum. (\%)'' denotes the total percentage of performance variations across all target and non-target domain tasks. Symbols ${\color{green} \uparrow}$ and ${\color{red} \downarrow}$ indicate an increase or decrease in the percentage of scores (\%) compared to the \textit{initial state} before supervised fine-tuning. The current target domain is highlighted using the corresponding domain color in \Cref{fig:pipeline}, which includes law, medicine, finance, science, code, and general fields. 
}
\end{table*}

\subsubsection{Performance Variations in Target and Non-Target Domains}
\label{subsubsec:appendix_performance_variations_of_target_and_non-target_domains}

\begin{figure*}[ht]
    \begin{center}
    \centerline{\includegraphics[width=\linewidth]{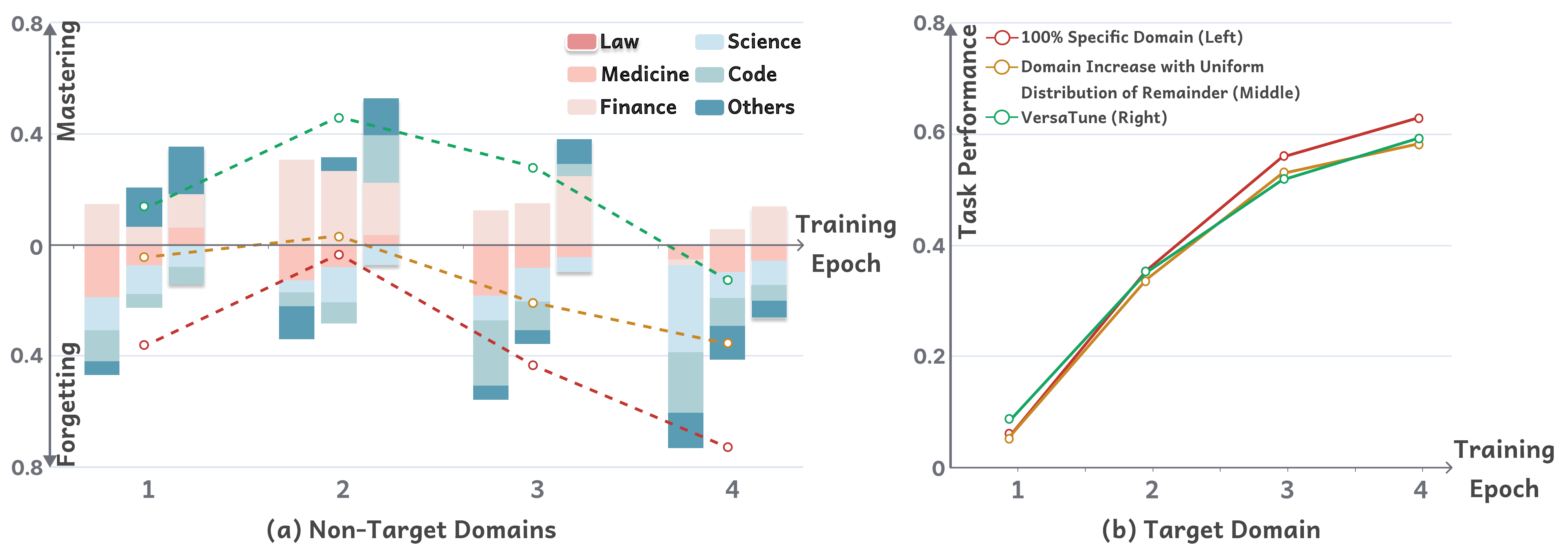}}
    \caption{Domain expansion results for the \textbf{\textit{law}} domain, including non-target domains (a) and target domain (b).}
    \label{fig:domain_expansion_law}
    \end{center}
    \vskip -0.2in
\end{figure*}

\begin{figure*}[ht]
    \begin{center}
    \centerline{\includegraphics[width=\linewidth]{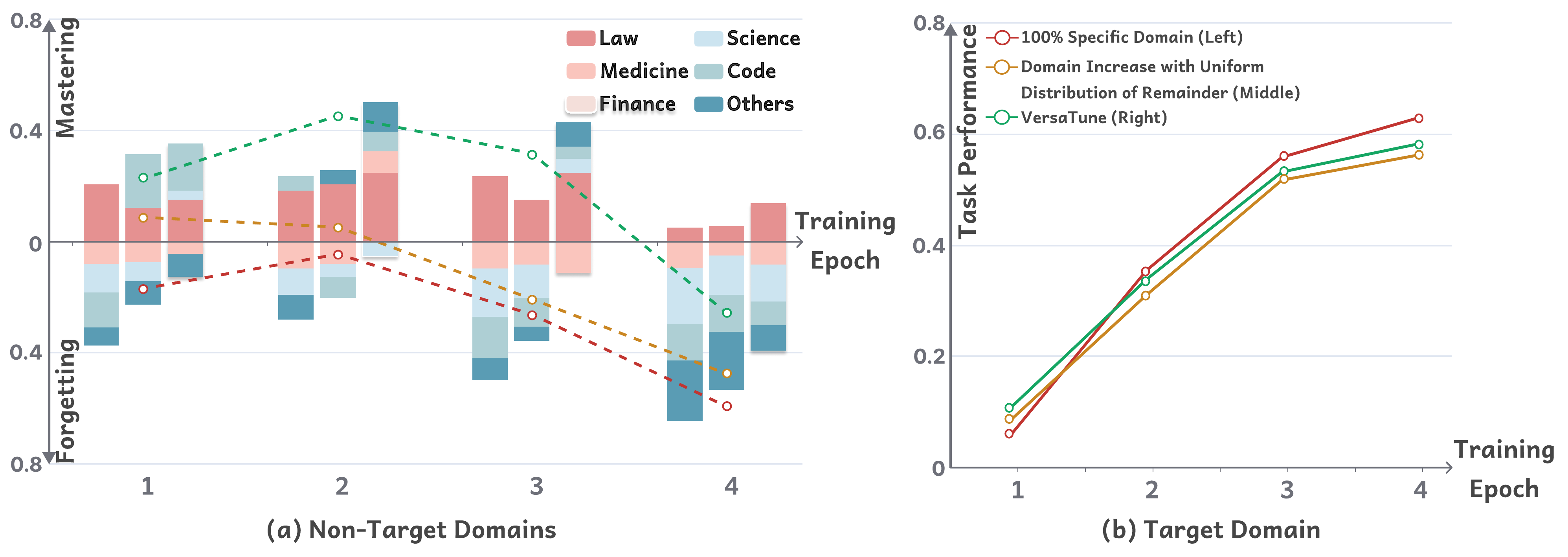}}
    \caption{Domain expansion results for \textbf{\textit{finance}} domain, including non-target domains (a) and target domain (b).}
    \label{fig:domain_expansion_finance}
    \end{center}
    \vskip -0.2in
\end{figure*}

\begin{figure*}[ht]
    \begin{center}
    \centerline{\includegraphics[width=\linewidth]{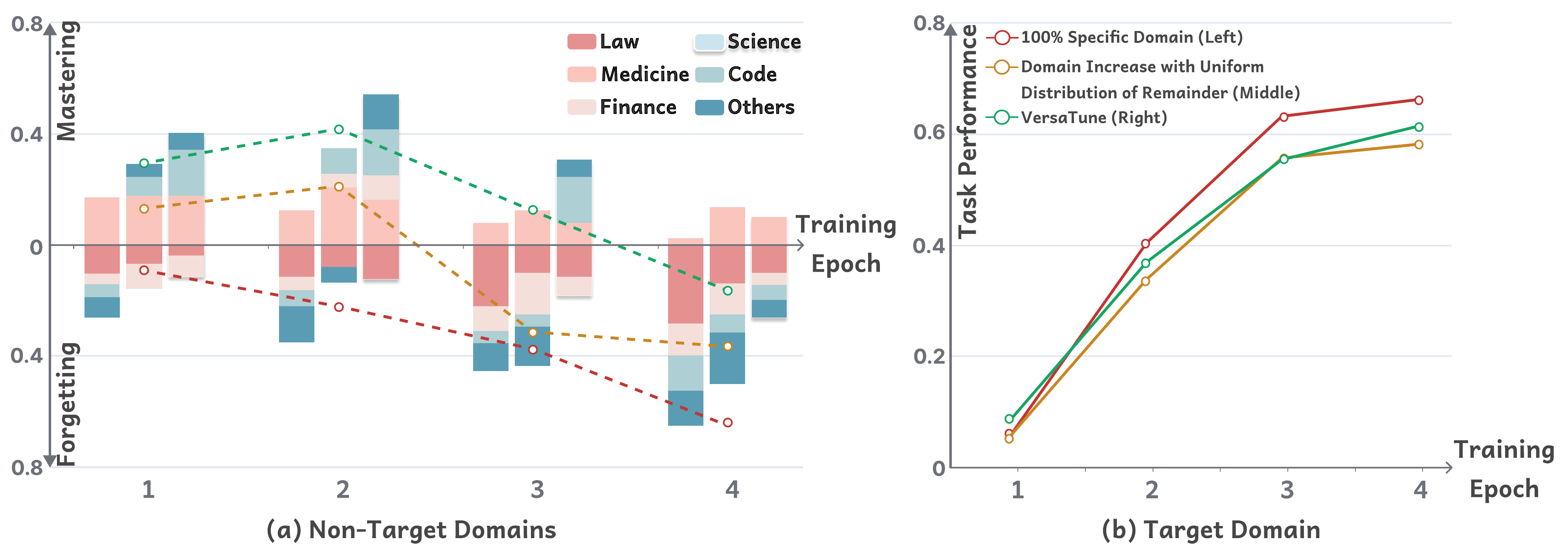}}
    \caption{Domain expansion results for \textbf{\textit{science}} domain, including non-target domains (a) and target domain (b).}
    \label{fig:domain_expansion_science}
    \end{center}
    \vskip -0.2in
\end{figure*}

\begin{figure*}[ht]
    \begin{center}
    \centerline{\includegraphics[width=\linewidth]{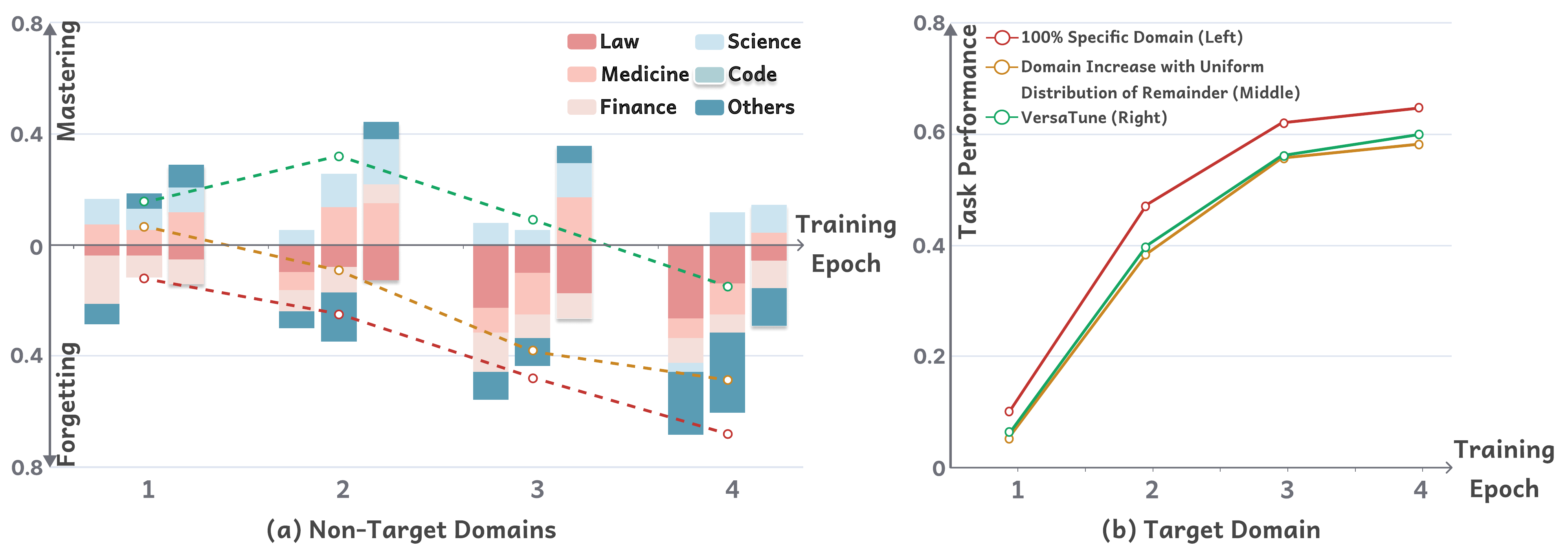}}
    \caption{Domain expansion results for the \textbf{\textit{code}} domain, including non-target domains (a) and target domain (b).}
    \label{fig:domain_expansion_code}
    \end{center}
    \vskip -0.2in
\end{figure*}

Here we exhibit the performance of the target domain and other non-target domains under the domain expansion scenario, as realized by \Cref{alg:domain_expansion}. 
\Cref{fig:domain_expansion_medicine}, \Cref{fig:domain_expansion_law}-\ref{fig:domain_expansion_code} illustrate the changes in target domain capabilities and non-target domain capabilities during the fine-tuning process when focused on a specific domain, providing experimental results for flexible domain expansion. 
In each figure, the \underline{stacked group bar chart (left)} depicts the percentage change in performance for non-target domains relative to their pre-fine-tuning states, with the positive direction on the y-axis indicating performance improvement and the negative direction signifying a decline. The \underline{line chart (right)} represents the overall change across all non-target domains for three distinct strategies, with color legends corresponding to those of the line chart on the right. The right-side chart depicts the percentage increase in performance for the current target domain. We employed the Qwen-2.5-7B and Qwen-2.5-14B models, and the mean percentage change in model performance when focusing on domain enhancement is presented in both the stacked group bar chart and the line chart. 
Three interesting phenomena are observed from the outcomes: 

\begin{itemize}[leftmargin=*]
    \item \textbf{Absolute Count vs. Proportion.} 
    Notably, by the \textit{second epoch} of training, the performance degradation across non-target domains tends to be mitigated to some extent, and there is even a positive trend in capability enhancement in some cases. We attribute this phenomenon to the fact that the absolute quantity of instances for each domain, relative to the domain distribution, has a predominant influence on model performance at this stage.
    \item \textbf{Domain Interactions.} 
    Domains are not entirely orthogonal to each other, and there is a degree of mutual reinforcement among them: \underline{\textbf{\textit{(1)}}} Enhancing capabilities in the medicine domain can boost performance in the science domain to a certain degree (\Cref{fig:domain_expansion_medicine}). 
    \underline{\textbf{\textit{(2)}}} Models' capabilities in law and finance are mutually reinforcing, promoting each other's development (\Cref{fig:domain_expansion_law} and \Cref{fig:domain_expansion_finance}). 
    \underline{\textbf{\textit{(3)}}} Augmenting the model's code-related capabilities can also, to some extent, improve its ability to solve scientific problems, which is likely due to the shared reasoning and logical structuring required across these domains (\Cref{fig:domain_expansion_code}). 
    \item \textbf{Domain Mastery Efficiency.} 
    From the slope of the target domain performance increase in \Cref{fig:domain_expansion_medicine}, \Cref{fig:domain_expansion_law}-\ref{fig:domain_expansion_code} (b), it is evident that the model's efficiency in mastering knowledge of a specific domain diminishes over training. In other words, as training progresses, the model's grasp of the target domain approaches saturation, while its performance on non-target domains declines sharply. Consequently, greater emphasis should be placed on mitigating losses in non-target domains during this phase, aiming to strike a balance between domain expansion and the salvage of capabilities in non-target domains, which is also shown in \Cref{fig:domain_expansion_ablation}.
\end{itemize}

\clearpage

\subsubsection{Importance of Proportion Thresholds}
\label{subsubsec:appendix_proportion_threshold}

\begin{figure*}[ht]
    \begin{center}
    \centerline{\includegraphics[width=\linewidth]{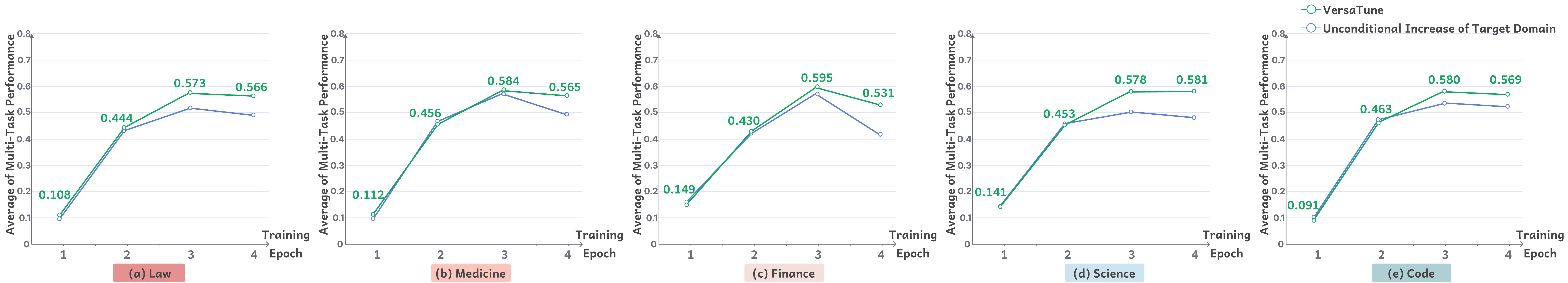}}
    \caption{Line chart of the multi-task performances of models across different domains during the domain expansion process. We calculated the average percentage change for both target and non-target domains comparing to the \textit{initial state}. Additionally, we highlighted the performance changes of the \ours at various checkpoints using green numerical annotations.}
    \label{fig:domain_expansion_ablation}
    \end{center}
    \vskip -0.2in
\end{figure*}

Here we describe the significance of establishing proportion thresholds for specific domains during domain expansion in detail. 
We compare \ours with those implementing an \textit{unconditional dynamic increase of the specific domain}, where we remove the implementation of Line 8 in \Cref{alg:domain_expansion}, to ablate the component of criteria for determining the upper limit of domain expansion. 
In \Cref{fig:domain_expansion_ablation}, we present the trends in the overall multi-domain performance of the models under specific domain expansion for each domain. It can be observed that, for the majority of domains, the gap in average multi-task performance between models trained with \ours and those without an upper limit on domain proportion becomes increasingly pronounced after the second or third epoch. 
We deduce that this occurs due to the fact that as training progresses, the models' ability to learn within the target domain becomes nearly maximized. Enhancing the emphasis on the current domain beyond this point yields marginal benefits and may even result in a substantial degradation of performance in other domains.
Notably, between the \underline{\textit{second}} and \underline{\textit{third}} epochs of supervised fine-tuning, the model reaches a balance where the efficiency of improvement in the target domain is matched by the rate of performance degradation in non-target domains. 
The finding shows that the criteria for determining the upper limit on the proportion of a specific domain during domain expansion, 
has mitigated the loss of capabilities in other domains experienced by the target model $M_\theta$ during the fine-tuning process. 
Moreover, it ensures gains in the capacity for the current domain of interest.

In summary, \ours exhibits the following properties:

\begin{itemize} 
    \item \textbf{Efficient.} \ours employs distribution consistency training of the domain knowledge proportion during models' SFT stage, providing an efficient data composition strategy for enhancing versatile capabilities (for \hyperlink{C2}{\textbf{\textit{C2}}}).
    \item \textbf{Flexible.} \ours can be flexibly adapted to scenarios that expand performance on specific domain tasks while minimizing the degradation of the model's capabilities in other non-target domains (for \hyperlink{C1}{\textbf{\textit{C1}}}, \hyperlink{C3}{\textbf{\textit{C3}}}).
    \item \textbf{Robust.} Our strategy achieves significant performance improvements in open-sourced models with parameter sizes ranging from 7B-32B, adding to the effectiveness of \ours (for \hyperlink{C1}{\textbf{\textit{C1}}}, \hyperlink{C2}{\textbf{\textit{C2}}} and \hyperlink{C3}{\textbf{\textit{C3}}}).
\end{itemize}

\clearpage

\section{Prompts}
\label{sec:appendix_prompts}

We present the prompts that are employed throughout our pipeline in \ours. 
Only the English version is presented due to LaTeX compilation issues with non-English languages. 

\begin{tcolorbox}[title=\textbf{Prompt: Domain Probability Inference}, colback=gray!10, colframe=gray!50!black, boxrule=1pt]
You are a data domain annotation expert, and you currently have the following six data domains: law, medical \&\& health care, finance, science, code, and other. Please classify the following text fragment based on their topic and structure by providing the probability distribution of its belonging to each category, where the sum of probabilities across all domain categories equals 1, without additional commentary: 

\vspace{0.5cm}

\textbf{\# Text}\\
\{text\_content\}

\vspace{0.5cm}

\hdashrule{\textwidth}{1pt}{3pt}

\vspace{0.5cm}

\textbf{Output Format:}  
\begin{verbatim}
```json
{
    "Law": "",
    "Medicine": "",
    "Finance": "",
    "Sciencee": "",
    "Code": "",
    "Other": ""
}
'''
\end{verbatim}

\end{tcolorbox}

\end{document}